\PassOptionsToPackage{table,xcdraw}{xcolor}
\documentclass[sn-mathphys]{sn-jnl}

\jyear{2021}%

\theoremstyle{thmstyleone}%
\theoremstyle{thmstyletwo}%

\theoremstyle{thmstylethree}%

\usepackage{enumitem}
\definecolor{mygray}{gray}{0.}

\usepackage{tabularx}
\usepackage{multirow}
\usepackage{pdflscape}
\usepackage{graphicx}

\usepackage{booktabs}      
\usepackage{longtable}     
\usepackage{multirow}      
\usepackage{array}         
\usepackage{caption}       
\usepackage{float}         
\usepackage{amsmath}
\usepackage{adjustbox} 
\usepackage{hhline}

\usepackage{amssymb}
\usepackage{pifont}
\newcommand{\cmark}{\ding{51}}%
\newcommand{\xmark}{\ding{55}}%

\begin{document}

\title[SLEEPYLAND]{SLEEPYLAND: trust begins with fair evaluation of automatic sleep staging models}


\author[1,2]{\fnm{Alvise} \sur{Dei Rossi}}\email{alvise.dei.rossi@usi.ch}

\author[2]{\fnm{Matteo} \sur{Metaldi}}\email{matteo.metaldi@supsi.ch}

\author[2,3]{\fnm{Michal} \sur{Bechny}}\email{michal.bechny@supsi.ch}

\author[4]{\fnm{Irina} \sur{Filchenko}}\email{irina.filchenko@insel.ch}

\author[4]{\fnm{Julia} \sur{van der Meer}}\email{julia.vandermeer@insel.ch}

\author[4]{\fnm{Markus H.} \sur{Schmidt}}\email{markus.schmidt@insel.ch}

\author[4]{\fnm{Claudio L.A.} \sur{Bassetti}}\email{claudio.bassetti@insel.ch}

\author[3,4]{\fnm{Athina} \sur{Tzovara}}\email{athina.tzovara@inf.unibe.ch}

\author[2]{\fnm{Francesca D.} \sur{Faraci}}\email{francesca.faraci@supsi.ch}

\author*[2,5]{\fnm{Luigi} \sur{Fiorillo}}\email{luigi.fiorillo@supsi.ch}

\affil[1]{\orgdiv{Faculty of informatics}, \orgname{Università della Svizzera Italiana}, \orgaddress{\street{Via Giuseppe Buffi 13}, \city{Lugano}, \postcode{6900}, \country{Switzerland}}}

\affil*[2]{\orgdiv{Institute of Digital Technologies for Personalized Healthcare $\vert$ MeDiTech}, Department of Innovative Technologies, \orgname{University of Applied Sciences and Arts of Southern Switzerland}, \orgaddress{\street{Via la Santa 1}, \city{Lugano}, \postcode{6962}, \country{Switzerland}}}

\affil[3]{\orgdiv{Institute of Informatics}, \orgname{University of Bern}, \orgaddress{\street{Neubr{\"u}ckstrasse 10}, \city{Bern}, \postcode{3012}, \country{Switzerland}}}

\affil[4]{\orgdiv{Sleep Wake Epilepsy Center $\vert$ NeuroTec},  Department of Neurology, \orgname{Inselspital, Bern University Hospital, University of Bern}, \orgaddress{\street{Freiburgstrasse}, \city{Bern}, \postcode{3010}, \country{Switzerland}}}

\affil*[5]{\orgdiv{Neurocenter of Southern Switzerland}, \orgname{Ente Ospedaliero Cantonale}, \orgaddress{\street{Via Tesserete 46}, \city{Lugano}, \postcode{6900}, \country{Switzerland}}}


\abstract{
Despite advances in deep learning for automatic sleep staging, clinical adoption remains limited due to challenges in fair model evaluation, generalization across diverse datasets, model bias, and variability in human annotations. We present SLEEPYLAND, an open-source sleep staging evaluation framework designed to address these barriers. It includes $\thicksim$220'000 hours in-domain sleep recordings, and $\thicksim$84'000 hours out-of-domain sleep recordings, spanning a broad range of ages, sleep-wake disorders, and hardware setups. We release pre-trained models based on high-performing state-of-the-art architectures (i.e., U-Sleep, DeepResNet and SleepTransformer), and evaluate them under standardized conditions across single- and multi-channel EEG/EOG configurations. We introduce SOMNUS, an ensemble combining models across architectures and channel setups via soft-voting. SOMNUS achieves robust performance across twenty-four different datasets, with macro-F1 scores between 68.7\% and 87.2\%, outperforming individual models in 94.9\% of cases. Notably, SOMNUS surpasses previous state-of-the-art methods, even including cases where compared models were trained in-domain while SOMNUS treated the same data as out-of-domain. Models trained on individual datasets show catastrophic generalization failures, while SOMNUS maintains robust performance across both in-domain and out-of-domain evaluations. Using a subset of the Bern Sleep-Wake Registry (N=6'633), we quantify model biases linked to age, gender, AHI, and PLMI, showing that while ensemble improves robustness, no model architecture consistently minimizes bias in performance and clinical markers estimation. In evaluations on out-of-domain multi-annotated datasets (DOD-H, DOD-O), SOMNUS exceeds the best human scorer, i.e., MF1 85.2\% vs 80.8\% on DOD-H, and 80.2\% vs 75.9\% on DOD-O, better reproducing the scorer consensus than any individual expert (kappa = 0.89/0.85 and ACS = 0.95/0.94 for healthy/OSA cohorts). Finally, we introduce ensemble disagreement metrics - entropy and inter-model divergence based - predicting regions of scorer disagreement with ROC AUCs up to 0.828, offering a data-driven proxy for human uncertainty.
}




\maketitle

\section*{Introduction}\label{intro} 

Polysomnography (PSG) is a golden standard of sleep diagnostics widely used common disorders like sleep breathing disorders, narcolepsy, and sleep-related movement disorders. PSG involves the recording of multiple biosignals during the whole night. These may include electroencephalographic activity (electroencephalogram - EEG), eye movements (electrooculogram - EOG), muscle activity (electromyogram EMG - derivations for the chin and legs), body position (video camera and accelerometer), heart rhythm (electrocardiogram), breathing (respiratory airflow, oxygen saturation, and respiratory effort indicators), as well as other vital parameters. Sleep staging is the procedure of extracting sleep cycle information from PSG signals. Wakefulness and sleep stages (i.e., NREM1, NREM2, NREM3 and REM) can be identified mainly analyzing the EEG, EOG and EMG signals. Sleep staging is performed worldwide by sleep experts, most commonly according to the AASM scoring manual (version 2.4) \cite{berry2017aasm}.\\

Many machine learning (ML) and deep learning (DL) based algorithms have been proposed to tackle the sleep staging task \cite{ronzhina2012sleep, csen2014comparative, radha2014comparison, aboalayon2016sleep, boostani2017comparative, fiorillo2019automated, fiorilloautomated}, achieving very good results in terms of overall accuracy. In particular, DL based staging algorithms have recently shown higher performances compared to the traditional ML approaches. However, none of these algorithms has ever been introduced into the daily clinical routine \cite{fiorilloautomated}. This gap is largely due to several interrelated challenges. \textit{Which algorithm should I use, and how can I fairly compare their performance?} There is a lack of standardized protocols for training, validating, and testing these algorithms, making direct performance comparisons across studies difficult and often unreliable. \textit{Will this algorithm work well beyond the dataset it was trained on?} Only a few of these algorithms have been evaluated on out-of-domain data, resulting in limited evidence of their ability to generalize well across different hardware configurations or sleep labs. \textit{Can the model learn from and adapt to data from diverse sources and patient populations?} Sleep data is heterogeneous, varying across acquisition systems, clinical sites, and patient characteristics, including those with different sleep-wake disorders. Models must be trained on and evaluated against this diversity, quantifying their bias against the heterogeneity, e.g., clinical confounders \cite{perslev2021u, olesen2021automatic, fiorillo2023u}. \textit{How can we account for differences in human scorers?} Inter- and intra-scorer variability is a major challenge, highlighting the need for systems that can be personalized or adapted to individual physicians’ scoring styles \cite{fiorillo2023multi}.\\ 

Addressing these challenges requires - as a first concrete step - access to large, diverse datasets to train and validate algorithms robustly. In this regard, the National Sleep Research Resource (NSRR) \cite{zhang2018national, zhang2024national} has emerged as a resource of primary importance, providing a centralized repository of harmonized sleep data from multiple cohorts and clinical trials. Sleep data are difficult to store, manage, and, most importantly, share with the sleep community for study research purposes (up to 1GB for each PSG). In 2014, the NSRR, supported by the National Heart, Lung, and Blood Institute (NHLBI), launched a game-changing open research data (ORD) program: "\textit{NSRR mission is to advance sleep and circadian science by supporting secondary data analysis, algorithmic development, and signal processing through the sharing of high-quality data sets}" \cite{zhang2018national, zhang2024national}. To date, the NSRR provides the opportunity to search and download large amounts of harmonized sleep data from multiple cohorts, clinical trials, and other data sources (up to 10TB currently stored on the resource). The NSRR repository currently hosts 36 datasets, including an archive of almost 53k individual represented, i.e., PSG recordings, actigraphy, questionnaires and demographic data. Most importantly, all datasets are harmonized, i.e., they are preprocessed and formatted using a common standard, facilitating integration and analysis across heterogeneous sources. \\

The NSRR ORD community has recently provided an open source software \emph{Luna} (\url{https://zzz.bwh.harvard.edu/luna/}) \cite{zhang2024national}, a C/C++ and R based open-source library able to further manually annotate and analyze the large numbers of NSRR sleep studies. \emph{Luna} is mainly focused on the overall automation of sleep analysis (e.g., time-frequency analysis, hypnograms, sleep patterns detection, etc...). The NSRR ORD community also released the \emph{Moonlight} (\url{https://remnrem.net/}) tool, a useful interactive viewer for PSG data implemented on top of the \emph{Luna} package. Combrisson et al. also developed \emph{Sleep} \cite{combrisson2017sleep}, an open-source python-based graphical user interface designed for the visualization, scoring, and analysis of sleep data. \emph{Sleep} allows to dynamically display polysomnographic data, including spectrograms, hypnograms, and topographic maps. It also integrates automated detection algorithms for essential sleep features, such as spindles, K-complexes, slow waves, and REM events, alongside essential signal processing tools like re-referencing and filtering. As \emph{Luna}, \emph{Sleep} is compatible as well with common formats like the European Data Format (EDF).\\

To date, the extensive NSRR environment does not yet include advanced DL-based staging models, nor has any open-source tool been released to systematically compare and evaluate recently proposed models. There is a clear need for a user-friendly and robust platform to benchmark high-performing algorithms, particularly as researchers seek generalizable and reliable solutions. To address the challenges mentioned above, and enhance the usability of the recently implemented sleep staging models, we introduce SLEEPYLAND, an open-source repository that centralizes and benchmarks high-performing ML- and DL-based sleep staging algorithms. Built on the foundation of multiple-diverse sleep datasets, including the highest number of PSG recordings from the NSRR repository, SLEEPYLAND facilitates the rigorous evaluation of different models while ensuring generalizability across different domains. To make our tool accessible to non-technical users, we also developed a user-friendly graphical user interface (GUI), lowering the entry barrier for researchers interested in testing and exploring its capabilities without requiring advanced computational expertise. The GUI runs locally on the user’s device, ensuring that all data remains private and secure, with no data leaving the user’s environment.\\

SLEEPYLAND key contributions can be summarized as follows: \\

\begin{enumerate}[label=(\roman*),font=\itshape]

\item \textbf{Benchmarking, ensembling and generalization}. We propose a unified protocol to train, validate and test high-performing sleep staging algorithms on the same set of data. For the first time, we publicly release all the models pre-trained on the largest collection of PSG recordings to date, enabling researchers to evaluate models on their own data and benchmark novel approaches using a common framework. We assess the generalizability of multiple algorithms and their ensemble, i.e., \textbf{SOMNUS} (\textbf{S}oft-voting \textbf{O}ver \textbf{M}ultiple \textbf{N}etworks for \textbf{U}nified \textbf{S}leep-staging), on the largest out-of-domain evaluation dataset used in literature to date. We demonstrate that SOMNUS leads to consistently superior performance across all evaluation settings. SOMNUS combines predictions from multiple architectures and channel configurations, and achieves macro-F1 scores that are better than any individual model. The ensemble mitigates weaknesses specific to network architecture and signal channel configurations, yielding consistently superior and more stable performance across diverse test conditions. SOMNUS surpasses previous state-of-the-art methods, even including cases where compared models were trained in-domain while SOMNUS treated the same data as out-of-domain.\\

\item \textbf{Model-bias on performance and clinical markers}. Exploiting the ensemble of the models, we quantify the bias on performance and prediction-derived clinical markers across demographics and clinical subgroups, including age, gender, apnea-hypopnea index (AHI) and periodic limb movement index (PLMI), on a huge out-of-domain dataset. SOMNUS does not reduce or compensate for demographic or clinical model bias. All models - including ensembles - show comparable bias patterns with respect to age, gender, AHI, and PLMI. No architecture seems to consistently minimizes errors in derived clinical markers.\\

\item \textbf{Model-ensemble versus human-ensemble}. We tested our tool on out-of-domain datasets scored by multiple human experts. We find that SOMNUS exceeds individual experts in alignment with the consensus, consistently aligning more closely with collective annotations than with any single expert. Its internal variability, quantified via entropy and model-to-model divergence, reliably identifies ambiguous regions, offering a data-driven proxy for human uncertainty in sleep staging.
\end{enumerate}

\section*{Results}

\subsection*{Datasets and experiments} 

We train and evaluate three state-of-the-art deep learning models\footnote{These models were chosen based on their high performance across multiple datasets and their distinct architectural approaches: U-Sleep employs a fully convolutional approach, DeepResNet incorporates recurrent compmonents, and SleepTransformer leverages attention-based modules.}, including U-Sleep \cite{perslev2021u}, DeepResNet \cite{olesen2021automatic}, and SleepTransformer \cite{phan2022sleeptransformer} using harmonized subsets of the NSRR repository - comprising up to $27'494$ recordings (details in Supplementary Notes: Datasets). The NSRR data are processed to facilitate fair benchmarking and generalization across the different models. Our experimental framework includes both single-channel (EEG or EOG) and multi-channel (e.g., EEG and EOG) configurations. We currently exclude the chin EMG channel mainly because not all recordings include at least one EMG channel (this choice allows us to retain the maximum number of recordings and ensure consistency in the amount of data across all experiments). In both single- and multi-channel configurations, we consider all possible EEG and EOG derivations available within each specific dataset (details in Supplementary Table~1). The data pre-processing and the data sampling across all the datasets is implemented as described in \cite{perslev2021u}, with minor adjustments to align with SLEEPYLAND’s modular architecture. No additional filtering applied to the signals during the pre-processing procedure.
To enhance the robustness and heterogeneity of the benchmark datasets, and to test the performance of the models across different clinical subpopulations, we exploit the Bern Sleep-Wake Registry (BSWR) \cite{calle20180733} - out-of-domain real-world dataset. The dataset comprises $8'410$ recordings from patients and healthy subjects aged $0-91$ years, collected at the Department of Neurology, Inselspital, University Hospital Bern, in a clinical routine between 2000 and 2021. This dataset, approved for secondary use by the cantonal ethics committee (KEK-Nr. 2022-00415), uniquely spans the full spectrum of sleep-wake disorders, including cases with multiple comorbidities and non-sleep-related conditions. Unlike publicly available datasets, the BSWR provides an exceptionally diverse PSG dataset, making it a critical addition to our benchmarking efforts.
Additionally, to evaluate the performance of the ensemble of the sleep staging models compared to the human experts, we leverage the Dreem Open Dataset-Healthy (DOD-H) and Dreem Open Dataset-Obstructive (DOD-O) \cite{guillot2020dreem}. These datasets consist of 80 PSG recordings, each scored by five independent physicians from three different sleep centers following AASM guidelines (version 2.4) \cite{berry2017aasm}.\\

Table~\ref{tab:db_overview} provides an overview of the demographic and dataset statistics, with additional details available in Supplementary Notes: Datasets. \\

\begin{table}[ht]
\caption{\textbf{Overview of the demographic and dataset statistics}. Datasets used for in-domain (ID) training (i.e., training, validation and test set split) and out-of-domain (OOD) testing. Missing values are due to study design or anonymized data. Datasets directly available online are identified by \checkmark, while datasets that require approval from a Data Access Committee are marked by (\checkmark). BSWR is a private dataset.}
\label{tab:db_overview}
\begin{center}
\begin{tabular}{lllll}
\hline
\noalign{\vskip 1mm}
\textbf{ID Dataset} & \textbf{PSGs} & \textbf{Age (y)} & \textbf{\%(F/M)} & \textbf{BMI} \\
\hline
\noalign{\vskip 1mm}
\cite{zhang2018national, bakker2018gastric} ABC (\checkmark) & $132$ & $48.8 \pm 9.8$ & $43/57$ & $38.9 \pm 3.0$ \\
\hline
\noalign{\vskip 1mm}
\cite{zhang2018national, moore2014design} APOE (\checkmark) & $712$ & $45.7 \pm 13.6$ & $41/59$ & $27.2 \pm 6.5$ \\
\hline
\noalign{\vskip 1mm}
\cite{zhang2018national, quan2011association} APPLES (\checkmark) & $1094$ & $50.1 \pm 12.9$ & $37/63$ & $32.1 \pm 7.8$ \\
\hline
\noalign{\vskip 1mm}
\cite{zhang2018national, rosen2003prevalence} CCSHS (\checkmark) & $515$ & $17.7 \pm 0.4$ & $50/50$ & $25.1 \pm 5.9$ \\
\hline
\noalign{\vskip 1mm}
\cite{zhang2018national, redline1995familial} CFS (\checkmark) & $730$ & $41.7 \pm 20.0$ & $55/45$ & $32.4 \pm 9.5$ \\
\hline
\noalign{\vskip 1mm}
\cite{zhang2018national, marcus2013randomized, redline2011childhood} CHAT (\checkmark) & $1638$ & $6.6 \pm 1.4$ & $52/48$ & $19.0 \pm 4.9$ \\
\hline
\noalign{\vskip 1mm}
\cite{zhang2018national, rosen2012multisite} HOMEPAP (\checkmark) & $246$ & $46.5 \pm 11.9$ & $43/57$ & $37.2 \pm 8.9$ \\
\hline
\noalign{\vskip 1mm}
\cite{zhang2018national, chen2015racial} MESA (\checkmark) & $2056$ & $69.4 \pm 9.1$ & $54/46$ & $28.7 \pm 5.6$ \\
\hline
\noalign{\vskip 1mm}
\cite{zhang2018national, stephansen2018neural} MNC-CNC (\checkmark) & $78$ & $28.5 \pm 16.9$ & $49/51$ & $23.2 \pm 11.5$ \\
\hline
\noalign{\vskip 1mm}
\cite{zhang2018national, stephansen2018neural} MNC-DHC (\checkmark) & $83$ & $33.4 \pm 14.8$ & $50/50$ & $24.8 \pm 4.9$ \\
\hline
\noalign{\vskip 1mm}
\cite{zhang2018national, stephansen2018neural} MNC-SSC (\checkmark) & $767$ & $45.4 \pm 13.8$ & $41/59$ & $23.9 \pm 6.5$ \\
\hline
\noalign{\vskip 1mm}
\cite{zhang2018national, blackwell2011associations, osteoporotic2015relationships} MROS (\checkmark) & $3930$ & $76.4 \pm 5.5$ & $0/100$ & $27.2 \pm 3.9$ \\
\hline
\noalign{\vskip 1mm}
\cite{zhang2018national, dipietro2021fetal, dipietro2023fetal} MSP (\checkmark) & $105$ & $26.8 \pm 5.9$ & $100/0$ & $42.4 \pm 6.6$ \\
\hline
\noalign{\vskip 1mm}
\cite{zhang2018national, lee2022large} NCHSDB (\checkmark) & $3950$ & $8.8 \pm 5.9$ & $44/56$ & $22.7 \pm 9.9$ \\
\hline
\noalign{\vskip 1mm}
\cite{zhang2018national, quan1997sleep} SHHS (\checkmark) & $8444$ & $63.1 \pm 11.2$ & $52/48$ & $28.2 \pm 5.1$ \\
\hline
\noalign{\vskip 1mm}
\cite{zhang2018national, cummings1990appendicular, spira2008sleep} SOF (\checkmark) & $453$ & $82.8 \pm 3.1$ & $100/0$ & $27.7 \pm 4.6$ \\
\hline
\noalign{\vskip 1mm}
\cite{zhang2018national, young2009burden} WSC (\checkmark) & $2569$ & $56.4 \pm 8.1$ & $46/54$ & $31.7 \pm 7.1$ \\
\hline
\hline
\noalign{\vskip 1mm}
\textbf{OOD Dataset} & \textbf{PSGs} & \textbf{Age (y)} & \textbf{\%(F/M)} & \textbf{BMI} \\
\hline
\noalign{\vskip 1mm}
BSWR \cite{calle20180733} & $8410$ & $47.9 \pm 18.4$ & $34/66$ & $-$ \\
\hline
\noalign{\vskip 1mm}
\cite{perslev2021u} DCSM \checkmark & $255$ & $-$ & $-$ & $-$ \\
\hline
\noalign{\vskip 1mm}
\cite{guillot2020dreem} DOD-H \checkmark & $25$ & $35.3 \pm 7.5$ & $24/76$ & $23.8 \pm 3.4$ \\
\hline
\noalign{\vskip 1mm}
\cite{guillot2020dreem} DOD-O \checkmark & $55$ & $45.6 \pm 16.5$ & $36/64$ & $29.6 \pm 6.4$ \\
\hline
\noalign{\vskip 1mm}
\cite{goldberger2000physiobank, ghassemi2018you} PHYS \checkmark & $994$ & $55.2 \pm 14.3$ & $33/67$ & $-$ \\
\hline
\noalign{\vskip 1mm}
\cite{goldberger2000physiobank, kemp2000analysis} SEDF-SC \checkmark & $153$ & $58.8 \pm 22.0$ & $53/47$ & $-$ \\
\hline
\noalign{\vskip 1mm}
\cite{goldberger2000physiobank, kemp2000analysis} SEDF-ST \checkmark & $44$ & $40.2 \pm 17.7$ & $68/32$ & $-$ \\
\hline
\end{tabular}
\end{center}
\end{table}


Below, we summarize our work key experiments: \\

\begin{enumerate}[label=(\roman*),font=\itshape]

\item \textbf{Benchmarking, ensembling and generalization}. We train the sleep staging models on all available NSRR datasets in both single-channel and multi-channel configurations. In Supplementary Table~2 we summarize the data split sets. We then fairly evaluate their performance both on in-domain (ID) datasets and out-of-domain (OOD) datasets (e.g., BSWR).\footnote{
In-domain (ID) datasets refer to data that closely matches the distribution of the training data, while out-of-domain (OOD) datasets come from a different distribution, often representing new or unseen scenarios not encountered during training.
} We examine how aggregating predictions from multiple models enhances the robustness of sleep stage classification, especially in cases where individual models produce inconsistent results. 
Given the large number of models in SLEEPYLAND, we focus on the performance achievable using the SOMNUS configuration, which combines predictions from all available model architectures and channel configurations. The experiments provides further insights into the generalizability of the different models, and their consistency across sleep studies. In addition, we explicitly evaluate how exposure to broader, more diverse training data shapes generalization and OOD performance, highlighting the critical role of data diversity in achieving robust generalization.\\

\item \textbf{Model-bias on performance and clinical markers}. 
As part of our external validation, we apply a previously proposed framework \cite{bechny2025,bechny2024framework} to the large and clinical rich OOD BSWR dataset. This analysis measures how each sleep staging model’s performance and its predictions of clinical markers (such as REM latency) may be biased by demographic factors (like age and gender) and clinical variables (such as AHI and PLMI). Evaluations are based on majority-vote predictions from each sleep staging architecture and their SOMNUS ensemble (details on the experimented models below in \ref{subsec:Benchmarking}). The framework enables detection of systematic biases (e.g., age- or gender-related) and supports detailed assessment of how performance and prediction errors vary across subpopulations, facilitating comparison of models' generalizability in real-world clinical settings.\\

\item \textbf{Model-ensemble versus human-ensemble}. We evaluate the SOMNUS ensemble on two OOD multi-scorer datasets, DOD-H and DOD-O \cite{guillot2020dreem}, comparing  its ability to predict the consensus-derived hypnogram against, (1) hypnogram of individual human scorers, and (2) hypnogram predicted from a baseline model, i.e., SimpleSleepNet \cite{guillot2020dreem}, originally trained on the consensus itself. We then assess whether SOMNUS - and its constituent models - tend to align more closely with the consensus than with individual human scorers. This evaluation is conducted both in terms of discrete agreement (Cohen’s $\kappa$) and similarity between predicted and consensus-derived hypnodensity graphs \cite{stephansen2018neural}, quantified exploiting the Averaged Cosine Similarity (ACS) metric \cite{fiorillo2023multi}. In addition, we compare the degree of agreement among SLEEPYLAND models to that observed among human scorers using the soft-agreement metric proposed by Guillot et al. \cite{guillot2020dreem}. Finally, we characterize ensemble variability using the entropy of the soft-voting output and the pairwise divergence between individual model predictions, and investigate their utility as indicators of regions with elevated human scoring uncertainty.

\end{enumerate}

\subsection*{Benchmarking, ensembling and generalization}
\label{subsec:Benchmarking}

In SLEEPYLAND, we evaluate the performance of all model architectures trained in both single-channel and multi-channel configurations. For each model, we predict the full hypnogram of all the PSGs exploiting all the available channel derivations for that record. The ensembling models, i.e., SOMNUS models, are constructed by combining the predicted probabilities from different subsets of model architectures and channel configurations available in SLEEPYLAND, using a soft-voting approach. We consider ensembling across two dimensions: the network architecture and the channel configurations, either considering EEG and EOG derivations jointly, or separately, i.e., using only EEG or only EOG signals. When ensembling is restricted to a fixed architecture or a fixed channel configuration, the corresponding fixed dimension is indicated as a subscript of SOMNUS (e.g., SOMNUS$_\text{U-Sleep}$ refers to the ensemble of all models based on the U-Sleep architecture, varying only in channel configuration setup). When no subscript is specified, the ensemble incorporates variation across both network architectures and channel configurations. Given the large number of experiments across all the models, we do not show the results for all channel derivations. \\

\begin{table}[h]
\caption{\textbf{SOMNUS performance overview}. MF1 (\%) and Class-Wise F1 score (\%) on recording-level for soft unweighted ensemble of all models (majority-vote across channel derivations).}
\label{tab:results_somnus}
\begin{center}
\resizebox{\textwidth}{!}{
\begin{tabular}{lllllll}
\hline
\noalign{\vskip 1mm}
\textbf{ID Dataset} & \textbf{MF1} & \textbf{W} & \textbf{N1} & \textbf{N2} & \textbf{N3} & \textbf{REM} \\
\hline
\noalign{\vskip 1mm}
ABC      & $78.9 \pm 8.7$ & $89.5 \pm 4.5$ & $60.5 \pm 8.7$ & $86.6 \pm 6.8$ & $66.6 \pm 28.8$ & $89.5 \pm 20.1$ \\
\hline
\noalign{\vskip 1mm}
APOE     & $72.7 \pm 10.4$ & $87.7 \pm 7.6$ & $43.8 \pm 18.0$ & $85.5 \pm 9.9$ & $57.6 \pm 32.1$ & $86.5 \pm 13.6$ \\
\hline
\noalign{\vskip 1mm}
APPLES   & $74.0 \pm 9.3$ & $91.2 \pm 5.1$ & $50.6 \pm 15.7$ & $84.6 \pm 13.5$ & $36.1 \pm 32.5$ & $87.4 \pm 15.0$ \\
\hline
\noalign{\vskip 1mm}
CCSHS    & $87.2 \pm 4.9$ & $97.3 \pm 1.7$ & $64.9 \pm 14.1$ & $91.9 \pm 4.8$ & $88.1 \pm 9.4$ & $93.8 \pm 3.5$ \\
\hline
\noalign{\vskip 1mm}
CFS      & $81.7 \pm 8.0$ & $96.3 \pm 3.9$ & $53.9 \pm 16.2$ & $88.8 \pm 9.9$ & $78.1 \pm 22.3$ & $91.6 \pm 8.2$ \\
\hline
\noalign{\vskip 1mm}
CHAT     & $84.2 \pm 4.3$ & $96.0 \pm 3.1$ & $58.9 \pm 11.4$ & $85.8 \pm 7.8$ & $89.9 \pm 7.2$ & $90.1 \pm 5.7$ \\
\hline
\noalign{\vskip 1mm}
HOMEPAP  & $75.5 \pm 8.3$ & $90.8 \pm 7.8$ & $41.6 \pm 15.5$ & $82.7 \pm 9.1$ & $74.4 \pm 25.3$ & $90.4 \pm 11.1$ \\
\hline
\noalign{\vskip 1mm}
MESA     & $75.2 \pm 9.2$ & $95.3 \pm 6.1$ & $51.0 \pm 14.6$ & $85.2 \pm 11.1$ & $53.6 \pm 29.7$ & $90.6 \pm 7.1$ \\
\hline
\noalign{\vskip 1mm}
MNC-CNC  & $78.6 \pm 6.1$ & $80.5 \pm 15.0$ & $53.6 \pm 13.0$ & $83.8 \pm 10.2$ & $86.2 \pm 6.7$ & $89.1 \pm 4.0$ \\
\hline
\noalign{\vskip 1mm}
MNC-DHC  & $81.3 \pm 6.1$ & $97.8 \pm 1.4$ & $51.6 \pm 14.4$ & $85.1 \pm 8.3$ & $82.2 \pm 9.1$ & $89.7 \pm 5.4$ \\
\hline
\noalign{\vskip 1mm}
MNC-SSC  & $68.7 \pm 10.7$ & $80.7 \pm 12.6$ & $31.2 \pm 17.3$ & $86.3 \pm 6.6$ & $59.2 \pm 31.4$ & $85.5 \pm 16.8$ \\
\hline
\noalign{\vskip 1mm}
MROS     & $75.2 \pm 8.4$ & $95.8 \pm 3.5$ & $45.9 \pm 16.1$ & $88.0 \pm 5.5$ & $57.5 \pm 29.1$ & $88.0 \pm 14.3$ \\
\hline
\noalign{\vskip 1mm}
MSP      & $78.7 \pm 7.6$ & $92.8 \pm 5.6$ & $52.0 \pm 9.7$ & $88.8 \pm 5.3$ & $68.0 \pm 26.4$ & $89.4 \pm 11.5$ \\
\hline
\noalign{\vskip 1mm}
NCHSDB   & $75.9 \pm 7.5$ & $86.0 \pm 10.6$ & $33.5 \pm 16.5$ & $85.6 \pm 12.6$ & $89.8 \pm 9.4$ & $85.3 \pm 14.6$ \\
\hline
\noalign{\vskip 1mm}
SHHS     & $77.9 \pm 8.8$ & $93.5 \pm 7.5$ & $46.1 \pm 20.7$ & $87.3 \pm 7.6$ & $72.5 \pm 21.7$ & $90.4 \pm 8.9$ \\
\hline
\noalign{\vskip 1mm}
SOF      & $77.1 \pm 7.1$ & $95.1 \pm 4.7$ & $41.6 \pm 15.8$ & $85.6 \pm 8.0$ & $71.0 \pm 19.3$ & $92.8 \pm 7.1$ \\
\hline
\noalign{\vskip 1mm}
WSC      & $74.0 \pm 9.8$ & $89.5 \pm 10.6$ & $50.0 \pm 16.6$ & $90.5 \pm 5.6$ & $48.6 \pm 29.0$ & $88.4 \pm 13.9$ \\
\hline
\hline
\noalign{\vskip 1mm}
\textbf{OOD Dataset} & \textbf{MF1} & \textbf{W} & \textbf{N1} & \textbf{N2} & \textbf{N3} & \textbf{REM} \\
\hline
\noalign{\vskip 1mm}
BSWR     & $71.6 \pm 11.2$ & $83.4 \pm 14.0$ & $41.0 \pm 17.1$ & $81.1 \pm 12.4$ & $66.7 \pm 28.2$ & $86.8 \pm 16.9$ \\
\hline
\noalign{\vskip 1mm}
DCSM     & $80.3 \pm 8.3$ & $98.3 \pm 2.4$ & $50.0 \pm 15.1$ & $85.7 \pm 9.7$ & $78.6 \pm 19.6$ & $89.5 \pm 14.4$ \\
\hline
\noalign{\vskip 1mm}
DOD-H    & $83.6 \pm 6.6$ & $89.3 \pm 9.4$ & $57.2 \pm 17.1$ & $91.5 \pm 3.8$ & $85.7 \pm 16.3$ & $94.2 \pm 4.8$ \\
\hline
\noalign{\vskip 1mm}
DOD-O    & $79.2 \pm 7.9$ & $92.1 \pm 5.6$ & $52.6 \pm 13.5$ & $89.2 \pm 6.2$ & $70.3 \pm 27.6$ & $92.7 \pm 7.0$ \\
\hline
\noalign{\vskip 1mm}
PHYS     & $69.4 \pm 9.7$ & $74.4 \pm 16.0$ & $38.7 \pm 15.6$ & $83.7 \pm 10.2$ & $66.0 \pm 26.7$ & $85.5 \pm 16.6$ \\
\hline
\noalign{\vskip 1mm}
SEDF-SC  & $74.0 \pm 8.1$ & $98.4 \pm 1.2$ & $37.5 \pm 14.0$ & $83.3 \pm 8.2$ & $61.0 \pm 28.1$ & $88.0 \pm 8.5$ \\
\hline
\noalign{\vskip 1mm}
SEDF-ST  & $75.6 \pm 7.6$ & $79.3 \pm 10.8$ & $48.1 \pm 15.3$ & $87.2 \pm 5.7$ & $73.3 \pm 24.3$ & $90.2 \pm 7.9$ \\
\hline
\hline
\noalign{\vskip 1mm}
\textbf{Avg ID}   & $77.5 \pm 4.5$ & $91.5 \pm 5.4$ & $48.9 \pm 8.9$ & $86.6 \pm 2.4$ & $69.4 \pm 15.7$ & $89.3 \pm 2.3$ \\
\hline
\noalign{\vskip 1mm}
\textbf{Avg OOD}  & $76.2 \pm 5.1$ & $87.9 \pm 9.3$ & $46.4 \pm 7.5$ & $86.0 \pm 3.6$ & $71.7 \pm 8.4$ & $89.6 \pm 3.1$ \\
\hline
\end{tabular}}
\end{center}
\end{table}

In Table~\ref{tab:results_somnus} we report the performance achievable using the SOMNUS configuration, which combines predictions from all available network architectures and channel configurations. For each ID and OOD dataset we report the mean and standard deviation of F1 scores for each sleep stage and the macro F1 (MF1) score across all stages (computed as the unweighted average of the class-wise F1 scores). The results are reported on a recording-level considering the majority-vote across channel derivations.
Overall, the SOMNUS ensemble consistently maintains strong performance across all evaluation settings, achieving macro-F1 scores ranging from $68.7\%$ to $87.2\%$ on ID datasets and from $69.4\%$ to $83.6\%$ on OOD datasets. 
Although we believe metrics across individual recordings to be the more clinically appropriate, in Supplementary Table~3 we also presents SOMNUS results in the overall (dataset-level) metric format, enabling straightforward comparison with previous and future studies that evaluate performance across all epochs in a dataset rather than on a per-recording basis.\\

To contextualize these outcomes, in Table~\ref{tab:sota_comparison} we compare SOMNUS with state-of-the-art results from automatic sleep scoring methods \cite{stephansen2018neural, perslev2021u, vallat2021open, olesen2021automatic, phan2022sleeptransformer, phan2023seqsleepnet, thapa2025multimodal}. In these comparisons, we select the performance metric that each original study reported as their primary measure - i.e., dataset-level or recording-level accuracy, Cohen's Kappa, and macro-F1 - thereby ensuring a fair evaluation. SOMNUS consistently meets or surpasses these established methods, providing strong performance in cross-dataset (out-of-domain) testing scenarios, demonstrating its superior generalization capability to previously unseen data. Even in comparisons where the state-of-the-art models had been trained directly on a specific dataset (in-domain) while SOMNUS treated it as out-of-domain, SOMNUS often matches or closely approaches their performance. These gains are primarily driven by the large amount and diversity of training data, combined with the unified training framework used in SLEEPYLAND. In addition, the complementary predictions from heterogeneous network architectures and channel configurations contribute to more stable and robust performance across sleep stages.

\begin{table}[]
\caption{\textbf{Comparison of SOMNUS with state-of-the-art automatic sleep scoring methods}. \textbf{$^\star$} denotes dataset-wise (overall) evaluation, and \textbf{$^\bullet$} denotes recording-wise evaluation. The last column reports results for each state-of-the-art (SOTA) models or SOMNUS model, with in-domain status indicated by \cmark (dateset used during training) or \xmark (dataset not used during training). Best results for each row are in bold.
\textit{Note:} SOMNUS metrics are computed using the same aggregation (mean, median, etc.) as reported in the original publication of the compared SOTA model; inconsistencies with SOMNUS values in other tables of this manuscript are due to this adaptation.}
\label{tab:sota_comparison}
\begin{center}
\footnotesize
\begin{tabular}{llll}
\hline
\noalign{\vskip 1mm}
\textbf{Model} & \textbf{Dataset} & \textbf{Metric} & \textbf{SOTA vs SOMNUS} \\
\hline
\noalign{\vskip 1mm}
\multirow{6}{*}{YASA \cite{vallat2021open}} & CCSHS$^\star$ & Acc & 0.90 (\cmark) vs \textbf{0.93} (\cmark) \\
\hhline{|~---|}
& MESA$^\star$ & Acc & 0.84 (\cmark) vs \textbf{0.89} (\cmark) \\
\hhline{|~---|}
& \multirow{2}{*}{DOD-H$^\bullet$} & Acc  & 0.87 (\xmark) vs \textbf{0.91} (\xmark) \\
& & MF1  & 0.79 (\xmark) vs \textbf{0.84} (\xmark) \\
\hhline{|~---|}
& \multirow{2}{*}{DOD-O$^\bullet$} & Acc  & 0.84 (\xmark) vs \textbf{0.89} (\xmark) \\
& & MF1  & 0.74 (\xmark) vs \textbf{0.81} (\xmark) \\
\hline
\noalign{\vskip 1mm}
\multirow{18}{*}{U-Sleep \cite{perslev2021u}} & ABC$^\star$ & MF1 & 0.77 (\cmark) vs \textbf{0.83} (\cmark) \\
\hhline{|~---|}
& CCSHS$^\star$ & MF1 & 0.85 (\cmark) vs \textbf{0.88} (\cmark) \\
\hhline{|~---|}
& CFS$^\star$ & MF1 & 0.82 (\cmark) vs \textbf{0.83} (\cmark) \\
\hhline{|~---|}
& CHAT$^\star$ & MF1 & \textbf{0.85} (\cmark) vs \textbf{0.85} (\cmark) \\
\hhline{|~---|}
& DCSM$^\star$ & MF1 & 0.81 (\cmark) vs \textbf{0.82} (\xmark) \\
\hhline{|~---|}
& HOMEPAP$^\star$ & MF1 & \textbf{0.78} (\cmark) vs \textbf{0.78} (\cmark) \\
\hhline{|~---|}
& MESA$^\star$ & MF1 & \textbf{0.79} (\cmark) vs \textbf{0.79} (\cmark) \\
\hhline{|~---|}
& MROS$^\star$ & MF1 & 0.77 (\cmark) vs \textbf{0.79} (\cmark) \\
\hhline{|~---|}
& PHYS$^\star$ & MF1 & \textbf{0.79} (\cmark) vs 0.73 (\xmark) \\
\hhline{|~---|}
& SEDF-SC$^\star$ & MF1 & \textbf{0.79} (\cmark) vs 0.75 (\xmark) \\
\hhline{|~---|}
& SEDF-ST$^\star$ & MF1 & 0.76 (\cmark) vs \textbf{0.78} (\xmark) \\
\hhline{|~---|}
& SHHS$^\star$ & MF1 & \textbf{0.80} (\cmark) vs \textbf{0.80} (\cmark) \\
\hhline{|~---|}
& SOF$^\star$ & MF1 & 0.78 (\cmark) vs \textbf{0.79} (\cmark) \\
\hhline{|~---|}
& DOD-H$^\star$ & MF1 & 0.82 (\xmark) vs \textbf{0.86} (\xmark) \\
\hhline{|~---|}
& DOD-O$^\star$ & MF1 & 0.79 (\xmark) vs \textbf{0.82} (\xmark) \\
\hline
\noalign{\vskip 1mm}
\multirow{5}{*}{Stephansen et al. \cite{stephansen2018neural}} & WSC$^\bullet$ & Acc & \textbf{0.86} (\cmark) vs \textbf{0.86} (\cmark) \\
\hhline{|~---|}
& \multirow{2}{*}{DOD-H$^\bullet$} & Acc  & 0.86 (\xmark) vs \textbf{0.91} (\xmark) \\
& & MF1  & 0.79 (\xmark) vs \textbf{0.84} (\xmark) \\
\hhline{|~---|}
& \multirow{2}{*}{DOD-O$^\bullet$} & Acc  & 0.85 (\xmark) vs \textbf{0.89} (\xmark) \\
& & MF1 & 0.70 (\xmark) vs \textbf{0.81} (\xmark) \\
\hline
\noalign{\vskip 1mm}
\multirow{8}{*}{DeepResNet \cite{olesen2021automatic}} & \multirow{2}{*}{MROS$^\star$} & Acc & 0.87 (\cmark) vs \textbf{0.90} (\cmark) \\
& & Kappa & 0.79 (\cmark) vs \textbf{0.85} (\cmark) \\
\hhline{|~---|}
& \multirow{2}{*}{SHHS$^\star$} & Acc & 0.87 (\cmark) vs \textbf{0.88} (\cmark) \\
& & Kappa & 0.81 (\cmark) vs \textbf{0.83} (\cmark) \\
\hhline{|~---|}
& \multirow{2}{*}{WSC$^\star$} & Acc & \textbf{0.86} (\cmark) vs \textbf{0.86} (\cmark) \\
& & Kappa & 0.77 (\cmark) vs \textbf{0.79} (\cmark) \\
\hhline{|~---|}
& \multirow{2}{*}{SSC$^\star$} & Acc & 0.81 (\cmark) vs \textbf{0.82} (\cmark) \\
& & Kappa & 0.70 (\cmark) vs \textbf{0.72} (\cmark) \\
\hline
\noalign{\vskip 1mm}
\multirow{6}{*}{SleepTransformer \cite{phan2022sleeptransformer}} & \multirow{3}{*}{SHHS$^\star$} & Acc & \textbf{0.88} (\cmark) vs \textbf{0.88} (\cmark) \\
& & MF1 & \textbf{0.80} (\cmark) vs \textbf{0.80} (\cmark) \\
& & Kappa & \textbf{0.83} (\cmark) vs \textbf{0.83} (\cmark) \\
\hhline{|~---|}
& \multirow{3}{*}{SEDF-SC$^\star$} & Acc & 0.85 (\cmark) vs \textbf{0.92} (\xmark) \\
& & MF1 & \textbf{0.79} (\cmark) vs 0.75 (\xmark) \\
& & Kappa & 0.79 (\cmark) vs \textbf{0.84} (\xmark) \\
\hline
\noalign{\vskip 1mm}
\multirow{3}{*}{L-SeqSleepNet \cite{phan2023seqsleepnet}} & \multirow{3}{*}{SHHS$^\star$} & Acc & \textbf{0.88} (\cmark) vs \textbf{0.88} (\cmark) \\
& & MF1 & \textbf{0.80} (\cmark) vs \textbf{0.80} (\cmark) \\
& & Kappa & \textbf{0.83} (\cmark) vs \textbf{0.83} (\cmark) \\
\hline
\noalign{\vskip 1mm}
\multirow{3}{*}{SleepFM \cite{thapa2025multimodal}} & MESA$^\star$ & MF1 & 0.78 (\cmark) vs \textbf{0.79} (\cmark) \\
\hhline{|~---|}
& MROS$^\star$ & MF1 & 0.75 (\cmark) vs \textbf{0.79} (\cmark) \\
\hhline{|~---|}
& SHHS$^\star$ & MF1 & 0.78 (\cmark) vs \textbf{0.80} (\cmark) \\
\hline
\end{tabular}
\end{center}
\end{table}

Due to space constraints, the evaluations of the individual models, changing in network architecture and channel configurations, are instead reported in Supplementary Tables~4-19.
Overall, SOMNUS achieves higher median recording-wise macro F1 performance than individual models (i.e., specific architecture and channel configuration) in 94.9\% of cases. Among these, it is significantly better in 72.7\% of comparisons, and in the remaining cases, it is never significantly worse (Wilcoxon one-sided paired test, adjusted for multiple comparisons). This advantage is further illustrated in Supplementary Figure~9, which displays the distribution of median performance differences between SOMNUS and individual models across datasets.

\paragraph{Achieving and assessing generalization}
The amount and diversity of training data are central to enhancing the generalization capabilities of sleep staging models \cite{stephansen2018neural, olesen2021automatic, fiorilloautomated}. To confirm this within our unified framework, where other potential confounders such as data splits, preprocessing, training procedures, and data sampling strategy are held constant, we compare two versions of SOMNUS: one trained on all the NSRR datasets and another trained on a reduced subset, excluding APOE, APPLES, MNC, MSP, NCHSDB, and WSC\footnote{The exclusion is based purely on the temporal sequence of data availability—these datasets were added to NSRR after 2022 and treated as additional sources not included in earlier analyses.}. As shown in Supplementary Figure~10, the more comprehensively trained SOMNUS version achieves better generalization on 5 out of 7 OOD datasets and improved performance on newly included NSRR datasets, although with a slight drop in performance on the shared training datasets, likely reflecting their reduced representation within the extended training pool.
This pattern is further supported by results reported in Supplementary Table~20 (and clearly visible in the EEG-only subplots in Supplementary Figures~1-2), where an EEG-only SleepTransformer model (typically among the best performing individual models in our benchmarking) trained exclusively on SHHS fails entirely to generalize to other datasets. Its median macro-F1 score approaches that of the counterpart trained on the full NSRR data only for SHHS itself, with substantial and sometimes dramatic performance declines for other datasets, particularly in OOD evaluations.\\

Together, these insights reinforce a crucial point: \emph{achieving and assessing true generalization depends on training and evaluating models across increasingly larger and more diverse PSG datasets. Single-dataset training and evaluation, even with well-designed architectures, do not provide meaningful insights into true generalizability.}

\subsection*{Model-bias on performance and clinical markers}

\paragraph{BSWR: demographics and clinical variables} 

For a detailed assessment of robustness of individual sleep staging architectures, we exploit the Generalized Additive Models for Location, Scale, and Shape (GAMLSS) framework \cite{bechny2025,bechny2024framework}, applied on the BSWR clinical database. The evaluation is conducted on a subset of $6'633$ recordings with complete information for bias-inducing covariates $X$, i.e., age, gender, apnea-hypopnea index (AHI), periodic limb movement index (PLMI), and the relevant performance/markers outcomes.
This subset has a mean age of $47.9$ years ($SD=18.2$; range: $0–88$), with females representing $35.1\%$ of the sample. The mean AHI is $17.7$ events/hour ($SD=19.2$; range: $0–141$), and the mean PLMI is $12.1$ events/hour ($SD=21.9$; range: $0–204$). Based on two-sided t-test, females are younger (mean difference: $–4.1$; $95\%$ $CI: [–5.02, –3.16]$), have lower AHI (mean difference: $–8.6$; $95\%$ $CI: [–9.48, –7.73]$), and also have lower PLMI (mean difference: $–5.0$; $95\% CI: [–5.99, –4.01]$), in comparison to males. 
Using Pearson's coefficient, age shows moderate positive correlations with AHI ($r = 0.30$, $95\%$ $CI: [0.28, 0.33]$) and PLMI ($r = 0.30$, $95\%$ $CI: [0.28, 0.32]$), while AHI and PLMI are weakly correlated ($r = 0.056$, $95\%$ $CI: [0.032, 0.080]$).

\begin{figure}[h]
\centering
\includegraphics[width=\textwidth]{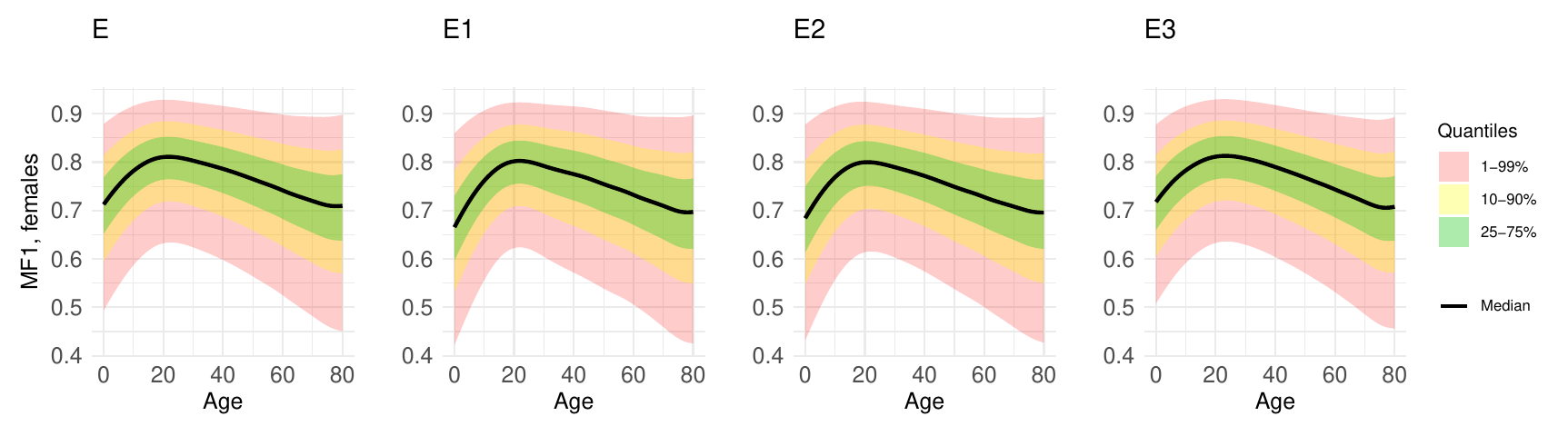}
\includegraphics[width=\textwidth]{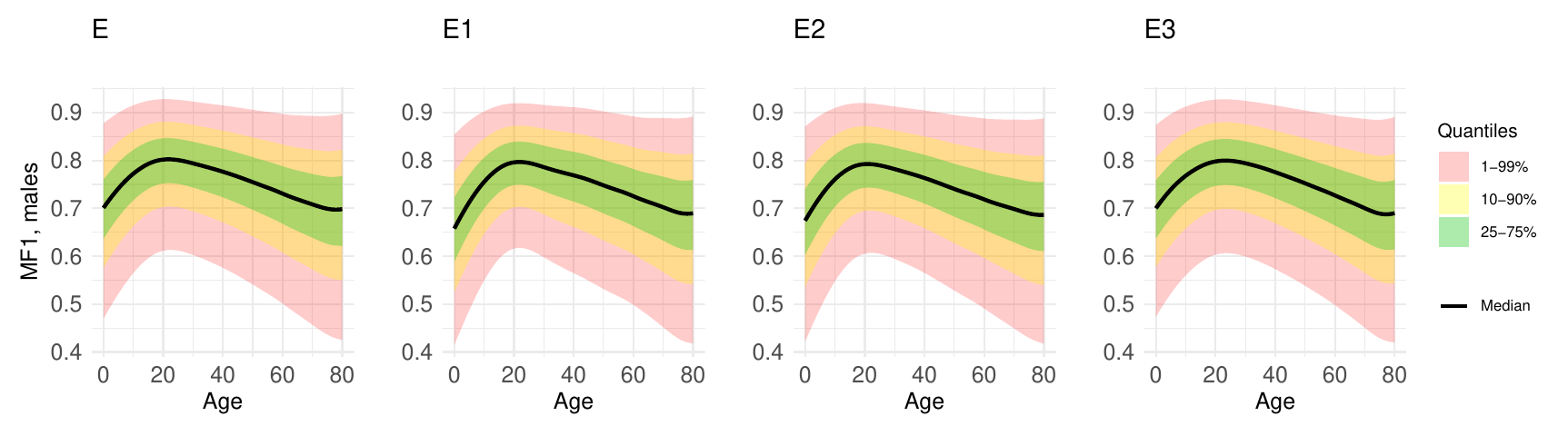}
\caption{\textbf{Age-conditioned expected distributions of MF1-scores quantiles for males and females}. Age-conditioned expected distributions of MF1-scores quantiles for males and females (under an optimistic scenario of AHI = PLMI = 0) across four sleep staging models, i.e., SOMNUS, SOMNUS$_\text{U‑Sleep}$, SOMNUS$_\text{DeepResNet}$, and SOMNUS$_\text{SleepTransformer}$.}
\label{fig:MF1_per_gender}
\end{figure}

\paragraph{Age, gender, AHI and PLMI bias}
In Table~\ref{tab:bias_performance} and in Table~\ref{tab:bias_markers} we report the bias/effects of the predictors $X$, i.e., age, gender, AHI, PLMI, on the both the average (location parameter, $\mu$) and variability (scale parameter, $\sigma$) of the outcomes. Specifically, we consider model performance metrics - MF1 and Class-Wise F1 scores - and clinically relevant sleep markers such as total sleep time (TST) and wake after sleep onset (WASO). The effect of each variable is modeled as a linear term in the distributional parameters (except for age), while the parameters for the \textit{inflated Beta distributions} ($\nu, \tau$) are kept constant. To clarify, the inflated Beta distribution allows us to appropriately model our outcomes, including the possibility of values exactly at the boundaries (0 or 1), by accounting for excess zeros or ones in the data; in this context, it helps capture rare cases of perfect or failed predictions. We identified which variables were most important for each outcome in the model. In the results tables, the presence of a numeric value indicates that the corresponding variable - i.e., gender, AHI, or PLMI - was retained in the final model for that parameter; blank cells indicate that the variable was excluded. Bold values flag the best outcomes - either higher performance, lower variability, or bias correction. If a predictor/covariate has no detectable effect on a metric, that setting is considered unbiased (and therefore most desirable). We do not report the age related predictor (modeled exploiting a cubic spline - see Figure~\ref{fig:MF1_per_gender}), mainly because a significant non-linear effect on $\mu$ and $\sigma$ is always identified for all the performance and clinical markers.\\

In this section, for simplicity, we will refer to the four architecture dependent ensemble sleep staging models - for which we quantify the bias - as follows: SOMNUS ($E$), SOMNUS$_\text{U‑Sleep}$ ($E1$), SOMNUS$_\text{DeepResNet}$ ($E2$), and SOMNUS$_\text{SleepTransformer}$ ($E3$).

\begin{table}[h]
\caption{\textbf{Bias on performance metrics}. Gender, AHI, and PLMI modeled as predictors of performance metrics for four sleep staging models, i.e., SOMNUS (\textbf{{E}}), SOMNUS$_\text{U‑Sleep}$ (\textbf{{E1}}), SOMNUS$_\text{DeepResNet}$ (\textbf{{E2}}), and SOMNUS$_\text{SleepTransformer}$ (\textbf{{E3}}) - exploiting the GAMLSS framework. Gender, AHI, and PLMI entered as linear terms. Bias estimated for the location ($\mu$, expectation) and scale ($\sigma$, variability) parameters, while the inflated‑Beta shape parameters ($\nu$, $\tau$) fixed at their intercepts.}
\label{tab:bias_performance}
\begin{center}
\resizebox{\textwidth}{!}{
\begin{tabular}{llllllllllllllllll}
\hline
\noalign{\vskip 1mm}
\multicolumn{2}{l}{\textbf{}} & \multicolumn{4}{c}{\cellcolor{gray!10}\textbf{(Intercept)}} & \multicolumn{4}{c}{\textbf{Gender}} & \multicolumn{4}{c}{\cellcolor{gray!10}\textbf{AHI}} & \multicolumn{4}{c}{\textbf{PLMI}} \\
&  &
\cellcolor{gray!10}\textbf{{E}} & \cellcolor{gray!10}\textbf{{E1}} & \cellcolor{gray!10}\textbf{{E2}} & \cellcolor{gray!10}\textbf{{E3}} &
\textbf{{E}} & \textbf{{E1}} & \textbf{{E2}} & \textbf{{E3}} &
\cellcolor{gray!10}\textbf{{E}} & \cellcolor{gray!10}\textbf{{E1}} & \cellcolor{gray!10}\textbf{{E2}} & \cellcolor{gray!10}\textbf{{E3}} &
\textbf{{E}} & \textbf{{E1}} & \textbf{{E2}} & \textbf{{E3}} \\
\hline
\noalign{\vskip 1mm}
\multirow{4}{*}{MF1} & $\mu$ & \cellcolor{gray!10}1.12 & \cellcolor{gray!10}1.06 & \cellcolor{gray!10}1.05 & \cellcolor{gray!10}\textbf{1.13}  & -0.06 & \textbf{-0.03} & -0.04 & -0.09 & \cellcolor{gray!10}-0.05 & \cellcolor{gray!10}-0.05 & \cellcolor{gray!10}-0.05 & \cellcolor{gray!10}-0.05 & -0.02 & -0.02 & -0.02 & -0.02 \\
& $\sigma$ & \cellcolor{gray!10}-1.51 & \cellcolor{gray!10}-1.46 & \cellcolor{gray!10}-1.47 & \cellcolor{gray!10}\textbf{-1.53}  & 0.06 &  &  & 0.08 & \cellcolor{gray!10} & \cellcolor{gray!10} & \cellcolor{gray!10} & \cellcolor{gray!10} & 0.01 & 0.01 &  & 0.02 \\
& $\nu$ & \cellcolor{gray!10}\textbf{-22.54} & \cellcolor{gray!10}\textbf{-22.54} & \cellcolor{gray!10}\textbf{ -22.54} & \cellcolor{gray!10}-21.54 &  &  &  &  & \cellcolor{gray!10} & \cellcolor{gray!10} & \cellcolor{gray!10} & \cellcolor{gray!10} &  &  &  &  \\
& $\tau$ & \cellcolor{gray!10}-22.63 & \cellcolor{gray!10}-22.63 & \cellcolor{gray!10}-22.63 & \cellcolor{gray!10}\textbf{-21.63} &  &  &  &  & \cellcolor{gray!10} & \cellcolor{gray!10} & \cellcolor{gray!10} & \cellcolor{gray!10} &  &  &  &  \\
\hline
\noalign{\vskip 1mm}
\multirow{4}{*}{F1\textsubscript{W}} & $\mu$ & \cellcolor{gray!10}\textbf{1.93} & \cellcolor{gray!10}\textbf{1.93} & \cellcolor{gray!10}1.90 & \cellcolor{gray!10}1.90  & -0.24 & \textbf{-0.23} & -0.24 & -0.24 & \cellcolor{gray!10}-0.07 & \cellcolor{gray!10}\textbf{-0.06} & \cellcolor{gray!10}-0.07 & \cellcolor{gray!10}-0.07 & 0.01 & 0.01 & 0.01 & 0.01 \\
& $\sigma$ & \cellcolor{gray!10}-0.97 & \cellcolor{gray!10}-0.98 & \cellcolor{gray!10}\textbf{-0.99} & \cellcolor{gray!10}-0.96  & 0.18 & 0.20 & 0.19 & \textbf{0.17} & \cellcolor{gray!10}0.02 & \cellcolor{gray!10}\textbf{0.01} & \cellcolor{gray!10}0.02 & \cellcolor{gray!10}0.03 &  &  &  &  \\
& $\nu$ & \cellcolor{gray!10}\textbf{-8.80} & \cellcolor{gray!10}-8.11 & \cellcolor{gray!10}\textbf{-8.80} & \cellcolor{gray!10}\textbf{-8.80} &  &  &  &  & \cellcolor{gray!10} & \cellcolor{gray!10} & \cellcolor{gray!10} & \cellcolor{gray!10} &  &  &  &  \\
& $\tau$ & \cellcolor{gray!10}-23.63 & \cellcolor{gray!10}-24.49 & \cellcolor{gray!10}-24.49 & \cellcolor{gray!10}\textbf{-8.80} &  &  &  &  & \cellcolor{gray!10} & \cellcolor{gray!10} & \cellcolor{gray!10} & \cellcolor{gray!10} &  &  &  &  \\
\hline
\noalign{\vskip 1mm}
\multirow{4}{*}{F1\textsubscript{N1}} & $\mu$ & \cellcolor{gray!10}-0.28 & \cellcolor{gray!10}\textbf{-0.22} & \cellcolor{gray!10}-0.26 & \cellcolor{gray!10}-0.39  &  & 0.03 & 0.03 &  & \cellcolor{gray!10}-0.03 & \cellcolor{gray!10}\textbf{-0.02} & \cellcolor{gray!10}\textbf{-0.02} & \cellcolor{gray!10}-0.04 & -0.02 & -0.02 & \textbf{-0.01} & -0.02 \\
& $\sigma$ & \cellcolor{gray!10}-0.75 & \cellcolor{gray!10}-0.79 & \cellcolor{gray!10}\textbf{-0.83} & \cellcolor{gray!10}-0.72  & \textbf{-0.05} & -0.03 & -0.04 & -0.04 & \cellcolor{gray!10} & \cellcolor{gray!10} & \cellcolor{gray!10} & \cellcolor{gray!10}0.02 &  &  &  &  \\
& $\nu$ & \cellcolor{gray!10}-4.82 & \cellcolor{gray!10}-5.10 & \cellcolor{gray!10}\textbf{-5.58} & \cellcolor{gray!10}-4.03 &  &  &  &  & \cellcolor{gray!10} & \cellcolor{gray!10} & \cellcolor{gray!10} & \cellcolor{gray!10} &  &  &  &  \\
& $\tau$ & \cellcolor{gray!10}-22.63 & \cellcolor{gray!10}-24.17 & \cellcolor{gray!10}\textbf{-20.63} & \cellcolor{gray!10}-23.62 &  &  &  &  & \cellcolor{gray!10} & \cellcolor{gray!10} & \cellcolor{gray!10} & \cellcolor{gray!10} &  &  &  &  \\
\hline
\noalign{\vskip 1mm}
\multirow{4}{*}{F1\textsubscript{N2}} & $\mu$ & \cellcolor{gray!10}\textbf{1.68} & \cellcolor{gray!10}1.59 & \cellcolor{gray!10}1.58 & \cellcolor{gray!10}1.67 &  & 0.04 & 0.04 &  & \cellcolor{gray!10}-0.09 & \cellcolor{gray!10}-0.09 & \cellcolor{gray!10}-0.09 & \cellcolor{gray!10}-0.09 & -0.03 & -0.03 & \textbf{-0.02} & -0.04 \\
& $\sigma$ & \cellcolor{gray!10}-1.35 & \cellcolor{gray!10}-1.31 & \cellcolor{gray!10}-1.32 & \cellcolor{gray!10}\textbf{-1.36}  &  &  &  &  & \cellcolor{gray!10}\textbf{0.05} & \cellcolor{gray!10}\textbf{0.05} & \cellcolor{gray!10}\textbf{0.05} & \cellcolor{gray!10}0.06 & 0.02 & \textbf{0.01} & \textbf{0.01} & 0.03 \\
& $\nu$ & \cellcolor{gray!10}\textbf{-24.47} & \cellcolor{gray!10}-23.54 & \cellcolor{gray!10}-24.14 & \cellcolor{gray!10}-24.14 &  &  &  &  & \cellcolor{gray!10} & \cellcolor{gray!10} & \cellcolor{gray!10} & \cellcolor{gray!10} &  &  &  &  \\
& $\tau$ & \cellcolor{gray!10}-24.49 & \cellcolor{gray!10}\textbf{-23.63} & \cellcolor{gray!10}-24.18 & \cellcolor{gray!10}-24.18 &  &  &  &  & \cellcolor{gray!10} & \cellcolor{gray!10} & \cellcolor{gray!10} & \cellcolor{gray!10} &  &  &  &  \\
\hline
\noalign{\vskip 1mm}
\multirow{4}{*}{F1\textsubscript{N3}} & $\mu$ & \cellcolor{gray!10}1.03 & \cellcolor{gray!10}0.78 & \cellcolor{gray!10}0.76 & \cellcolor{gray!10}\textbf{1.25} & -0.14 & \textbf{-0.05} & -0.09 & -0.20 & \cellcolor{gray!10}-0.03 & \cellcolor{gray!10}-0.03 & \cellcolor{gray!10}-0.03 & \cellcolor{gray!10}-0.03 & \textbf{-0.02} & -0.03 & -0.03 & -0.03 \\
& $\sigma$ & \cellcolor{gray!10}-0.34 & \cellcolor{gray!10}-0.21 & \cellcolor{gray!10}-0.22 & \cellcolor{gray!10}\textbf{-0.47}  & 0.10 &  &  & 0.15 & \cellcolor{gray!10}0.02 & \cellcolor{gray!10}0.02 & \cellcolor{gray!10}0.02 & \cellcolor{gray!10}\textbf{0.01} & \textbf{0.01} & \textbf{0.01} & \textbf{0.01} & 0.02 \\
& $\nu$ & \cellcolor{gray!10}-2.88 & \cellcolor{gray!10}-2.64 & \cellcolor{gray!10}-2.71 & \cellcolor{gray!10}\textbf{-3.10} &  &  &  &  & \cellcolor{gray!10} & \cellcolor{gray!10} & \cellcolor{gray!10} & \cellcolor{gray!10} &  &  &  &  \\
& $\tau$ & \cellcolor{gray!10}-8.75 & \cellcolor{gray!10}-7.63 & \cellcolor{gray!10}-24.13 & \cellcolor{gray!10}\textbf{-7.37} &  &  &  &  & \cellcolor{gray!10} & \cellcolor{gray!10} & \cellcolor{gray!10} & \cellcolor{gray!10} &  &  &  &  \\
\hline
\noalign{\vskip 1mm}
\multirow{4}{*}{F1\textsubscript{REM}} & $\mu$ & \cellcolor{gray!10}\textbf{2.09} & \cellcolor{gray!10}2.03 & \cellcolor{gray!10}2.04 & \cellcolor{gray!10}2.00 &  &  &  &  & \cellcolor{gray!10}-0.05 & \cellcolor{gray!10}\textbf{-0.04} & \cellcolor{gray!10}-0.05 & \cellcolor{gray!10}\textbf{-0.04} &  &  &  &  \\
& $\sigma$ & \cellcolor{gray!10}\textbf{-0.95} & \cellcolor{gray!10}-0.91 & \cellcolor{gray!10}-0.94 & \cellcolor{gray!10}-0.85 &  & 0.05 &  & \textbf{-0.04} & \cellcolor{gray!10}0.02 & \cellcolor{gray!10}\textbf{0.01} & \cellcolor{gray!10}0.02 & \cellcolor{gray!10}\textbf{0.01} & \textbf{0.04} & \textbf{0.04} & 0.05 & \textbf{0.04} \\
& $\nu$ & \cellcolor{gray!10}-4.06 & \cellcolor{gray!10}-4.16 & \cellcolor{gray!10}\textbf{-4.23} & \cellcolor{gray!10}-4.03 &  &  &  &  & \cellcolor{gray!10} & \cellcolor{gray!10} & \cellcolor{gray!10} & \cellcolor{gray!10} &  &  &  &  \\
& $\tau$ & \cellcolor{gray!10}-4.65 & \cellcolor{gray!10}-4.77 & \cellcolor{gray!10}-5.07 & \cellcolor{gray!10}\textbf{-4.60} &  &  &  &  & \cellcolor{gray!10} & \cellcolor{gray!10} & \cellcolor{gray!10} & \cellcolor{gray!10} &  &  &  &  \\
\hline
\end{tabular}}
\end{center}
\end{table}

\begin{table}[h]
\caption{\textbf{Bias on prediction-derived clinical markers}. Gender, AHI, and PLMI modeled as predictors of clinical markers for four sleep staging models, i.e., SOMNUS (\textbf{{E}}), SOMNUS$_\text{U‑Sleep}$ (\textbf{{E1}}), SOMNUS$_\text{DeepResNet}$ (\textbf{{E2}}), and SOMNUS$_\text{SleepTransformer}$ (\textbf{{E3}}) - exploiting the GAMLSS framework. Gender, AHI, and PLMI entered as linear terms. Bias estimated for the location ($\mu$, expectation) and scale ($\sigma$, variability) parameters.}
\label{tab:bias_markers}
\begin{center}
\resizebox{\textwidth}{!}{
\begin{tabular}{llllllllllllllllll}
\hline
\noalign{\vskip 1mm}
\multicolumn{2}{l}{\textbf{}} & \multicolumn{4}{c}{\cellcolor{gray!10}\textbf{(Intercept)}} & \multicolumn{4}{c}{\textbf{Gender}} & \multicolumn{4}{c}{\cellcolor{gray!10}\textbf{AHI}} & \multicolumn{4}{c}{\textbf{PLMI}} \\
 &  &
\cellcolor{gray!10}\textbf{{E}} & \cellcolor{gray!10}\textbf{{E1}} & \cellcolor{gray!10}\textbf{{E2}} & \cellcolor{gray!10}\textbf{{E3}} &
\textbf{{E}} & \textbf{{E1}} & \textbf{{E2}} & \textbf{{E3}} &
\cellcolor{gray!10}\textbf{{E}} & \cellcolor{gray!10}\textbf{{E1}} & \cellcolor{gray!10}\textbf{{E2}} & \cellcolor{gray!10}\textbf{{E3}} &
\textbf{{E}} & \textbf{{E1}} & \textbf{{E2}} & \textbf{{E3}} \\
\hline
\noalign{\vskip 1mm}
\multirow{2}{*}{TST} & $\mu$ & \cellcolor{gray!10}-9.05 & \cellcolor{gray!10}\textbf{-8.11} & \cellcolor{gray!10}-8.98 & \cellcolor{gray!10}-9.96  & -1.37 & \textbf{-0.99} & -1.35 & -1.84 & \cellcolor{gray!10}-1.95 & \cellcolor{gray!10}\textbf{-1.69} & \cellcolor{gray!10}-1.77 & \cellcolor{gray!10}-2.28 & -0.47 &  &  & -0.50 \\
 & $\sigma$ & \cellcolor{gray!10}\textbf{2.86} & \cellcolor{gray!10}2.88 & \cellcolor{gray!10}2.88 & \cellcolor{gray!10}2.88  & \textbf{0.08} & 0.10 & \textbf{0.08} & 0.10 & \cellcolor{gray!10}0.03 & \cellcolor{gray!10}\textbf{0.02} & \cellcolor{gray!10}0.04 & \cellcolor{gray!10}0.04 & 0.04 & 0.04 & \textbf{0.03} & 0.04 \\
\hline
\noalign{\vskip 1mm}
\multirow{2}{*}{WASO} & $\mu$ & \cellcolor{gray!10}5.70 & \cellcolor{gray!10}\textbf{5.52} & \cellcolor{gray!10}5.87 & \cellcolor{gray!10}6.23  & 1.91 & \textbf{1.73} & 1.94 & 2.22 & \cellcolor{gray!10}1.53 & \cellcolor{gray!10}\textbf{1.24} & \cellcolor{gray!10}1.42 & \cellcolor{gray!10}1.77 &  &  &  &  \\
 & $\sigma$ & \cellcolor{gray!10}2.78 & \cellcolor{gray!10}2.82 & \cellcolor{gray!10}2.80 & \cellcolor{gray!10}\textbf{2.77}  & \textbf{0.08} & 0.09 & 0.09 & 0.10 & \cellcolor{gray!10}0.06 & \cellcolor{gray!10}\textbf{0.05} & \cellcolor{gray!10}0.06 & \cellcolor{gray!10}0.07 & 0.04 & \textbf{0.03} & \textbf{0.03} & 0.04 \\
\hline
\noalign{\vskip 1mm}
\multirow{2}{*}{N1} & $\mu$ & \cellcolor{gray!10}-27.54 & \cellcolor{gray!10}\textbf{-25.77} & \cellcolor{gray!10}-25.81 & \cellcolor{gray!10}-28.76  & -1.70 & \textbf{-1.25} & -1.47 & -2.06 & \cellcolor{gray!10}-5.91 & \cellcolor{gray!10}\textbf{-5.47} & \cellcolor{gray!10}-5.56 & \cellcolor{gray!10}-6.30 & -1.53 & -1.40 & \textbf{-1.35} & -1.65 \\
 & $\sigma$ & \cellcolor{gray!10}3.03 & \cellcolor{gray!10}3.01 & \cellcolor{gray!10}\textbf{3.00} & \cellcolor{gray!10}3.07  &  &  &  &  & \cellcolor{gray!10}0.10 & \cellcolor{gray!10}0.10 & \cellcolor{gray!10}0.10 & \cellcolor{gray!10}0.10 & 0.05 & 0.05 & \textbf{0.04} & \textbf{0.04} \\
\hline
\noalign{\vskip 1mm}
\multirow{2}{*}{N2} & $\mu$ & \cellcolor{gray!10}36.55 & 4\cellcolor{gray!10}3.38 & \cellcolor{gray!10}43.47 & \cellcolor{gray!10}\textbf{22.68}  & -3.77 & \textbf{-5.37} & -4.34 & -1.99 & \cellcolor{gray!10}3.68 & \cellcolor{gray!10}3.10 & \cellcolor{gray!10}\textbf{2.99} & \cellcolor{gray!10}4.33 & 1.09 & \textbf{0.89} & 1.01 & 1.17 \\
 & $\sigma$ & \cellcolor{gray!10}3.39 & \cellcolor{gray!10}3.46 & \cellcolor{gray!10}3.45 & \cellcolor{gray!10}\textbf{3.36} &  &  & -0.03 &  & \cellcolor{gray!10}\textbf{0.05} & \cellcolor{gray!10}\textbf{0.05} & \cellcolor{gray!10}\textbf{0.05} & \cellcolor{gray!10}0.06 & 0.01 &  & 0.01 & 0.02 \\
\hline
\noalign{\vskip 1mm}
\multirow{2}{*}{N3} & $\mu$ & \cellcolor{gray!10}-22.69 & \cellcolor{gray!10}-28.76 & \cellcolor{gray!10}-30.13 & \cellcolor{gray!10}\textbf{-9.59}  & 5.54 & \textbf{7.09} & 6.37 & 3.27 & \cellcolor{gray!10}0.44 & \cellcolor{gray!10}0.72 & \cellcolor{gray!10}\textbf{0.79} & \cellcolor{gray!10} & 0.17 & 0.23 & 0.19 & \textbf{0.25} \\
 & $\sigma$ & \cellcolor{gray!10}\textbf{3.26} & \cellcolor{gray!10}3.34 & \cellcolor{gray!10}3.32 & \cellcolor{gray!10}3.29  & -0.05 & \textbf{-0.07} & \textbf{-0.07} & -0.03 & \cellcolor{gray!10}\textbf{-0.04} & \cellcolor{gray!10}\textbf{-0.04} & \cellcolor{gray!10}\textbf{-0.04} & \cellcolor{gray!10}-0.03 & 0.01 & 0.01 &  & 0.01 \\
\hline
\noalign{\vskip 1mm}
\multirow{2}{*}{REM} & $\mu$ & \cellcolor{gray!10}4.33 & \cellcolor{gray!10}\textbf{2.60} & \cellcolor{gray!10}3.21 & \cellcolor{gray!10}5.61  & -0.65 &  & \textbf{-0.88} & -0.73 & \cellcolor{gray!10}-0.19 & \cellcolor{gray!10} & \cellcolor{gray!10} & \cellcolor{gray!10}\textbf{-0.27} & \textbf{-0.37} & -0.32 & -0.29 & -0.42 \\
 & $\sigma$ & \cellcolor{gray!10}\textbf{2.42} & \cellcolor{gray!10}2.43 & \cellcolor{gray!10}2.39 & \cellcolor{gray!10}2.47  & -0.03 & -0.03 & \textbf{-0.04} & \textbf{-0.04} & \cellcolor{gray!10} & \cellcolor{gray!10} & \cellcolor{gray!10}0.01 & \cellcolor{gray!10}-0.01 & 0.02 & 0.02 & 0.02 & 0.02 \\
\hline
\noalign{\vskip 1mm}
\multirow{2}{*}{REML} & $\mu$ & \cellcolor{gray!10}1.80 & \cellcolor{gray!10}-1.01 & \cellcolor{gray!10}\textbf{-0.57} & \cellcolor{gray!10}0.78  &  & 1.74 &  &  & \cellcolor{gray!10}1.10 & \cellcolor{gray!10}0.50 & \cellcolor{gray!10}0.80 & \cellcolor{gray!10}0.71 & 0.35 &  &  & 0.93 \\
 & $\sigma$ & \cellcolor{gray!10}3.76 & \cellcolor{gray!10}3.88 & \cellcolor{gray!10}3.76 & \cellcolor{gray!10}\textbf{3.75}  &  & \textbf{-0.09} & 0.05 &  & \cellcolor{gray!10}\textbf{0.03} & \cellcolor{gray!10}0.04 & \cellcolor{gray!10}0.04 & \cellcolor{gray!10}0.04 & \textbf{0.02} & 0.03 & \textbf{0.02} & 0.03 \\
\hline
\noalign{\vskip 1mm}
\multirow{2}{*}{AwH} & $\mu$ & \cellcolor{gray!10}0.32 & \cellcolor{gray!10}\textbf{0.25} & \cellcolor{gray!10}0.34 & \cellcolor{gray!10}0.40  & 0.23 & \textbf{0.21} & 0.23 & 0.24 & \cellcolor{gray!10}0.08 & \cellcolor{gray!10}\textbf{0.06} & \cellcolor{gray!10}0.07 & \cellcolor{gray!10}0.10 & -0.02 & -0.02 & -0.02 & \textbf{-0.03} \\
 & $\sigma$ & \cellcolor{gray!10}\textbf{0.03} & \cellcolor{gray!10}0.05 & \cellcolor{gray!10}\textbf{0.03} & \cellcolor{gray!10}0.05  & 0.15 & 0.14 & \textbf{0.13} & 0.17 & \cellcolor{gray!10}0.12 & \cellcolor{gray!10}\textbf{0.11} & \cellcolor{gray!10}0.12 & \cellcolor{gray!10}0.12 & 0.01 & 0.01 & 0.01 &  \\
\hline
\noalign{\vskip 1mm}
\multirow{2}{*}{TrH} & $\mu$ & \cellcolor{gray!10}-4.93 & \cellcolor{gray!10}-4.65 & \cellcolor{gray!10}\textbf{-4.48} & \cellcolor{gray!10}-4.76  & \textbf{0.41} & 0.32 & \textbf{0.41} & 0.36 & \cellcolor{gray!10}-0.18 & \cellcolor{gray!10}\textbf{-0.06} & \cellcolor{gray!10}-0.14 & \cellcolor{gray!10}-0.10 & 0.09 & \textbf{0.13} & 0.07 & 0.01 \\
 & $\sigma$ & \cellcolor{gray!10}1.76 & \cellcolor{gray!10}1.82 & \cellcolor{gray!10}1.79 & \cellcolor{gray!10}\textbf{1.72}  &  &  &  & 0.03 & \cellcolor{gray!10}0.12 & \cellcolor{gray!10}0.12 & \cellcolor{gray!10}0.12 & \cellcolor{gray!10}0.12 & 0.03 & 0.03 & 0.03 & 0.03 \\
 \hline
\end{tabular}}
\end{center}
\end{table}

Performance of all sleep staging models at baseline (defined as a 50-year-old female with AHI = PLMI = 0) is comparable, with macro-F1 scores, ranging from $0.74$ to $0.76$, the latter also exhibiting the highest consistency (i.e., lowest variability). Probabilities of complete misclassification or perfect scoring are negligible across models, with log-inflation parameters near $–22$, and corresponding probabilities $<10^{-9}$. Performance and precision decrease with increasing AHI, PLMI, and male gender, with $E3$ showing the largest gender-related MF1 drop. Stage-wise F1-scores are highest for REM and wake ($\approx0.87–0.89$), followed by N2 ($\approx0.83–0.84$), N3 ($\approx0.68–0.78$), and N1 ($\approx0.40–0.45$), which also shows the most variability. $E3$ yields the best N3 performance and matches others on REM and N2. Gender especially affects wake and N1 variability, while AHI and PLMI amplify variability in N2, N3, and REM. Across age, MF1 and F1 for N1, N2, and N3 (in older adults) decline at both ends of the lifespan, with REM and N3 showing wide uncertainty bands (refer to Figure~\ref{fig:MF1_per_gender}, and Supplementary Figures~3-4).\\

Sleep parameters prediction (i.e., TST, WASO, and REM duration) by all models at baseline (defined as a 50-year-old female with AHI = PLMI = 0) shows limited bias, with absolute deviations $<10$ minutes. TST is slightly underestimated ($–10$ to $–8$ min), WASO overestimated ($+5$ to $+6$ min), i.e., likely linked to higher sensitivity of sleep staging models in detecting (micro)arousals, and REM modestly overestimated (+2 to +6 min), leading to consistent underestimation of sleep efficiency. Biases are more pronounced in stage durations: N1 and N2 are overestimated (N1: $+ 26–29$ min; N2: $+ 23–43$ min), while N3 is underestimated ($–30$ min in DeepResNet; $–10$ min in SleepTransformer). REM latency (REML) shows small, model-dependent fluctuations. All models underestimate stage transitions (TrH: $–4.5$ to $–5$ /h) and overestimate awakenings (AwH: $+0.25$ to $+0.40$ /h), suggesting oversmoothing and sensitivity to microarousals. Based on GAMLSS, $E1$ performs best for TST, WASO, N1, REM, and AwH; $E2$ for REML and TrH; and $E3$ for N2 and N3. $E$ seems to not minimize any baseline bias. \\

Increased AHI and PLMI generally worsen bias and variability but partially correct underestimation in N3 and REM due to their natural reduction in sleep-disordered individuals. Male gender amplifies most biases, but reduces N3 and REM errors, reflecting naturally lower amounts in older males. Age-related trends are non-linear and consistent across models (refer to Supplementary Figures~5-8). WASO is underestimated in children and overestimated in older adults. N1 is increasingly underestimated at age extremes, N2 is overestimated in middle age, N3 is underestimated in youth and REM is slightly overestimated in children. Overall, variability rises in children and elderly, especially for N3 and REM. As a result, TST is slightly overestimated in children and underestimated in adults, compensating for WASO biases. REM latency remains generally stable with increased variability at age extremes. Awakenings are overestimated and transitions underestimated, especially in middle-aged subjects, with variability sharply increasing in older adults. These trends likely reflect age-related changes in sleep depth and fragmentation. No model architecture consistently minimizes bias across markers.

\subsection*{Model-ensemble versus human-ensemble}

Evaluating automatic sleep staging algorithms against human experts requires careful consideration of the intrinsic variability between scorers. Disagreements among trained sleep technologists are well-documented \cite{rosenberg2013american}, particularly at transitions between sleep stages \cite{stephansen2018neural}, and any robust automatic method must demonstrate consistent performance in the face of such ambiguities. Notably, medical consensus consistently outperforms individual expert judgment \cite{barnett2019comparative}, indicating that reliable models should aim to replicate consensus-based scoring rather than emulate any single scorer’s decisions \cite{fiorillo2023multi}. To investigate this aspect in the context of SLEEPYLAND models, we leverage the multi-scored Dreem Open Datasets introduced by Guillot et al. \cite{guillot2020dreem}. These datasets include polysomnographic recordings from 25 healthy individuals (DOD-H) and 55 patients diagnosed with obstructive sleep apnea (DOD-O), each independently annotated by five experienced sleep technologists from different clinical centers.

\paragraph{Benchmarking against human experts and SimpleSleepNet}\label{sec:dod_benchmarking}

\begin{figure}
\centering
\includegraphics[width=1\linewidth]{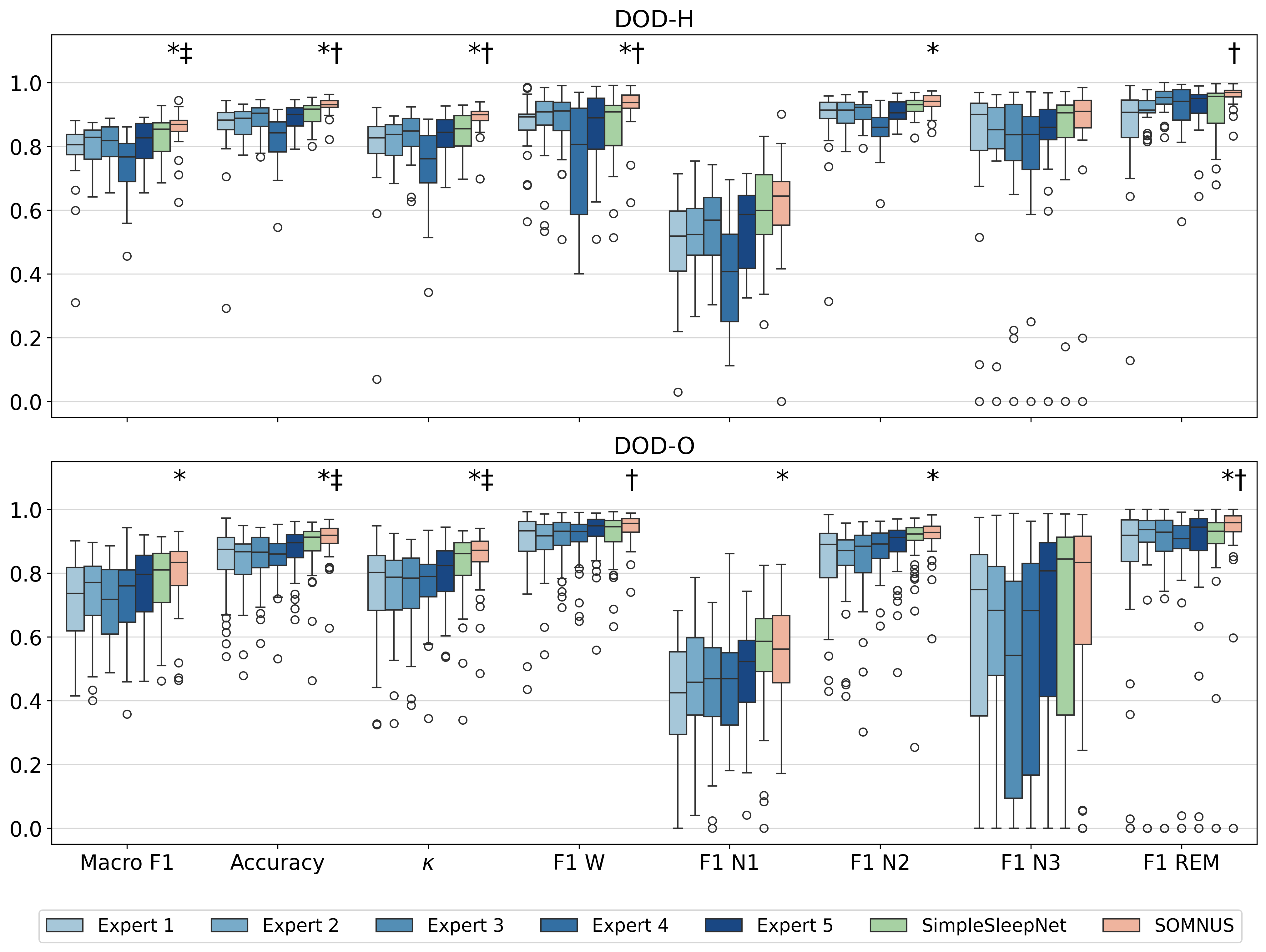}
\caption{\textbf{Distributions of recording-wise SOMNUS performance metrics for the OOD multi-scored dataset, relative to consensus labels}. Boxplots show the distributions of recording-wise performance metrics for the DOD-H (top) and DOD-O (bottom) datasets. Results are shown with respect to consensus annotations. Statistical significance markers above each boxplot indicate superior performance of SOMNUS: $\ast : p_{adj} < 0.05$ vs all experts, $\dag : p_{adj} < 0.05$ vs SimpleSleepNet, $\ddag : 0.05 \leq p_{adj} < 0.10$ vs SimpleSleepNet.}
\label{fig:dod_boxplots}
\end{figure}

We evaluate SOMNUS, our ensemble of high-performing models, against two benchmarks: (i) the individual human scorers, along with their average performance, and (ii) SimpleSleepNet, a model originally developed and trained specifically on DOD-H and DOD-O \cite{guillot2020dreem}. SOMNUS was never trained on DOD-H or DOD-O, making this a true out-of-domain evaluation that tests its robustness to inter-scorer variability. In Table~\ref{tab:dod_results} we report average performance across metrics, whilst in Figure~\ref{fig:dod_boxplots} we illustrate the distribution of recording-level performance metrics for both datasets. Evaluation protocols closely follow those defined in Guillot et al. \cite{guillot2020dreem}.

\begin{table}[h]
\caption{\textbf{Performance comparison with expert consensus on DOD-H and DOD-O datasets}. MF1 (\%) and Class-Wise F1 score (\%) on recording-level of the experts and of the models on DOD-H and DOD-O with respect to the soft-consensus among the experts. Recording-level mean ± standard deviation of MF1 (\%) and Class-Wise F1 score (\%), and Cohen’s $\kappa$ for individual human scorers, their average (Avg), SimpleSleepNet, and SOMNUS on DOD-H and DOD-O. All metrics are computed with respect to the expert-derived consensus.}
\label{tab:dod_results}
\centering
\resizebox{\textwidth}{!}{
\begin{tabular}{lllllllll}
\hline
\noalign{\vskip 1mm}
& \textbf{Scorer} & \textbf{MF1} & \textbf{$\kappa$} & \textbf{W} & \textbf{N1} & \textbf{N2} & \textbf{N3} & \textbf{REM} \\
\hline
\noalign{\vskip 1mm}
\multirow{8}{*}{\rotatebox{90}{DOD-H}} & Expert 1 & $77.8 \pm 11.6$ & $78.7 \pm 16.3$ & $86.2 \pm 9.7$ & $49.2 \pm 14.4$ & $87.8 \pm 12.6$ & $80.5 \pm 24.3$ & $85.0 \pm 17.2$ \\
& Expert 2 & $80.1 \pm 6.7$ & $81.5 \pm 6.5$ & $86.7 \pm 12.1$ & $52.2 \pm 11.4$ & $90.3 \pm 4.6$ & $79.9 \pm 22.9$ & $91.2 \pm 4.7$ \\
& Expert 3 (Best) & $80.8 \pm 6.1$ & $83.0 \pm 7.6$ & $87.7 \pm 10.6$ & $55.2 \pm 13.3$ & $90.5 \pm 4.4$ & $76.4 \pm 25.1$ & $94.4 \pm 4.2$ \\
& Expert 4 & $73.7 \pm 10.6$ & $73.3 \pm 12.9$ & $75.2 \pm 17.9$ & $40.3 \pm 16.5$ & $84.7 \pm 6.9$ & $76.8 \pm 21.8$ & $91.6 \pm 8.9$ \\
& Expert 5 & $80.4 \pm 7.5$ & $82.8 \pm 7.2$ & $85.6 \pm 12.0$ & $54.7 \pm 11.8$ & $91.1 \pm 3.7$ & $78.9 \pm 24.8$ & $91.8 \pm 8.0$ \\
& Experts (Avg) & $78.6 \pm 9.2$ & $79.9 \pm 11.4$ & $84.3 \pm 13.6$ & $50.3 \pm 14.7$ & $88.9 \pm 7.6$ & $78.5 \pm 23.9$ & $90.8 \pm 10.3$ \\
& SimpleSleepNet & $82.4 \pm 7.1$ & $84.6 \pm 6.5$ & $86.1 \pm 11.5$ & $59.8 \pm 14.4$ & $92.4 \pm 3.1$ & $82.5 \pm 23.0$ & $91.4 \pm 8.3$ \\
& \textbf{SOMNUS} & $\mathbf{85.2 \pm 6.6}$ & $\mathbf{88.9 \pm 4.7}$ & $\mathbf{91.8 \pm 7.7}$ & $\mathbf{60.9 \pm 16.9}$ & $\mathbf{93.5 \pm 3.2}$ & $\mathbf{84.1 \pm 22.7}$ & $\mathbf{95.8 \pm 3.4}$ \\
\hline
\hline
\noalign{\vskip 1mm}
\multirow{8}{*}{\rotatebox{90}{DOD-O}} & Expert 1 & $71.4 \pm 12.7$ & $74.9 \pm 15.3$ & $89.7 \pm 10.4$ & $41.0 \pm 16.7$ & $83.8 \pm 12.9$ & $60.0 \pm 32.1$ & $82.6 \pm 25.9$ \\
& Expert 2 & $73.5 \pm 11.9$ & $75.3 \pm 12.2$ & $89.5 \pm 8.3$ & $45.8 \pm 16.1$ & $84.0 \pm 11.4$ & $60.4 \pm 29.7$ & $87.9 \pm 21.7$ \\
& Expert 3 & $70.6 \pm 11.4$ & $75.4 \pm 12.4$ & $90.9 \pm 7.2$ & $44.5 \pm 16.5$ & $84.5 \pm 12.0$ & $46.6 \pm 33.3$ & $86.5 \pm 21.8$ \\
& Expert 4 & $72.2 \pm 12.1$ & $76.4 \pm 10.6$ & $91.0 \pm 7.3$ & $44.6 \pm 15.2$ & $87.4 \pm 6.7$ & $53.7 \pm 33.5$ & $84.3 \pm 24.0$ \\
& Expert 5 (Best) & $75.9 \pm 11.4$ & $80.5 \pm 9.5$ & $92.7 \pm 6.9$ & $48.5 \pm 15.3$ & $88.3 \pm 8.5$ & $63.7 \pm 33.9$ & $86.5 \pm 22.4$ \\
& Experts (Avg) & $72.7 \pm 12.1$ & $76.5 \pm 12.3$ & $90.8 \pm 8.2$ & $44.9 \pm 16.2$ & $85.6 \pm 10.7$ & $56.9 \pm 33.1$ & $85.6 \pm 23.3$ \\
& SimpleSleepNet & $77.6 \pm 11.4$ & $82.3 \pm 11.2$ & $91.7 \pm 7.4$ & $\mathbf{55.4 \pm 16.8}$ & $89.7 \pm 10.5$ & $64.8 \pm 36.0$ & $86.5 \pm 22.5$ \\
& \textbf{SOMNUS} & $\mathbf{80.2 \pm 10.1}$ & $\mathbf{85.4 \pm 8.0}$ & $\mathbf{94.3 \pm 4.3}$ & $55.3 \pm 13.7$ & $\mathbf{91.6 \pm 5.9}$ & $\mathbf{68.5 \pm 32.0}$ & $\mathbf{91.1 \pm 18.6}$ \\
\hline
\end{tabular}}
\end{table}

\textbf{DOD-H}. SOMNUS significantly outperforms all human experts across all global metrics: Macro F1 score ($p_{adj} = 0.002$, $r = 0.70$), Accuracy ($p_{adj} < 0.001$, $r = 0.82$), Cohen's $\kappa$ ($p_{adj} < 0.001$, $r = 0.82$), as well as for the staging of Wake ($p_{adj} = 0.008$, $r = 0.64$) and N2 ($p_{adj} = 0.020$, $r = 0.59$).
No significant differences are observed for N1, N3, or REM sleep, although small to moderate effect sizes consistently favor SOMNUS. 
When compared to SimpleSleepNet, SOMNUS achieves significantly higher Accuracy ($p_{adj} = 0.025$, $r = 0.57$), Cohen's $\kappa$ ($p_{adj} = 0.020$, $r=0.59$), and improved classification of Wake ($p_{adj} = 0.008$, $r=0.64$) and REM stages ($p_{adj} = 0.030$, $r=0.55$).
For Macro F1 score, SOMNUS is marginally superior ($p_{adj} = 0.096$, $r=0.44$). No significant differences are found for NREM stages, though median performances again trend in favor of SOMNUS.\\

\textbf{DOD-O}. SOMNUS consistently outperforms all individual human experts in terms of Macro F1 score ($p_{adj} < 0.001$, $r = 0.53$), Accuracy ($p_{adj} < 0.001$, $r = 0.54$), Cohen's $\kappa$ ($p_{adj} < 0.001$, $r = 0.52$), and in the determination of N1 ($p_{adj} = 0.001$, $r = 0.50$), N2 ($p_{adj} < 0.001$, $r = 0.55$), and REM sleep ($p_{adj} < 0.001$, $r = 0.51$). No significant differences are found for the Wake and N3 stages.
When compared to SimpleSleepNet, SOMNUS achieves slightly higher overall performance, although the differences are only weakly significant for both Accuracy ($p_{adj} = 0.059$, $r = 0.32$) and Cohen's $\kappa$ ($p_{adj} = 0.050$, $r = 0.34$).
As observed for the healthy cohort, SOMNUS clearly outperforms SimpleSleepNet in the classification of Wake ($p_{adj} = 0.026$, $r = 0.37$) and REM sleep ($p_{adj} < 0.001$, $r = 0.62$).

\paragraph{Alignment with human consensus}

To assess whether SOMNUS tends to replicate a consensus of human scorers rather than the style of individual experts, we evaluate the inter-scorer agreement using Cohen’s $\kappa$, comparing the model’s predictions both to individual annotations and to the consensus labels. Additionally, we quantify the similarity between SOMNUS predicted hypnodensity graphs and those computed by the soft-consensus using the ACS metric \cite{fiorillo2023multi}. In Table~\ref{tab:consensus_agreement} we report both the Cohen’s $\kappa$ agreement scores and the ACS values for DOD-H and DOD-O datasets. In both cases, the agreement with the consensus is significantly higher than with any individual scorer ($p_{adj} < 0.001$), suggesting that SOMNUS predictions are more closely aligned with collective scoring patterns. Moreover, the high mean ACS values observed, where a value of $1$ would imply perfect agreement, indicate that SOMNUS not only matches the consensus labels, but also closely captures the underlying probability distributions reflecting scorer uncertainty.

\begin{table}[h]
\caption{\textbf{Agreement metrics between SOMNUS and human annotations.}. Agreement of SOMNUS with individual human scorers, their consensus (Cohen’s $\kappa$, mean $\pm$ std), and soft-consensus, quantified using Averaged Cosine Similarity (ACS), on the DOD-O and DOD-H datasets.}
\resizebox{\textwidth}{!}{
\begin{tabular}{llllllll}
\hline
\noalign{\vskip 1mm}
Dataset & Expert 1 & Expert 2 & Expert 3 & Expert 4 & Expert 5 & Consensus & ACS \\
\hline
\noalign{\vskip 1mm}
DOD-H & 78.8 ± 17.0 & 81.7 ± 6.1 & 81.0 ± 10.2 & 74.7 ± 12.1 & 82.1 ± 8.7 & \textbf{88.9 ± 4.7} & 0.947 ± 0.021 \\
DOD-O & 74.9 ± 13.6 & 77.5 ± 10.3 & 73.0 ± 12.7 & 78.9 ± 9.7 & 79.6 ± 9.6 & \textbf{85.4 ± 8.0} & 0.939 ± 0.024 \\
\hline
\end{tabular}}
\label{tab:consensus_agreement}
\end{table}

We further investigate whether the observed alignment with consensus was specific to SOMNUS or extended to other model configurations. In Supplementary Table~21 we report Cohen’s $\kappa$ and ACS scores between various ensemble models, individual models, and the human consensus across DOD-H and DOD-O datasets. The consensus-oriented behavior is consistent across all models. Nonetheless, combining predictions across different architectures and channel configurations, as done in SOMNUS, yields additional gains, producing the highest levels of consensus alignment across both datasets.
We observe that the inter-model soft-agreement \cite{guillot2020dreem} within SLEEPYLAND exceeds the soft-agreement observed among human experts. 
On DOD-H, the average soft-agreement across human scorers is $0.894 \pm 0.028$, compared to $0.944 \pm 0.019$ among the models. Similarly, on DOD-O, models show slightly higher agreement ($0.902 \pm 0.081$) compared to human scorers ($0.885 \pm 0.014$).

\paragraph{Ensemble disagreement as an indicator of scoring ambiguity}

Building on the observation that SLEEPYLAND models align more closely with the consensus than with individual scorers, we further examine whether model-ensemble variability can serve as a proxy for highlighting regions of ambiguity in sleep staging. We characterize ensemble disagreement exploiting two complementary metrics: the \textit{entropy} of the soft-voting ensemble output; and the \textit{pairwise cosine distance} between predictions of individual models. The \textit{entropy} and the \textit{pairwise cosine distance} are computed against the ground-truth soft-consensus (i.e., level of agreement among human scorers at each epoch). This analysis explores to what extent variability within the ensemble corresponds to regions of human uncertainty, and whether inter-scorer disagreement can be predicted from model-derived indicators.\\

\begin{figure}
\centering
\includegraphics[width=1\linewidth]{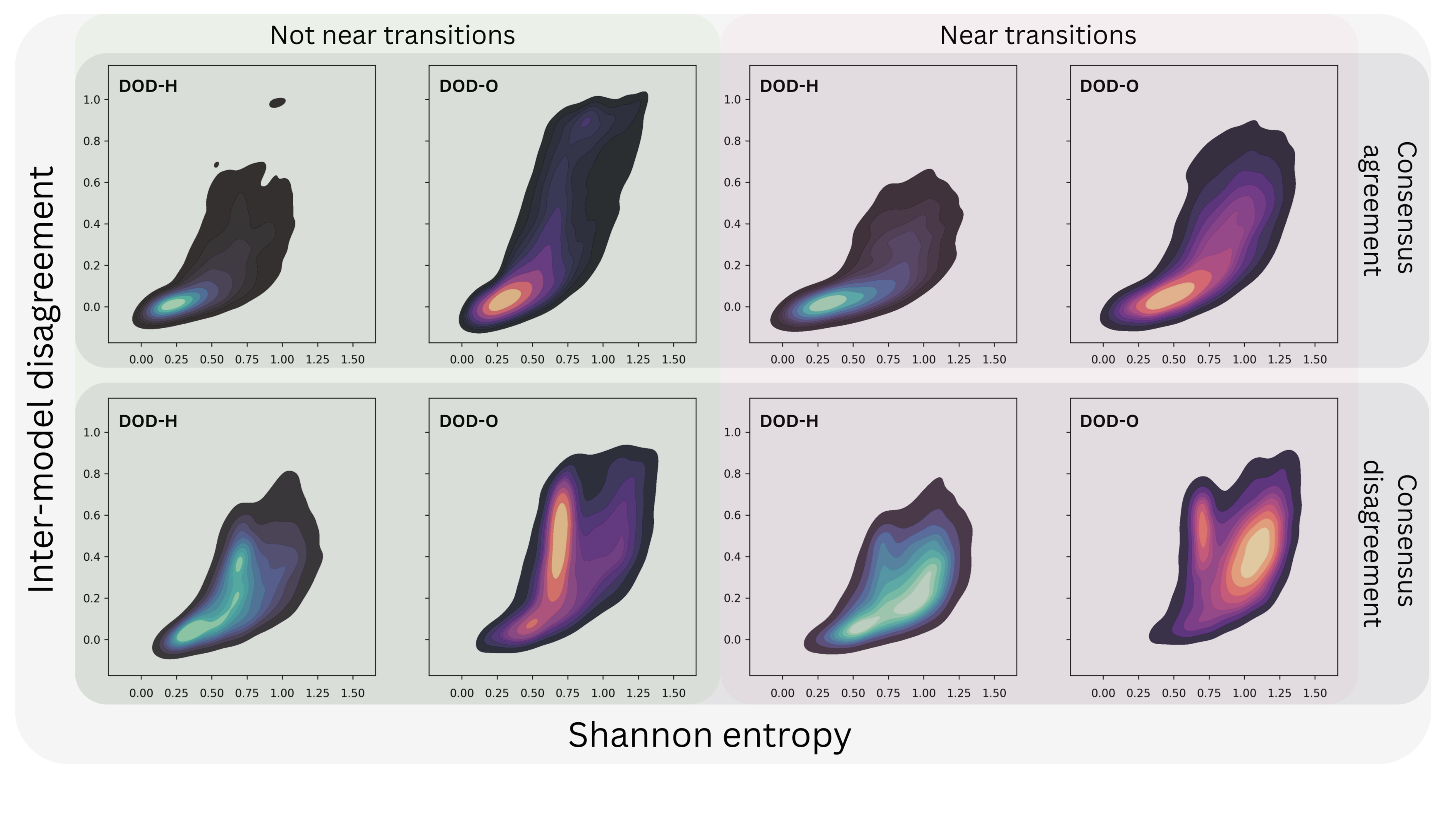}
\caption{\textbf{Joint distributions of ensemble entropy and inter-model disagreement}. Kernel density estimates illustrate the relationship between Shannon entropy and inter-model disagreement across sleep epochs, stratified by consensus agreement and proximity to sleep stage transitions. Columns represent proximity to transitions, and rows correspond to consensus conditions (full agreement vs. disagreement). Subplots display distributions separately for DOD-H (left) and DOD-O (right). Inter-model disagreement is represented by the first principal component derived from summary statistics (mean, standard deviation, maximum) of pairwise cosine distances among ensemble constituents.}
\label{fig:dod_disagreement}
\end{figure}

Figure~\ref{fig:dod_disagreement} shows the joint distribution of ensemble \textit{Shannon entropy} and \textit{inter-model disagreement}, stratified by ground-truth consensus agreement (full agreement vs. disagreement) and proximity to sleep stage transitions - defined as either within one minute of a transition (\textit{near}) or outside that window (\textit{not near}) - based on the consensus hypnogram.
We observe a substantial increase in both ensemble entropy and inter-model disagreement in epochs where the soft-consensus reveals disagreement among human scorers. This trend holds across both DOD-H and DOD-O datasets. In regions distant from sleep stage transitions, consensus disagreement is associated primarily with increased inter-model disagreement, particularly in DOD-O where inter-model soft-agreement is lower and performances are more variable. In contrast, entropy shows a more marked increase near transitions, reflecting the intrinsic ambiguity of these periods. Importantly, elevated inter-model disagreement persists even during transitions, suggesting that while entropy is especially sensitive to local uncertainty at boundaries, disagreement among components of the ensemble also captures broader ambiguities beyond stage shifts.\\

We tested whether ensemble entropy and inter-model disagreement could predict human consensus disagreement, finding that exploiting both led to improved performance. \textit{Simple logistic regression models using features from both sources of variability yielded the highest performance, with mean ROC AUCs of 0.823 on DOD-H and 0.828 on DOD-O, significantly outperforming counterparts using entropy alone (DOD-H: 0.814, DOD-O: 0.751) or features derived from pairwise cosine distance alone (DOD-H: 0.808, DOD-O: 0.816)}.

\section*{Discussions}

SLEEPYLAND is the first tool to fairly evaluate automatic sleep staging models by exploiting an unprecedented scale of harmonized, diverse datasets, including both in-domain and out-of-domain populations. SLEEPYLAND represents the first open-source platform providing sleep staging models pre-trained on $\thicksim$220'000 hours of harmonized PSG data from diverse populations covering 24 distinct clinical cohorts. Our framework is unique for two primary research-related reasons, and three additional strengths that are crucial in clinical practice. First, the system exploits a standardized evaluation protocol across different architectures, and signal configurations, eliminating dataset selection bias prevalent in prior studies. Second, we release the model weights - pre-trained on the hugest amount of data to date - openly, which lets researchers quickly fine-tune the network for specific, underrepresented groups (such as children with narcolepsy or older adults with mild sleep apnea) without the expense of starting training from scratch. Third, the system can run directly in a sleep lab, each clinic can test it on its own patients while keeping all data on-site and private. Privacy of sensitive sleep recordings is implicit, since data remain locally in the intranet of the clinics or research institute. Four, predictions of individual PSG recordings can be performed “on-the-fly” in the morning after the data collection, and transferred back to the local clinical or research tool for re-scoring within few minutes. Technicians with no advanced informatics or programming skills can perform all these steps. Five, the sleep staging (hypnogram), together with the epoch-wise scoring certainty score (hypnodensity graph plus, e.g., entropy-derived confidence) is provided to the physicians.\\

Overall, we show that deep learning models, when trained on such heterogeneous data, can generalize robustly across a wide spectrum of clinical populations and recording protocols. The models leverage any available EEG and/or EOG derivation, offering flexibility in deployment. Consistent with prior work, majority voting across available EEG and/or EOG derivations reliably outperforms individual channel configurations, and often surpasses the best-performing single-channel setup. Theoretically there is no need for an exhaustive channel selection. The SOMNUS ensemble, which combines predictions across both architectures and channel configurations via soft-voting, consistently exceeds previously reported state-of-the-art performance across a wide range of datasets and evaluation scenarios. SOMNUS demonstrate robust generalization capabilities in both in-domain and out-of-domain contexts. It also surpasses all previously reported state-of-the-art results on the out-of-domain multi-scored Dreem Open Datasets, consistently outperforming (without ever having seen the data before) both all individual human experts, and the originally proposed model \cite{guillot2020dreem} with the data in-training. Our ensemble model not only aligns more closely with the consensus, but also exhibits greater consistency, with significantly lower variability in per-recording performance. SOMNUS outperforms individual models across datasets in nearly 95\% of cases, and it is never significantly inferior, highlighting the value of ensemble approaches for robust, plug-and-play deployment in diverse settings. Our benchmarking shows that while certain individual models may outperform others on specific recordings/datasets, ensembling consistently output more reliable results across all the recordings/datasets. \textit{Unless there is clear evidence that a particular model is exceptionally well-suited to a specific dataset, ensemble methods remain the most dependable option for maintaining high and consistent performance across a wide range of evaluation scenarios}. Furthermore, exploiting ensembling approaches allows to complement uncertainty quantification methods based on softmax-derived metrics, by incorporating the available inter-model disagreement between ensemble components.\\

While we acknowledge that more sophisticated ensembling strategies, such as stacked ensembles or meta-learners, might yield even better performance, and that more precise uncertainty quantification might be achievable through dedicated auxiliary confidence networks \cite{bechny2024bridging}, our focus in this work is on promoting readily available ensembling techniques within SLEEPYLAND. The simplicity of the SOMNUS approach requires no additional training or architecture-specific tuning, making it particularly well-suited for an open-access platform. In this way users can easily define custom ensembles across different subsets of models.\\

\textit{SLEEPYLAND emphasizes that advances in automated sleep staging are driven more by expanding and diversifying the training data than by introducing new model architectures}. Exposure to such heterogeneous data and diverse scorer styles during training inherently promotes consensus-level generalization and mitigates scorer-specific biases, even without explicitly incorporating multi-annotated data. Models within SLEEPYLAND surpass previous methods that directly leverage multiple annotations during training \cite{guillot2020dreem, fiorillo2023multi}, even achieving superior alignment with soft-consensus. This underscores that large-scale, heterogeneous training alone can be sufficient to learn the generalizable representations necessary to match expert-level consensus. As more PSG data from varied populations and recording settings become publicly available, such as through continued expansion of NSRR, we anticipate further improvements in generalizability.
Moreover, while single-dataset experiments can provide useful proof-of-concept validations for model designs, robust claims of generalization demand comprehensive evaluation across multiple, diverse datasets. Moving forward, \textit{adopting multi-dataset training and rigorous in-domain and out-of-domain evaluation should become the standard practice to advance the field and ensure models are truly robust and clinically applicable}.\\

Considerable variability in macro-F1 scores persists across datasets, driven largely by specific sleep phases, e.g., N1 and N3 stage performance. Datasets with more fragmented or atypical sleep - such as those with narcolepsy (e.g., MNC-SSC data), or multiple comorbidities - remain particularly challenging. Our analysis reveals that automated sleep staging inherits biases rooted in both physiological variability and data limitations. While ensemble approaches like SOMNUS improve robustness, they fail to resolve fundamental disparities linked to demographic and clinical factors. Here four key insights emerge: (1) Performance degrades non-linearly at pediatric and elderly extremes, driven by AASM scoring rule differences between age groups and age-related EEG/EOG signal changes. For pediatric populations, mismatches arise from distinct sleep patterns criteria, while elderly declines reflect unmodeled comorbidities beyond AHI/PLMI. Our bias-findings contrast with our earlier experiments \cite{fiorillo2023u} - where the sub-optimal inclusion of age seemed to offer no benefit, and we therefore did not quantify, e.g., the underlying age-related bias. (2) Apparent gender differences in N3/REM accuracy largely derive from clinical confounders - males in the cohort were older and had higher AHI/PLMI. When controlling for these factors, gender effects diminished but persisted in N3, suggesting residual biological or scoring biases. (3) High AHI/PLMI increased N2/N3/REM variability due to sleep fragmentation. Models partially "corrected" N3 underestimation in severe OSA populations - a consequence of training data oversampling disordered sleep. This may explain how data imbalances can create misleading compensatory behaviors. (4) No model architecture consistently outperformed others in bias mitigation. Ensembling reduced performance gaps but preserved systemic biases, confirming that architectural innovations alone cannot resolve data-driven disparities.\\

Perhaps, the shortening of scoring periods from 30 seconds to, e.g., 5 seconds might better account for and reflect overall sleep physiology \cite{follin2025inter}. The AASM sleep staging rules have certain limitations to reflect sleep physiology, especially in persons with altered sleep as compared to healthy young adults, like children, elderly and sleep-disordered people. With automatic sleep staging, we have the opportunity to soften the strict historic rules for manual sleep staging, accounting better for actual sleep physiology. In parallel, a recent work has also highlighted the potential of self-supervised approaches for learning general-purpose sleep representations directly from large-scale raw data \cite{thapa2025multimodal}. While their performance on sleep staging remains slightly below that of supervised models like SOMNUS, their independence from scorer-derived labels offers two key advantages: the ability to sidestep scorer-specific biases, and enhanced generalization to downstream clinical prediction tasks. Together, these findings suggest that self-supervised learning represents an important frontier for future development in the field.\\

Looking ahead, several recommendations emerge:\\

\begin{enumerate}

\item Leverage \textbf{\textit{deep learning model-ensemble}}, exploiting their higher performance, model-disagreement derived metrics to flag ambiguous epochs for expert review.

\item Develop \textbf{\textit{bias-aware training paradigms}}, including explicit conditioning/weighting on demographic and clinical variables and/or underrepresented groups.

\item Expand \textbf{\textit{data diversity}}, prioritizing massively expanded datasets with intentional diversity sampling.

\item Develop \textbf{\textit{hybrid training strategies}}, combining large-scale pre-training on heterogeneous datasets with targeted fine-tuning for specific subgroups or scorer styles.

\item Integrate \textbf{\textit{human-in-the-loop workflows}}, allowing clinicians to review and correct automated outputs, with the potential for model personalization, while carefully managing the risk of overfitting to scorer-specific biases.

\item Establish \textbf{\textit{standardized reporting}}, i.e, noting prediction-derived clinical markers [e.g., SE] along with a flag [high AHI bias], ensuring transparency of model performance across subpopulations.\\

\end{enumerate}

As the field moves toward broader clinical adoption, a central technical question remains: will the combination of large-scale, diverse data, open-source tools, and adaptive bias-aware workflows be sufficient to achieve equitable and high-performance sleep staging in all patient populations? \\

So, is it then possible to start using these models in clinical routine? \\
Obviously, not yet. As a next step, tools like SLEEPYLAND will need to be further integrated within PSG software platforms commonly used in clinical and research institutions. Certification through programs such as the 2024 AASM Autoscoring Certification Pilot Program, as well as regulatory approval as a biomedical product under the EU AI Act, will be essential to transition automatic scoring tools from research into clinical practice. Continued collaboration between researchers, clinicians, and regulatory bodies will be crucial to ensure robust validation, user trust, and safe deployment in real-world healthcare environments.

\section*{Methods}\label{sec:methods}

SLEEPYLAND key features are the following: \emph{fair training and benchmarking}, i.e., the DL models integrated within the tool are trained and validated on the same set of harmonized data from the NSRR repository; \emph{modular architecture}, i.e, the modular design allows for flexibility and scalability through containerized services; \emph{local privacy} for user data, i.e., recognizing the sensitive nature of medical/sleep data, the tool ensures that all analysis (e.g., evaluating the models on their own data) is performed locally on the user’s device.

\subsection*{The architecture}

\begin{figure}[ht]
\centering
\includegraphics[width=\linewidth]{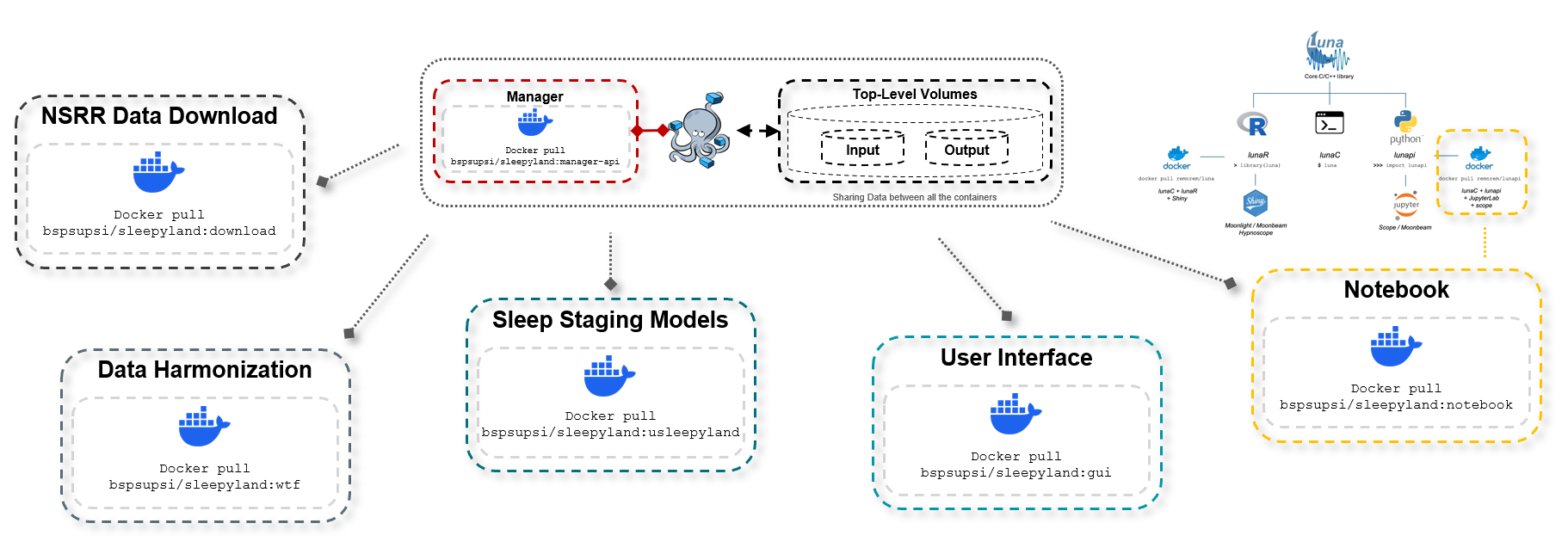}
\caption{\textbf{The SLEEPYLAND architecture}. Overview of the modular architecture composed of multiple containerized services. The \textit{manager-api} service coordinate communication across components. The \textit{gui} provides a web interface for data management and visualization. The \textit{nsrr-download} service automates data retrieval from NSRR repositories. The \textit{wild-to-fancy} service standardizes inputs by converting and harmonizing sleep recordings and annotations. Sleep staging is performed by the \textit{usleepyland} service, which integrates multiple ML/DL models. A dedicated Jupyter \textit{notebook} container supports advanced and customizable analyses.}
\label{fig:architecture}
\end{figure}

SLEEPYLAND has been developed using Docker Compose, with each key service running in a dedicated container. All Docker images and containers are stored in \href{https://hub.docker.com/r/bspsupsi/sleepyland}{Docker Hub}. By following the instructions in the available \href{https://github.com/biomedical-signal-processing}{GitHub} repository, the users can easily execute the tool on Linux, Mac, or Windows, automatically setting up all components in the background. The SLEEPYLAND architecture integrates multiple containerized services, each designed to perform a specific function within a modular and scalable framework (see Figure~\ref{fig:architecture}).\\

SLEEPYLAND is composed of multiple containerized services:\\

\begin{itemize}
    \item The \emph{\textbf{manager-api}} service: a central coordinating service that manages communication between all the implemented components. \\
    
    \item The \emph{\textbf{gui}} service, implemented using Flask, provides an intuitive web-based interface for data upload/download, task execution, and interactive visualization of results. \\
    
    \item To facilitate the use of datasets from the NSRR repository mentioned in Table~\ref{tab:db_overview}, the \emph{\textbf{nsrr-download}} service automates the retrieval of sleep data and annotations, allowing users to select specific cohorts and subjects. \\
    
    \item Data harmonization is handled by the \emph{\textbf{wild-to-fancy}} service, which converts sleep recordings from .edf to .h5, resamples data, and standardizes annotations across a range of formats. \\
    
    \item Sleep staging is performed by the \emph{\textbf{usleepyland}} service, which integrates state-of-the-art ML and DL models, including YASA \cite{vallat2021open}, U-Sleep, DeepResNet, and SleepTransformer. \\
    
    \item A Jupyter \emph{\textbf{notebook}} container is included for users who require customizable analyses or the ability to generate detailed reports.\\
\end{itemize}

SLEEPYLAND utilizes Docker volumes and mounted directories. Raw input data is stored in a shared input volume accessible to all services, while processed outputs are saved in an output volume. A dedicated directory is mounted for the Jupyter notebook container, allowing users to save and to generate custom analyses and reports. The choice of Docker Compose was driven by several key advantages. First, the use of containerization ensures isolation and modularity, allowing each service to operate independently. This modularity is particularly beneficial in contexts where components require different dependencies or need to be updated without affecting the overall system. Second, Docker Compose simplifies the deployment process by enabling the entire environment to be launched with a single command, reducing the complexity of managing multiple services. Third, Docker provides consistency across computing environments, eliminating potential discrepancies arising from differences in machine configurations or operating systems. It ensures cross-platform compatibility, making SLEEPYLAND accessible to a wide range of users.


\subsection*{Sleep staging algorithms}

The primary goal of SLEEPYLAND is to consolidate the top-performing sleep staging algorithms, trained and evaluated on a wide variety of datasets, including tens of thousands of PSG recordings from the publicly available NSRR data repository \cite{zhang2018national}, into a single open-source repository. To ensure a fair evaluation framework, a standardized approach is followed for data harmonization and the training procedures, including consistent splits into training, validation, and test sets across all models. In the following sections, we focus primarily on the core \emph{usleepyland} service, which integrates well known high-performing sleep staging models. Below we provide deeper insights into the deep learning architectures exploited, the training procedures, and the data management/sampling strategies.

\subsubsection*{Deep Learning models}\label{sec:dl_models}

SLEEPYLAND includes a diverse selection of deep learning models for sleep staging, selected for their reported performance, robustness, architectural diversity - including convolutional, recurrent, and attention-based frameworks - and their ability to process different input representations, such as raw signals or spectrograms.  Each model’s architecture and hyperparameters closely adhere to the original implementations, with only minimal modifications introduced to ensure compatibility with the proposed training procedure and to facilitate fair comparisons across architectures. These adjustments are strictly limited to aspects required for consistency, compatibility or numerical stability, without additional hyperparameter tuning. Moreover, while the original implementations were often tailored to specific signal channels or fixed derivations, SLEEPYLAND supports both single-channel and multi-channel configurations that can operate on any available derivation, enabling researchers to choose models that best align with their data and needs.\\

\begin{itemize}

    \item \emph{\textbf{U-Sleep}} \cite{perslev2021u}. The model is a purely feed-forward convolutional neural network developed for sleep staging, originally inspired by the U-Net architecture \cite{Ronneberger2015, perslev2019u}. The architecture comprises three main sub-modules. An \emph{encoder module}, composed of sequential blocks that alternate between convolution, ELU activation function, batch normalization, and max pooling. It is designed to extract abstract feature maps from the input signals, progressively reducing their temporal resolution to capture high-level features. The encoder is followed by a \emph{decoder module}, which mirrors the structure of the encoder and gradually reconstructs the temporal resolution of the encoded features, producing a high-frequency sleep stage representation. Skip connections between corresponding encoder and decoder layers help preserve crucial temporal details. Finally, a \emph{segment classifier} processes the high-frequency, intermediate representations generated by the decoder and maps them to a probability distribution across sleep stages, providing the final sleep stage predictions at a desired $\tau$ seconds interval. In all our experiments, U-Sleep is trained with hyperparameters in line with the original implementation, with the exception of an increased initial learning rate  $\eta = 10^{-5}$, adjusted from the original $\eta = 10^{-7}$. We set $\tau=30$ during both training and evaluation. \\
    
    \item \emph{\textbf{DeepResNet}} \cite{olesen2021automatic}. The model is based on an adaptation of a ResNet architecture \cite{He2016}, designed for processing one-dimensional, time-dependent PSG data. The network is organized into four modules. The \emph{mixing module} performs non-linear channel mixing, followed by a \emph{feature extraction module}, composed of a sequence of convolutional residual blocks, capturing key features at intra-epoch level from the signal. The \emph{temporal processing module} then leverages bidirectional GRUs to capture temporal dependencies, allowing the model to consider both preceding and following context for each time segment. Finally, a \emph{classification module} computes the probability distribution for each epoch across sleep stage classes. Notably, in model instances designed for single-channel input, the mixing module is omitted. All SLEEPYLAND versions of DeepResNet closely follow the original architecture, with key adjustments limited to increasing learning rate scheduler patience from $5$ to $50$ epochs and lowering the initial learning rate to $\eta=10^{-5}$. \\
    
    \item \emph{\textbf{SleepTransformer}} \cite{phan2022sleeptransformer}. The model is a fully transformer-based architecture \cite{Vaswani2017} tailored for sequence-to-sequence sleep staging. The architecture comprises two main attention-based components. The \emph{epoch transformer}, consisting of four encoder blocks, processes each 30-second sleep epoch by treating it as a sequence of spectral columns, capturing intra-epoch relationships. An additional attention mechanism at the end of this module produces a compact feature vector that encapsulates essential information from each epoch. Following this, the \emph{sequence transformer}, composed of four additional transformer encoder blocks, models inter-epoch dependencies by processing the sequence of feature vectors derived from consecutive epochs. A standard fully connected classification module finally provides probabilistic sleep stage predictions across the sequence. Only minimal modifications were made to adapt SleepTransformer to the proposed framework and improve numerical stability. In both the epoch and sequence transformers $d_{model}$ was adjusted to align with the number of channels processed in SLEEPYLAND; specifically, it was set to the product of the number of frequency bins and the number of input channels, allowing adaptability to various channel configurations. We also applied standard numerical stabilization strategies, including softmax output clipping, gradient clipping, and increasing the $\epsilon$ parameter of layer normalization to $10^{-6}$.
    
\end{itemize}

\paragraph{Training procedure and data sampling}\label{sec:training_procedures}

The training procedure and the data sampling from multiple datasets draws extensively from the original approach proposed in \cite{perslev2021u}. A \textit{training "epoch"} is defined differently from the conventional deep learning usage; in this context, it refers to the model processing $10^{6}$ sleep segments. Training is conducted for a maximum of $10000$ epochs, with early stopping triggered by a patience of $200$ epochs based on the mean validation F1 score across sleep stage classes. In practice, early stopping consistently occurred before reaching the maximum epoch limit. A training batch size of $64$ is maintained across all models, with each batch generated as follows: \emph{dataset sampling}, a training dataset is sampled based on a $50$/$50$ probability split between discrete uniform sampling and size-based selection; \emph{subject/recording and channel sampling}, from the chosen dataset, a PSG record is randomly selected, with channel(s) picked uniformly according to availability and model specifications; \emph{segment sampling}, a sleep stage class is then sampled, and a corresponding sleep segment/epoch is chosen randomly within the PSG record. Finally, a segment of the desired length is created to include the selected epoch in a random position. The segment length is fixed at 35 sleep epochs ($17.5$ minutes of PSG data) for all models. When the model requires time-frequency images instead of raw signals, spectrograms are generated "on-the-fly" following the procedure described in Phan et al. \cite{phan2022sleeptransformer}. Each channel undergoes robust scaling at the individual instance level within the batch. Data augmentation is applied for both raw and spectrogram inputs, with a $0.1$ probability of replacing parts of sequences with random noise or dropping channels, except for single-channel models where channel dropout is omitted. All models were trained to minimize cross-entropy loss between predicted sleep stages and ground-truth labels. Model definition and training were carried out within a unified TensorFlow 2 framework, with all training conducted on a single NVIDIA L40S GPU. For more details refer to the \emph{\textbf{usleepyland}} service repository \url{https://github.com/biomedical-signal-processing/uSLEEPYLAND}.

\paragraph{Inference procedure and evaluation}

Test-time inference in SLEEPYLAND partially follows the approach outlined in \cite{perslev2021u}, with adaptations to accommodate architectural differences across the available models. All models predict sleep stages while excluding sleep epochs labeled with annotations unrelated to sleep stages or left unscored by human annotators. Predictions can be generated using all available EEG and/or EOG channel combinations and a majority voting scheme is applied to fuse predictions across channels. Evaluation is conducted comprehensively, assessing both overall performance and class-specific metrics. Global performance is measured using Accuracy, Macro F1, and Cohen’s Kappa \cite{Mchugh2012}, while per-class F1 scores are computed for each of the five sleep stages. Due to architectural differences, inference procedures vary across models. U-Sleep, being fully convolutional, allows for one-shot evaluation, processing an entire PSG in a single forward pass, as in the original implementation. In contrast, DeepResNet, which includes recurrent components, and SleepTransformer, which relies on positional encoding, do not support this approach. Empirical results indicate that evaluating these models on extended sequences in a single forward pass leads to consistent performance degradation. Therefore, both DeepResNet and SleepTransformer are evaluated on fixed-length segments of 35 sleep epochs, mirroring their training setup. These segments are processed as independent instances within a batch, and the final sleep stage predictions are obtained by concatenating the outputs across segments. While overlapping window evaluation could theoretically be applied to refine predictions, the associated computational cost significantly outweighs the minor performance gains observed, and is therefore not implemented in SLEEPYLAND.

\subsubsection*{Model-ensembling}
\label{sec:model_ensembling}

To combine predictions from multiple automatic sleep staging algorithms within SLEEPYLAND, we adopt a soft-voting ensemble strategy. In this approach, each model $m \in \{1, \ldots , M\}$ produces a probability distribution $\hat{y}_m^{t} \in \mathbb{R}^5$ over the five sleep stages for each sleep epoch $t$. The ensemble output is computed as the element-wise average across all constituent model predictions:
\begin{equation}
    \hat{y}_E^{t} = \frac{1}{M} \sum_{m=1}^M \hat{y}_m^{t}
\end{equation}
We refer to such ensembles as SOMNUS models, constructed by aggregating predictions across different subsets of network architectures and channel configurations available in SLEEPYLAND. These configurations may be restricted to a fixed model architecture or signal modality, or may span all available models in a heterogeneous ensemble.
We choose soft-voting as the default ensembling strategy in SLEEPYLAND due to its flexibility and compatibility with the toolbox’s modular design. Provided that a model produces probabilistic output at the sleep-epoch level, it can be seamlessly integrated into a soft-voting ensemble.



\subsection*{Bias on performance and clinical markers}
\label{sec:bias_performance}

To evaluate robustness of different sleep staging approaches, we adopted the GAMLSS framework \cite{bechny2025, bechny2024framework, stasinopoulos2017flexible}, which models the distribution of performance metrics and prediction-derived clinical markers as a function of external covariates $X$. This approach captures systematic shifts in algorithmic behavior - referred to as \textit{biases} - and extends standard validation techniques (e.g., correlations with clinical variables, Bland–Altman plots). By enabling the estimation of arbitrary bias and performance quantiles conditional on subjects' characteristics $X$, it provides detailed insights essential for the clinical adoption of sleep-scoring algorithms. Recording-level performance metrics, including the macro F1-score (MF1) and stage-specific F1-scores (F1\textsubscript{W}, F1\textsubscript{N1}, F1\textsubscript{N2}, F1\textsubscript{N3}, F1\textsubscript{REM}), bounded in $[0,1]$, were modeled using a zeros-and-ones-inflated Beta distribution \cite{rigby2019distributions}. The location ($\mu$) and scale ($\sigma$) distributional parameters were modeled with a logit link, and the zero- and one-inflation parameters ($\nu$, $\tau$) - capturing completely incorrect (0\%) and perfect (100\%) sleep-scoring predictions - with a log link. Due to the limited number of cases with perfect or completely incorrect hypnogram predictions (e.g., $\leq4$ cases for MF1 across all models), covariate effects (cubic spline for age; linear terms for AHI, PLMI, and gender) were modeled only for the location ($\mu$; expectation) and scale ($\sigma$; variability) distributional parameters. The parameters ($\nu, \tau$) of the inflated Beta distributions were treated as constants using intercept only. Prediction-derived -clinical markers were quantified by modeling the differences $\hat{y} - y_{\text{ref.}}$ between predicted and (expert-scored) reference hypnogram-derived sleep markers using a generalized Normal distribution, with $\mu$ and $\sigma$ modeled by identity and log links, respectively. The markers considered included total sleep time (TST), wake after sleep onset (WASO), stage-specific sleep durations (N1, N2, N3, REM) [minutes], and REM latency (REML) [minutes], and the rates of awakenings (AwH) and transitions (TrH) per hour [N/hour]. Potential \textit{bias-inducing covariates} ($X$) included age (in years above 50), gender (binary, female as reference), apnea-hypopnea index (AHI, divided by 10), and periodic limb movement index (PLMI, divided by 10). The effect of age was modeled using a penalized cubic spline to capture known non-linear associations \cite{bechny2025}, while AHI and PLMI were modeled linearly, assuming proportional effects with event frequency. Gender was included as a binary covariate with female as the baseline. The optimal set of predictors for each distributional parameter was determined using forward stepwise regression based on the generalized Akaike information criterion (GAIC) \cite{stasinopoulos2017flexible}.\\

In Supplementary Notes: Calculation of expected values from GAMLSS, we report concrete examples of converting the table’s estimates into expected values. For derivation of conditional variance and quantiles, see \cite{stasinopoulos2017flexible}.

\subsection*{Evaluation Protocol for Multi-Annotated Datasets}
\label{sec:methods_consensus}
To investigate the well known challenge of inter-scorer variability in sleep staging \cite{rosenberg2013american} and evaluate the robustness of SLEEPYLAND models with respect to it, we adopt the multi-annotator evaluation framework and the Dreem Open Datasets introduced by Guillot et al. \cite{guillot2020dreem}, allowing for comparison of automatic models against annotations from five expert scorers. 

To assess statistical significance of median performance differences and consistency, quantified as the absolute deviation from the median Cohen’s $\kappa$, across recordings, we conduct one-sided Wilcoxon signed-rank tests with  Bonferroni-Holm correction for multiple comparisons. Effect size are reported as $r=\frac{Z}{\sqrt{N}}$. Comparisons against human experts are made relative to the best-performing individual scorer for each metric, providing the most challenging baseline. 

Given $S$ human scorers, let $y_s^t \in \{0, 1, 2, 3, 4\}$ denote the label assigned by scorer $s$ to epoch at time $t \in \{1, \ldots, T\}$, corresponding respectively to the five sleep stages Wake, N1, N2, N3, and REM. Let $\hat{y}_s^t \in \{0,1\}^{5}$ be the one-hot encoding of $y_s^t$. The probabilistic consensus for scorer $s$ at epoch $t$, denoted $\hat{z}_s^t \in [0,1]^5$, is defined as:
\begin{equation}
    \hat{z}_s^t = \frac{\sum\limits_{i=1; \ i \neq s}^S \hat{y}_i^t}{\max \left(\sum\limits_{i=1;\ i \neq s}^S \hat{y}_i^t\right)} 
\end{equation}
This assigns each sleep stage a score proportional to the number of other scorers who selected it, normalized such that the most frequent stage(s) receive a value of 1. In the case of ties, multiple stages may share the maximum value. 
The \emph{soft-agreement} $\in [0,1]$ of scorer $s$ over a recording is then computed as: 
\begin{equation}
    \text{Soft-Agreement}_s = \frac{1}{T} \sum_{t=1}^T \hat{z}_s^t[y_s^t]
\end{equation}
This metric quantifies the extent to which a scorer's annotations align with the collective judgment of the other scorers, weighted by the level of inter-scorer agreement at each timepoint. A value of 1 indicates perfect alignment with the majority at all epochs, while 0 indicates complete disagreement. Soft-agreement is computed for all human scorers across all recordings in the multi-scored datasets. The same formulation is also used to quantify agreement among SLEEPYLAND models, treating each model as an independent scorer. 
The \emph{consensus hypnogram} for a recording is obtained via majority voting across scorers at each epoch, with ties resolved using the decision of the most reliable scorer, defined as the scorer with the highest soft-agreement for that recording. For human annotators, the consensus hypnogram is computed excluding the scorer being evaluated and is derived from the remaining four scorers. For automated methods, the consensus is computed from the set $\mathfrak{C}$ consisting of the four most reliable human scorers across the dataset.

In addition to the discrete consensus hypnogram, we compute a \emph{soft-consensus} vector $\hat{y}_{\mathfrak{C}}^t$ at each epoch $t$, as introduced in \cite{fiorillo2023multi}, defined as: 

\begin{equation}
\hat{y}_{\mathfrak{C}}^t = \frac{1}{\lvert\mathfrak{C}\rvert} \sum_{s \in \mathfrak{C}} \hat{y}_s^t 
\end{equation}

This represents the empirical distribution of the annotations provided by scorers in $\mathfrak{C}$.

The \emph{hypnodensity graph} is a visualization tool that represents the probability distribution over sleep stages for each epoch across time, providing a continuous depiction of uncertainty and transitions in sleep staging \cite{stephansen2018neural}. For human annotations, it is constructed from the soft-consensus vectors $\hat{y}_{\mathfrak{C}}^t$ computed at each epoch $t$, offering a probabilistic summary of inter-scorer agreement over the course of a recording. Similarly, for automatic models, the hypnodensity graph is obtained by stacking the model’s predicted probability distributions $\hat{y}_m^t$ across epochs. In both cases, hypnodensities capture the temporal dynamics and ambiguity in staging decisions, enabling comparison between models and human annotators beyond discrete labels.

The cosine similarity between two probability distributions over sleep stages, such as the predicted distribution $\hat{y}_m^t$ from a model and the reference distribution $\hat{y}_{\mathfrak{C}}^t$ from the soft-consensus, at epoch $t$ is defined as:
\begin{equation}
\text{sim}_{m, \mathfrak{C}}^t = \frac{ \hat{y}_m^t \cdot \hat{y}_{\mathfrak{C}}^t }{ \| \hat{y}_m^t \| \| \hat{y}_{\mathfrak{C}}^t \| }
\end{equation}
This metric quantifies the angular similarity between two vectors, with values closer to 1 indicating greater alignment.
To summarize similarity across two hypnodensity graphs, the Averaged Cosine Similarity (ACS) \cite{fiorillo2023multi} is computed as the mean cosine similarity over all $T$ epochs:
\begin{equation}
\text{ACS}_{m, \mathfrak{C}} = \frac{1}{T} \sum_{t=1}^{T} \text{sim}_{m, \mathfrak{C}}^t
\end{equation}
In our work, ACS provides a single scalar value reflecting how closely a model’s hypnodensity output aligns with the soft-consensus of human scorers over the full recording.

To quantify variability within SOMNUS, we compute two complementary measures: \emph{Shannon entropy} of the ensemble output and inter-model disagreement based on \emph{pairwise cosine distance}.
Given a probability distribution $p \in \mathbb{R}^C$ over $C$ classes, the Shannon entropy $H$ is defined as
\begin{equation}
    H = - \sum_{c=1}^{C} p[c] \cdot \log \left( p[c] \right)
\end{equation}
Entropy reaches its maximum when the distribution is uniform and minimum when all probability mass is concentrated on a single class. In our work, we compute the entropy of SOMNUS at each epoch $t$, considering the probability distribution $\hat{y}_E^t$ over the five sleep stages. Higher entropy indicates greater uncertainty in the ensemble's output distribution.
To characterize inter-model disagreement, we compute the set of pairwise cosine distances between all unique pairs of model predictions at each epoch $t$:
\begin{equation}
    D^t = \left\{ 1 - \text{sim}_{m,n}^t \; \mid \; m < n, \; m,n \in \{1, \ldots, M\} \right\}
\end{equation}

where $\text{sim}_{m,n}^t$ denotes the cosine similarity between the predicted probability distributions of models $m$ and $n$, components of the ensemble.

Human consensus disagreement prediction is framed as a binary classification problem, with the target indicating whether human scorers disagreed at a given epoch. Logistic regression classifiers are trained using features derived from ensemble variability, namely, the Shannon entropy of SOMNUS output and summary statistics (mean, standard deviation, and maximum values at each epoch) of the pairwise cosine distance set $D^t$, capturing different facets of disagreement among ensemble constituents. Performance is evaluated separately on DOD-H and DOD-O using leave-one-recording-out cross-validation.

\section*{Ethical approval}

The secondary usage of the BSWR dataset was approved by the local ethics committee (Kantonale Ethikkommission Bern [KEK]-Nr. 2022-00415), ensuring compliance with the Human Research Act (HRA) and Ordinance on Human Research with the Exception of Clinical Trials (HRO). All methods were carried out in accordance with relevant guidelines and regulations. Written informed consent was obtained from all participants as of the introduction of the general consent process at Inselspital in 2015. Data were maintained with confidentiality throughout the study.

\section*{Data availability}

Due to patient confidentiality and ethical restrictions, the BSWR dataset from Inselspital, University Hospital Bern, is not publicly available. However, de-identified data may be obtained upon reasonable request, subject to a data sharing agreement and approval by the relevant ethics committees. For all other dataset access, kindly refer to the specific access details associated with each of them.

\section*{CODE AVAILABILITY}


SLEEPYLAND is an open-source toolbox publicly available at \url{https://github.com/biomedical-signal-processing/sleepyland}. The complete framework, including the pre-trained models, has been containerized and made accessible via Docker Hub at \url{https://hub.docker.com/r/bspsupsi/sleepyland}, ensuring reproducibility and ease of deployment. Additionally, we provide the uSLEEPYLAND repository \url{https://github.com/biomedical-signal-processing/uSLEEPYLAND}, which serves as an extension of the U-Time framework \cite{perslev2021u}. This revised version has been adapted for integration into the SLEEPYLAND project, incorporating additional ML models to enhance its utility in sleep staging analysis. uSLEEPYLAND expands the scope of available methods by exposing feature-based and deep learning models, which can be systematically evaluated within the SLEEPYLAND framework.

\bibliography{sn-bibliography}


\section*{Acknowledgements}

L.F. was supported by "\textit{SLEEPYLAND: A python library to analyze the large amount of NSRR sleep data via deep learning algorithms}" project, from the Swiss Open Research Data Grants funding program from swissuniversities (P-5B, Open Science (2021-2024) - TP80). The National Sleep Research Resource was supported by the National Heart, Lung, and Blood Institute (R24 HL114473, 75N92019R002). The Apnea, Bariatric surgery, and CPAP study (ABC Study) was supported by National Institutes of Health grants R01HL106410 and K24HL127307. Philips Respironics donated the CPAP machines and supplies used in the perioperative period for patients undergoing bariatric surgery. The Sleep Disordered Breathing, ApoE and Lipid Metabolism Study was supported by the National Institute of Health (5R01HL71515-3). The Apnea Positive Pressure Long-term Efficacy Study (APPLES) was supported by the National Heart, Lung, and Blood Institute (U01HL68060). The Cleveland Children's Sleep and Health Study (CCSHS) was supported by grants from the National Institutes of Health (RO1HL60957, K23 HL04426, RO1 NR02707, M01 Rrmpd0380-39). The Cleveland Family Study (CFS) was supported by grants from the National Institutes of Health (HL46380, M01 RR00080-39, T32-HL07567, RO1-46380). The Childhood Adenotonsillectomy Trial (CHAT) was supported by the National Institutes of Health (HL083075, HL083129, UL1-RR-024134, UL1 RR024989). The Home Positive Airway Pressure study (HomePAP) was supported by the American Sleep Medicine Foundation 38-PM-07 Grant: Portable Monitoring for the Diagnosis and Management of OSA. The Multi-Ethnic Study of Atherosclerosis (MESA) Sleep Ancillary study was funded by NIH-NHLBI Association of Sleep Disorders with Cardiovascular Health Across Ethnic Groups (RO1 HL098433). MESA is supported by NHLBI funded contracts HHSN268201500003I, N01-HC-95159, N01-HC-95160, N01-HC-95161, N01-HC-95162, N01-HC-95163, N01-HC-95164, N01-HC-95165, N01-HC-95166, N01-HC-95167, N01-HC-95168 and N01-HC-95169 from the National Heart, Lung, and Blood Institute, and by cooperative agreements UL1-TR-000040, UL1-TR-001079, and UL1-TR-001420 funded by NCATS. The Mignot Nature Communications research was mostly supported by a grant from Jazz Pharmaceuticals to E.M. Additional funding came from: NIH grant R01HL62252 (to P.E.P.); Ministry of Science and Technology 2015CB856405 and National Foundation of Science of China 81420108002,81670087 (to F.H.); H. Lundbeck A/S, Lundbeck Foundation, Technical University of Denmark and Center for Healthy Aging, University of Copenhagen (to P.J. and H.B.D.S). Additional support was provided by the Klarman Family, Otto Mønsted, Stibo, Vera \& Carl Johan Michaelsens, Knud Højgaards, Reinholdt W. Jorck and Hustrus and Augustinus Foundations (to A.N.O.). The National Heart, Lung, and Blood Institute provided funding for the ancillary MrOS Sleep Study, "Outcomes of Sleep Disorders in Older Men," under the following grant numbers: R01 HL071194, R01 HL070848, R01 HL070847, R01 HL070842, R01 HL070841, R01 HL070837, R01 HL070838, and R01 HL070839. Data provided for this report was funded by National Institutes of Health grant R01 HD079411, awarded to Janet A. DiPietro, Johns Hopkins Bloomberg School of Public Health, Johns Hopkins University. NCH Sleep DataBank was supported by the National Institute of Biomedical Imaging and Bioengineering of the National Institutes of Health under Award Number R01EB025018. The Sleep Heart Health Study (SHHS) was supported by National Heart, Lung, and Blood Institute cooperative agreements U01HL53916 (University of California, Davis), U01HL53931 (New York University), U01HL53934 (University of Minnesota), U01HL53937 and U01HL64360 (Johns Hopkins University), U01HL53938 (University of Arizona), U01HL53940 (University of Washington), U01HL53941 (Boston University), and U01HL63463 (Case Western Reserve University). The Study of Osteoporotic Fractures (SOF) was supported by National Institutes of Health grants (AG021918, AG026720, AG05394, AG05407, AG08415, AR35582, AR35583, AR35584, RO1 AG005407, R01 AG027576-22, 2 R01 AG005394-22A1, 2 RO1 AG027574-22A1, HL40489, T32 AG000212-14). This Wisconsin Sleep Cohort Study was supported by the U.S. National Institutes of Health, National Heart, Lung, and Blood Institute (R01HL62252), National Institute on Aging (R01AG036838, R01AG058680), and the National Center for Research Resources (1UL1RR025011).

\section*{Author contribution}

A.D.R., M.M., M.B., F.D.F., and L.F. contributed to the design of the study. A.D.R., M.M. and L.F. implemented the SLEEPYLAND system, and they conducted all the experiments related to the benchmarking, ensembling and generalization, plus all the experiments related to the model-ensemble versus human-ensemble quantification. M.B. conducted all the experiments on the quantification of model-bias on performance and clinical markers. L.F., A.D.R. and M.B. contributed to drafting figures and tables. L.F., A.D.R., M.B. and F.D.F. wrote the paper with feedback from I.F., J.v.d.M, and M.H.S. We thank A.T. and C.L.A.B. for their critical reading. The BSWR dataset was extracted/prepared by J.v.d.M, L.F. All authors approved the final paper.

\section*{Competing interests}

The authors declare no competing interests.

\end{document}


\title[SLEEPYLAND]{SLEEPYLAND: trust begins with fair evaluation of automatic sleep staging models}


\author[1,2]{\fnm{Alvise} \sur{Dei Rossi}}\email{alvise.dei.rossi@usi.ch}

\author[2]{\fnm{Matteo} \sur{Metaldi}}\email{matteo.metaldi@supsi.ch}

\author[2,3]{\fnm{Michal} \sur{Bechny}}\email{michal.bechny@supsi.ch}

\author[4]{\fnm{Irina} \sur{Filchenko}}\email{irina.filchenko@insel.ch}

\author[4]{\fnm{Julia} \sur{van der Meer}}\email{julia.vandermeer@insel.ch}

\author[4]{\fnm{Markus H.} \sur{Schmidt}}\email{markus.schmidt@insel.ch}

\author[4]{\fnm{Claudio L.A.} \sur{Bassetti}}\email{claudio.bassetti@insel.ch}

\author[3,4]{\fnm{Athina} \sur{Tzovara}}\email{athina.tzovara@inf.unibe.ch}

\author[2]{\fnm{Francesca D.} \sur{Faraci}}\email{francesca.faraci@supsi.ch}

\author*[2,5]{\fnm{Luigi} \sur{Fiorillo}}\email{luigi.fiorillo@supsi.ch}

\affil[1]{\orgdiv{Faculty of informatics}, \orgname{Università della Svizzera Italiana}, \orgaddress{\street{Via Giuseppe Buffi 13}, \city{Lugano}, \postcode{6900}, \country{Switzerland}}}

\affil*[2]{\orgdiv{Institute of Digital Technologies for Personalized Healthcare $\vert$ MeDiTech}, Department of Innovative Technologies, \orgname{University of Applied Sciences and Arts of Southern Switzerland}, \orgaddress{\street{Via la Santa 1}, \city{Lugano}, \postcode{6962}, \country{Switzerland}}}

\affil[3]{\orgdiv{Institute of Informatics}, \orgname{University of Bern}, \orgaddress{\street{Neubr{\"u}ckstrasse 10}, \city{Bern}, \postcode{3012}, \country{Switzerland}}}

\affil[4]{\orgdiv{Sleep Wake Epilepsy Center $\vert$ NeuroTec},  Department of Neurology, \orgname{Inselspital, Bern University Hospital, University of Bern}, \orgaddress{\street{Freiburgstrasse}, \city{Bern}, \postcode{3010}, \country{Switzerland}}}

\affil*[5]{\orgdiv{Neurocenter of Southern Switzerland}, \orgname{Ente Ospedaliero Cantonale}, \orgaddress{\street{Via Tesserete 46}, \city{Lugano}, \postcode{6900}, \country{Switzerland}}}

\maketitle

\newpage

\section*{List of Supplementary Figures}

\begin{figure}[h]
    \renewcommand{\figurename}{Supplementary Figure}
    \centering
    \includegraphics[width=1\linewidth]{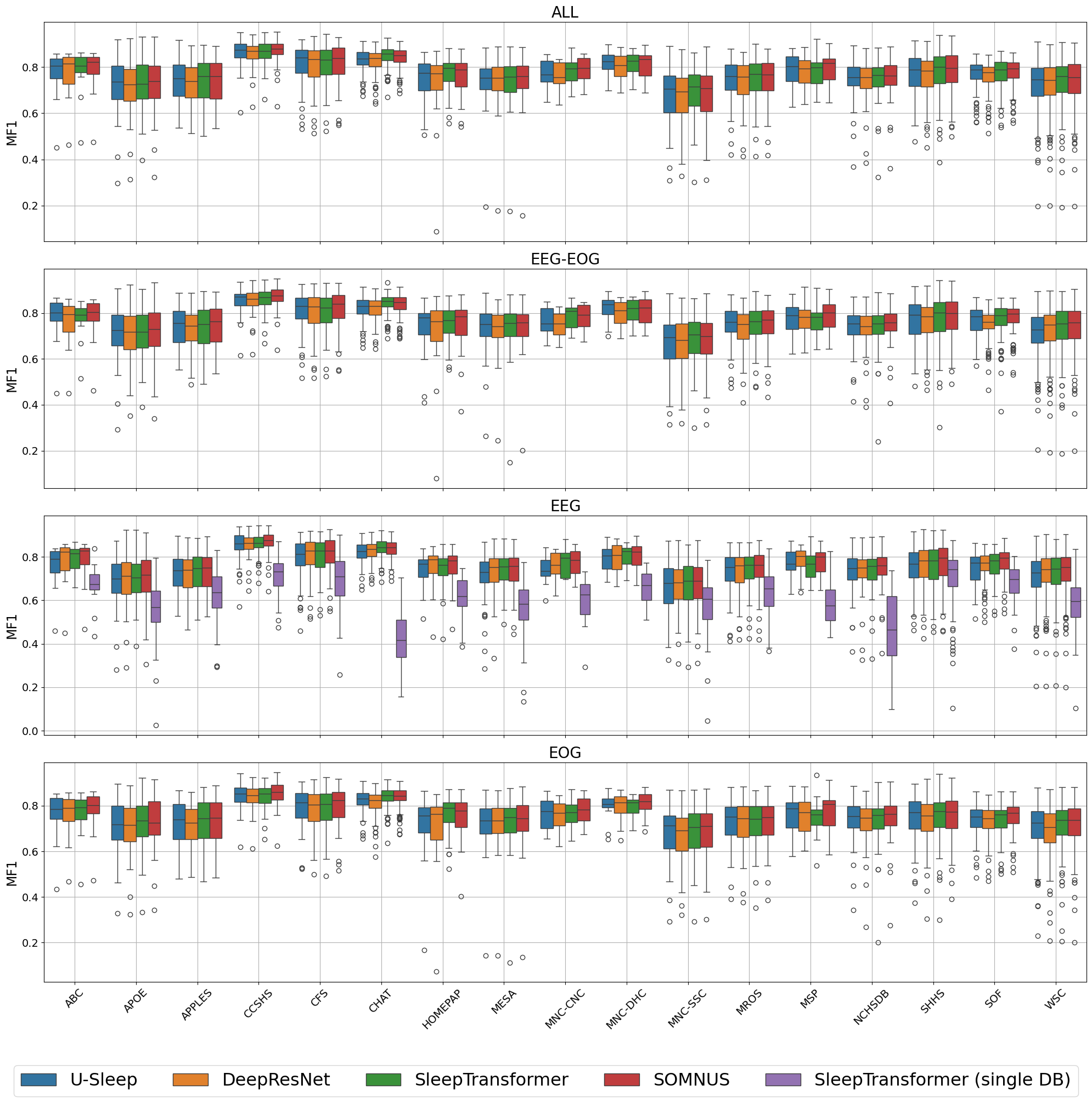}
    \caption{Recording-wise Macro-F1 distributions for all models in SLEEPYLAND on all the test set partitions of in-domain (ID) datasets.}
    \label{fig:boxplots_ID}
\end{figure}

\begin{figure}
    \renewcommand{\figurename}{Supplementary Figure}
    \centering
    \includegraphics[width=1\linewidth]{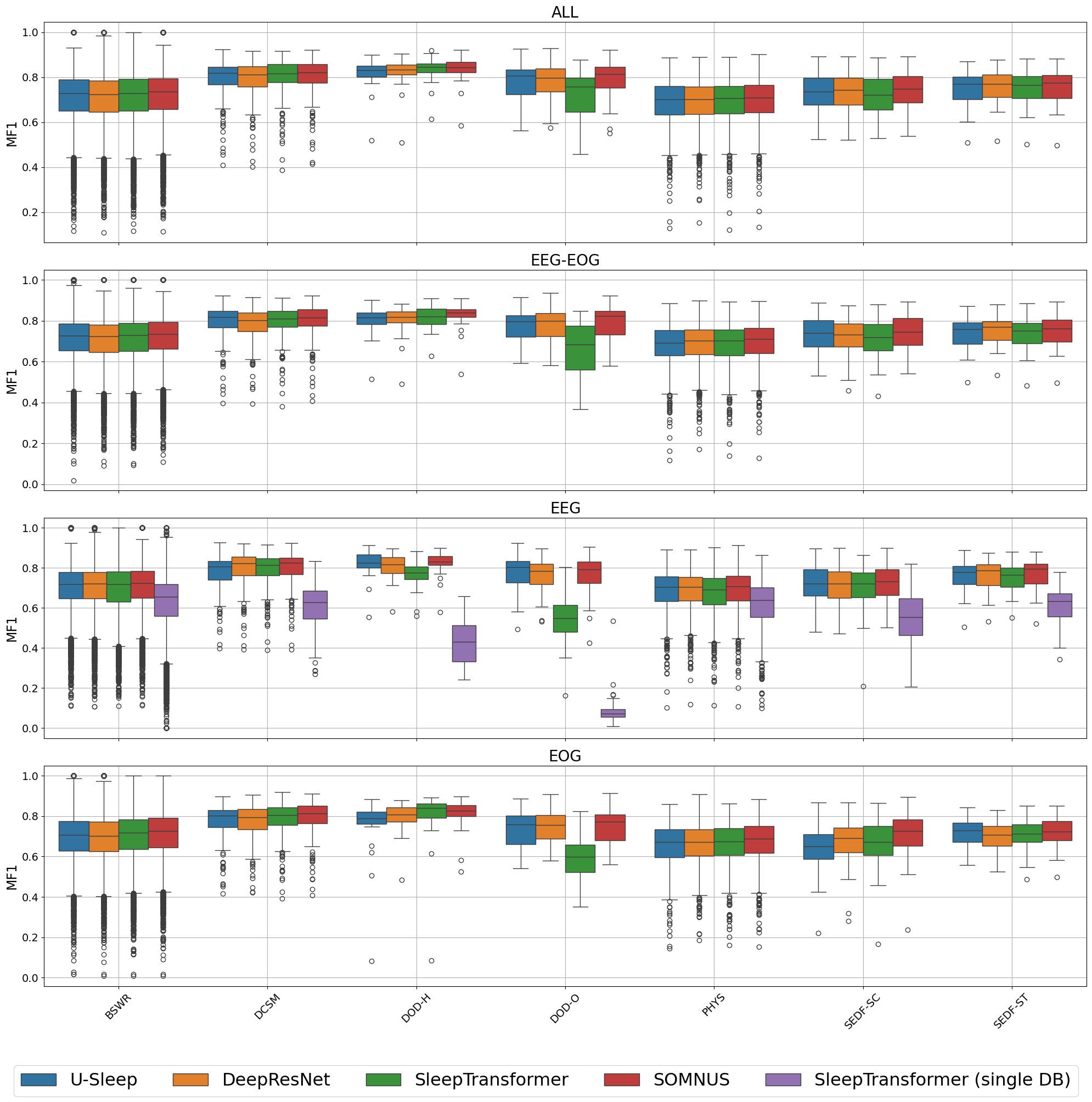}
    \caption{Recording-wise Macro-F1 distributions for all models in SLEEPYLAND on out-of-domain (OOD) datasets.}
    \label{fig:boxplots_OOD}
\end{figure}

\begin{figure}[htbp]
    \renewcommand{\figurename}{Supplementary Figure}
    \centering
    \includegraphics[width=\textwidth]{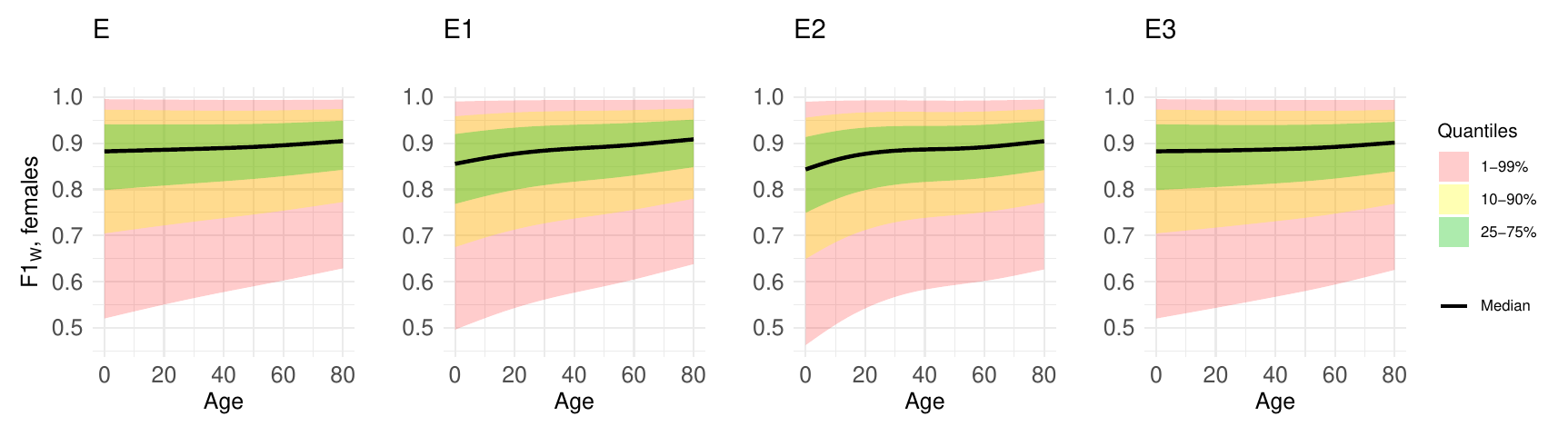}
    \includegraphics[width=\textwidth]{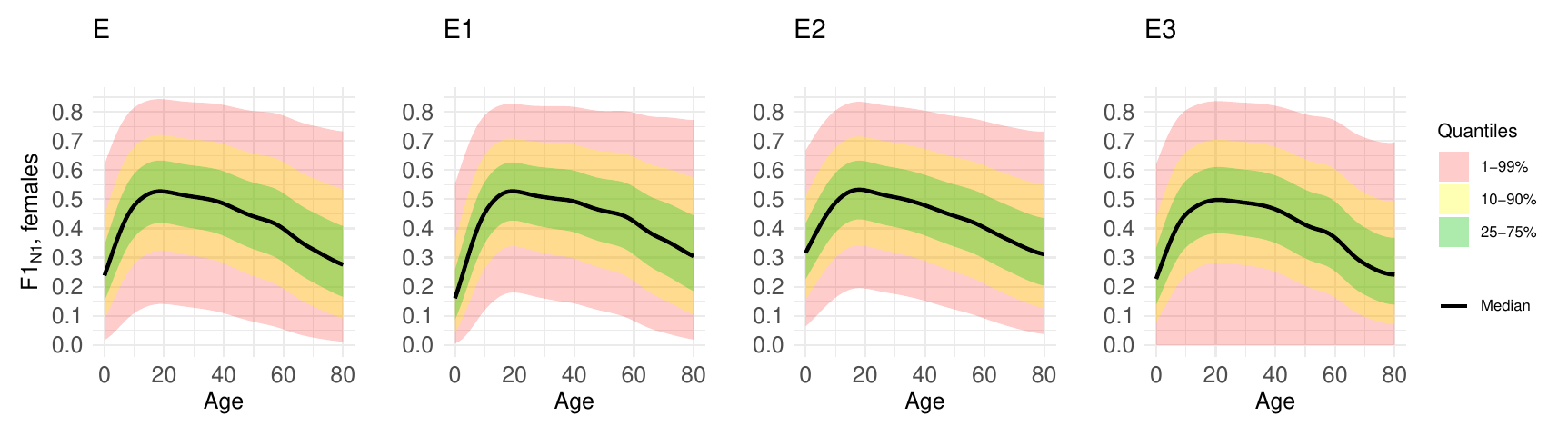}
    \includegraphics[width=\textwidth]{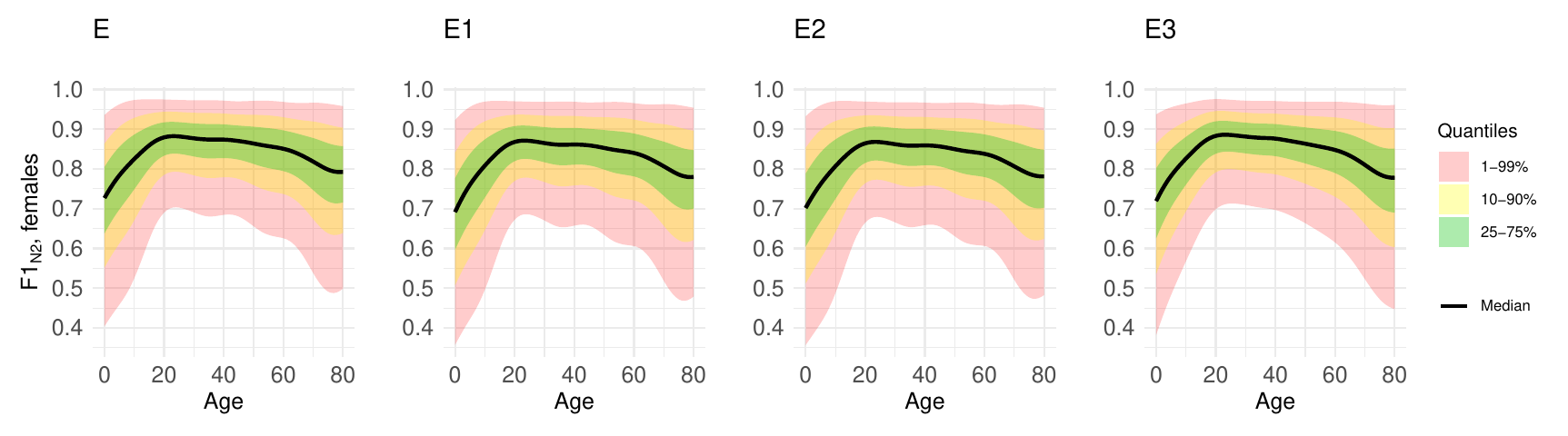}
    \includegraphics[width=\textwidth]{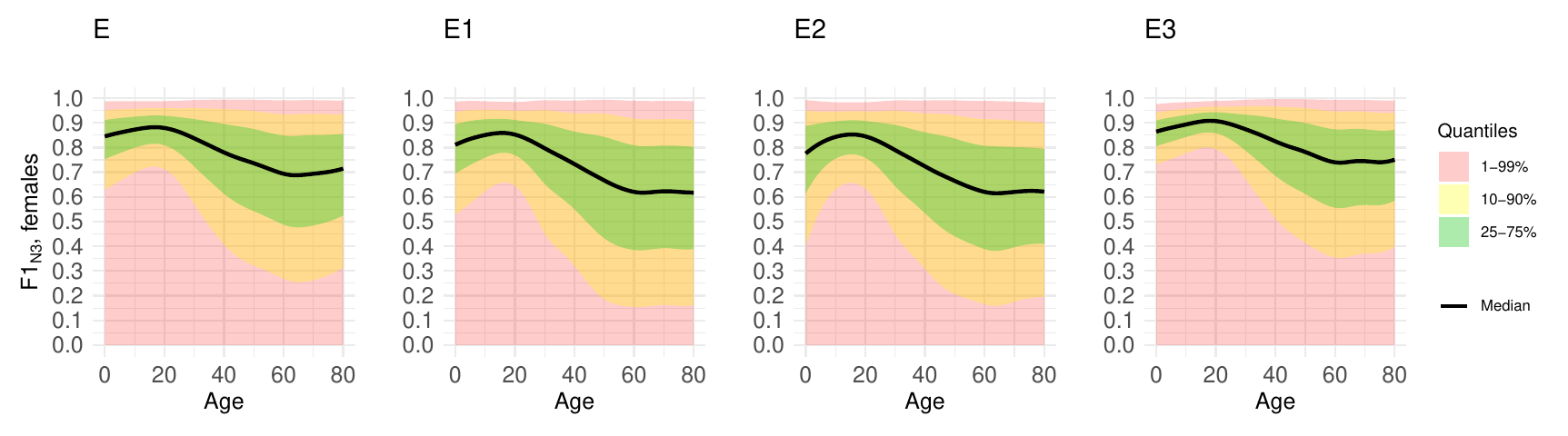}
    \includegraphics[width=\textwidth]{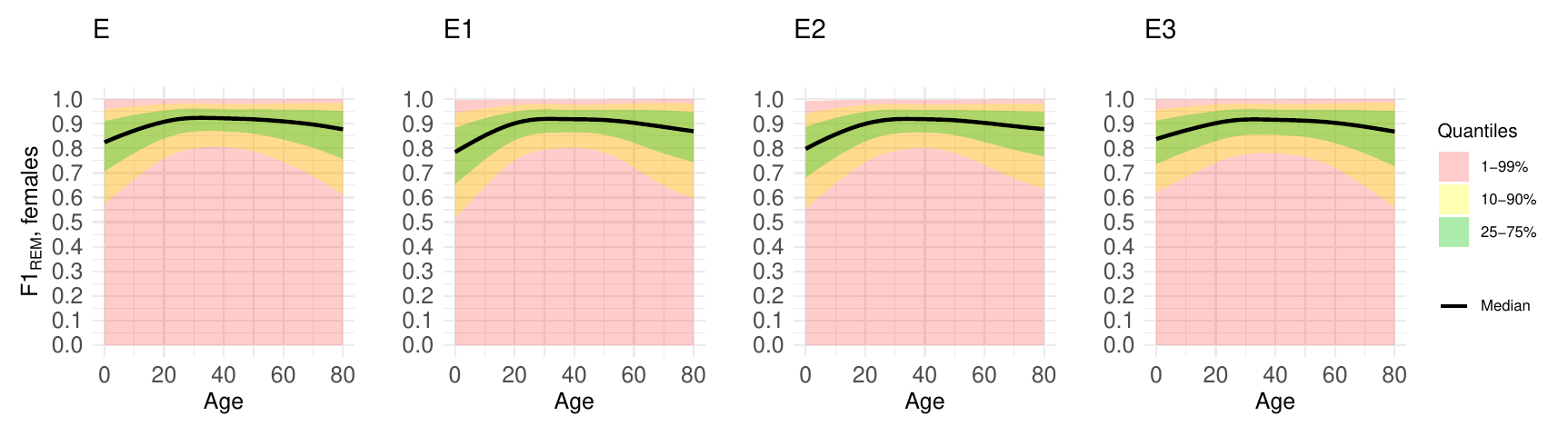}
    \caption{Age-conditioned expected distributions of stage-specific F1-scores quantiles for females (under an optimistic scenario of AHI = PLMI = 0) across four sleep scoring models, i.e., SOMNUS, SOMNUS$_\text{U‑Sleep}$, SOMNUS$_\text{DeepResNet}$, and SOMNUS$_\text{SleepTransformer}$.}
    \label{supp_fig:f1_stages_female}
\end{figure}

\begin{figure}[htbp]
    \renewcommand{\figurename}{Supplementary Figure}
    \centering
    \includegraphics[width=\textwidth]{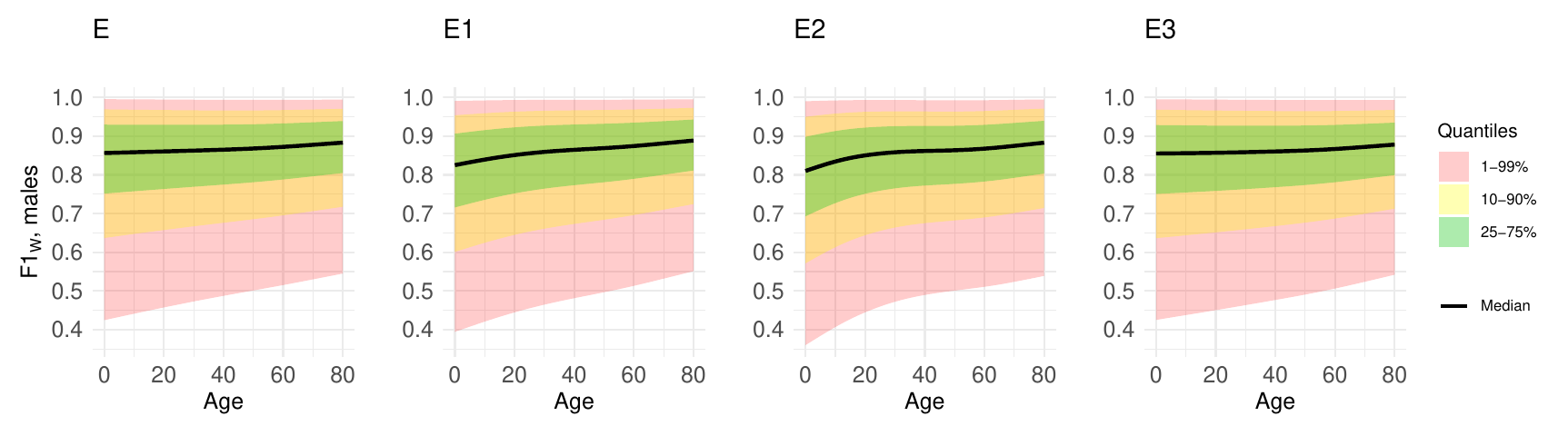}
    \includegraphics[width=\textwidth]{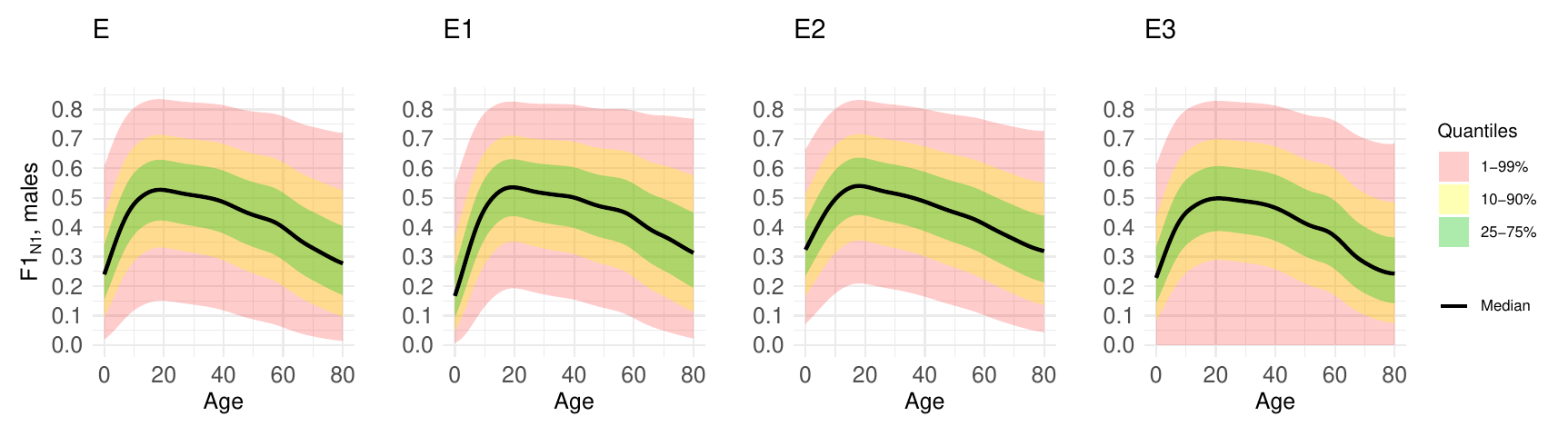}
    \includegraphics[width=\textwidth]{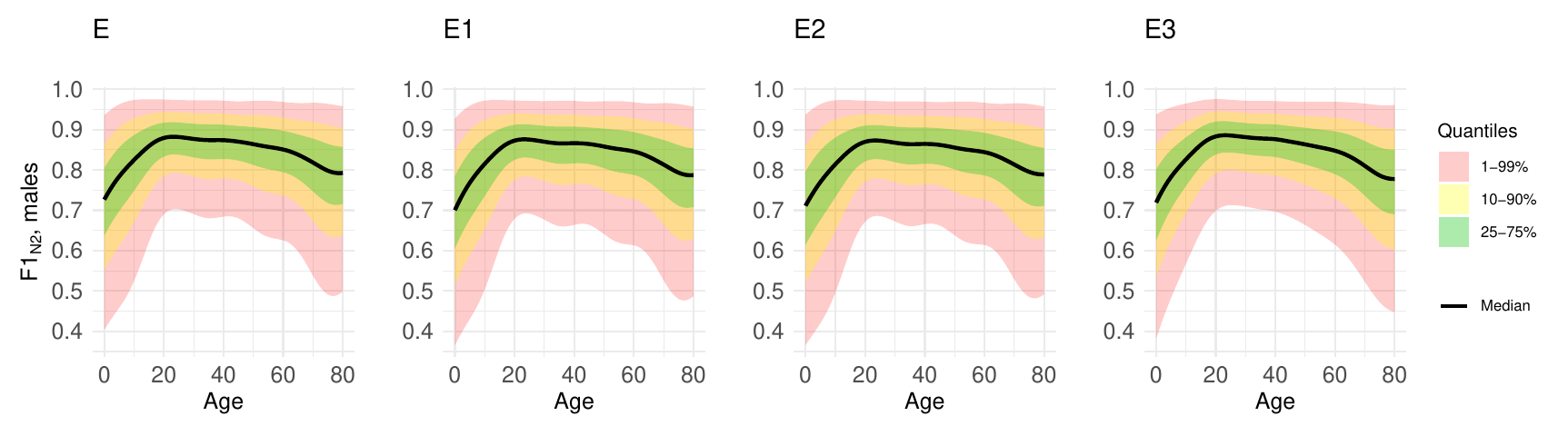}
    \includegraphics[width=\textwidth]{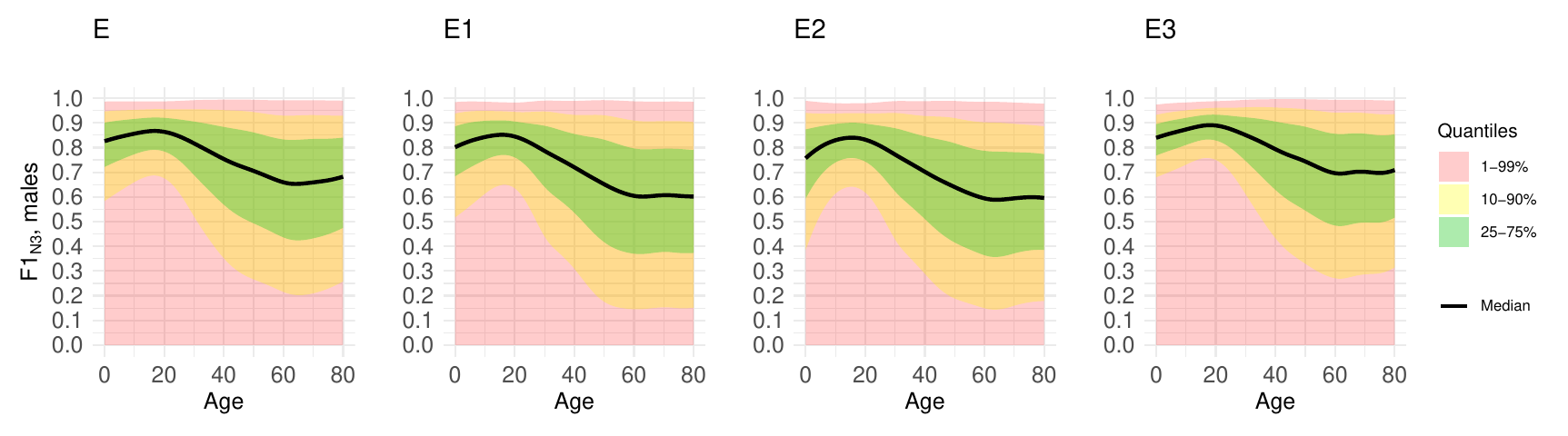}
    \includegraphics[width=\textwidth]{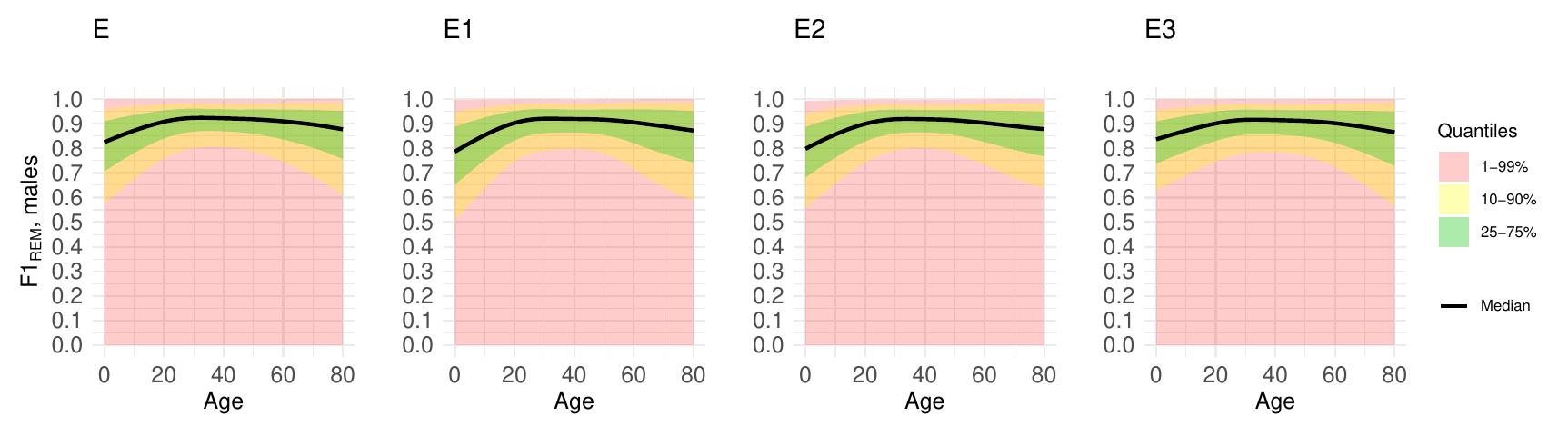}
    \caption{Age-conditioned expected distributions of stage-specific F1-scores quantiles for males (under an optimistic scenario of AHI = PLMI = 0) across four sleep scoring models, i.e., SOMNUS, SOMNUS$_\text{U‑Sleep}$, SOMNUS$_\text{DeepResNet}$, and SOMNUS$_\text{SleepTransformer}$.}
    \label{supp_fig:f1_stages_male}
\end{figure}

\begin{figure}[htbp]
    \renewcommand{\figurename}{Supplementary Figure}
    \centering
    \includegraphics[width=\textwidth]{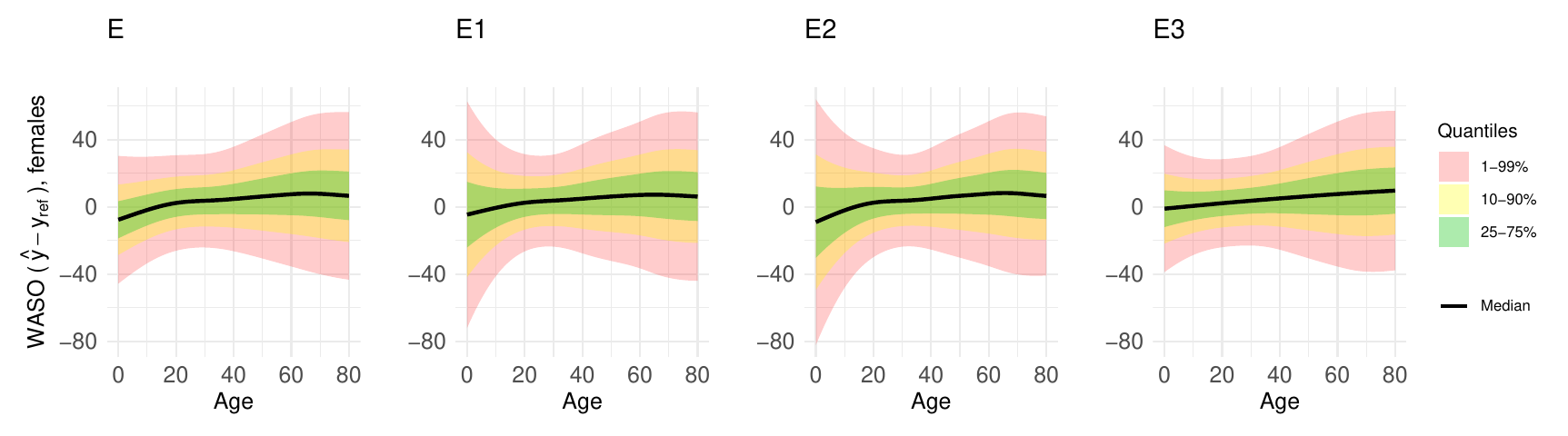}
    \includegraphics[width=\textwidth]{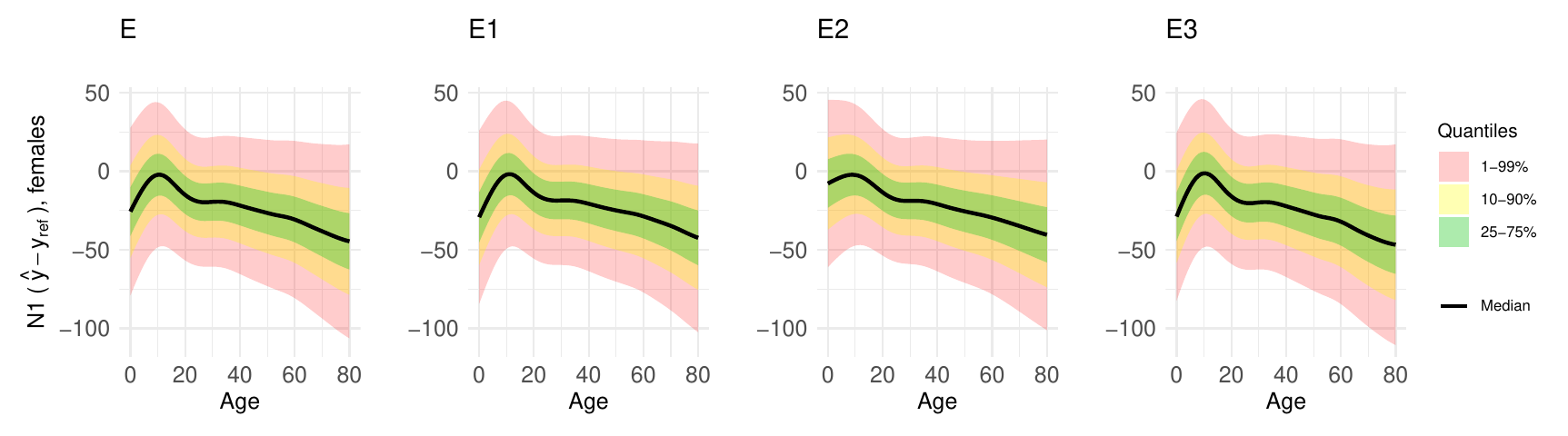}
    \includegraphics[width=\textwidth]{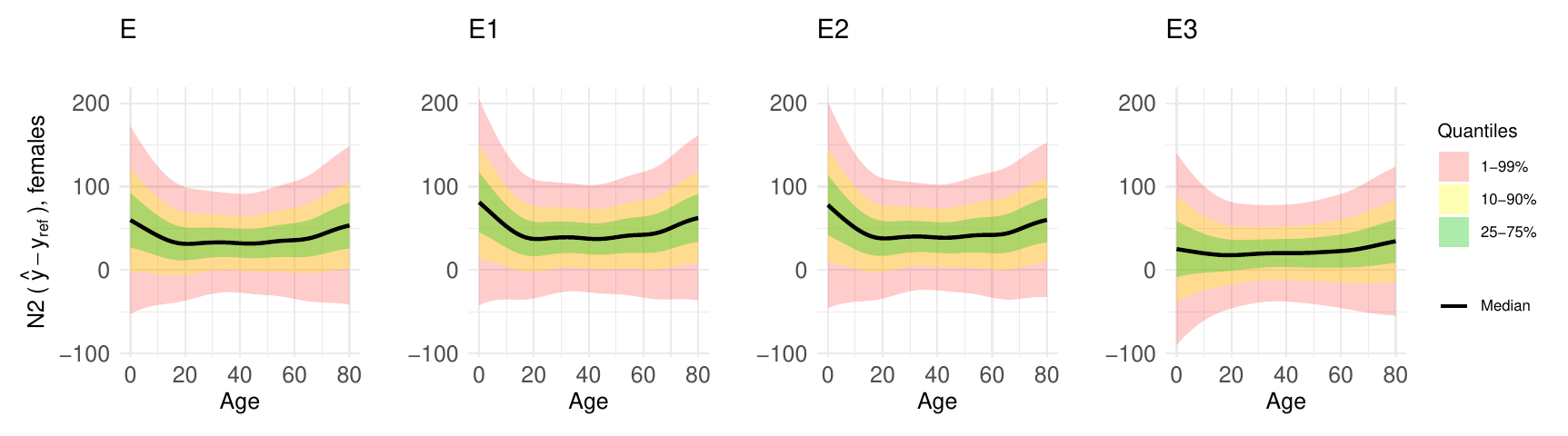}
    \includegraphics[width=\textwidth]{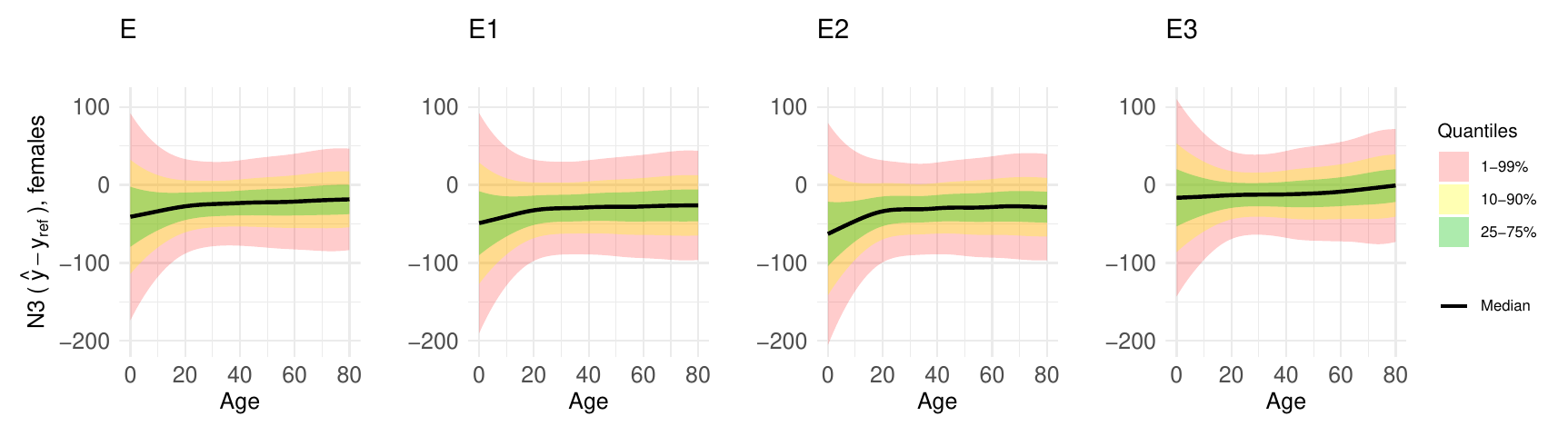}
    \includegraphics[width=\textwidth]{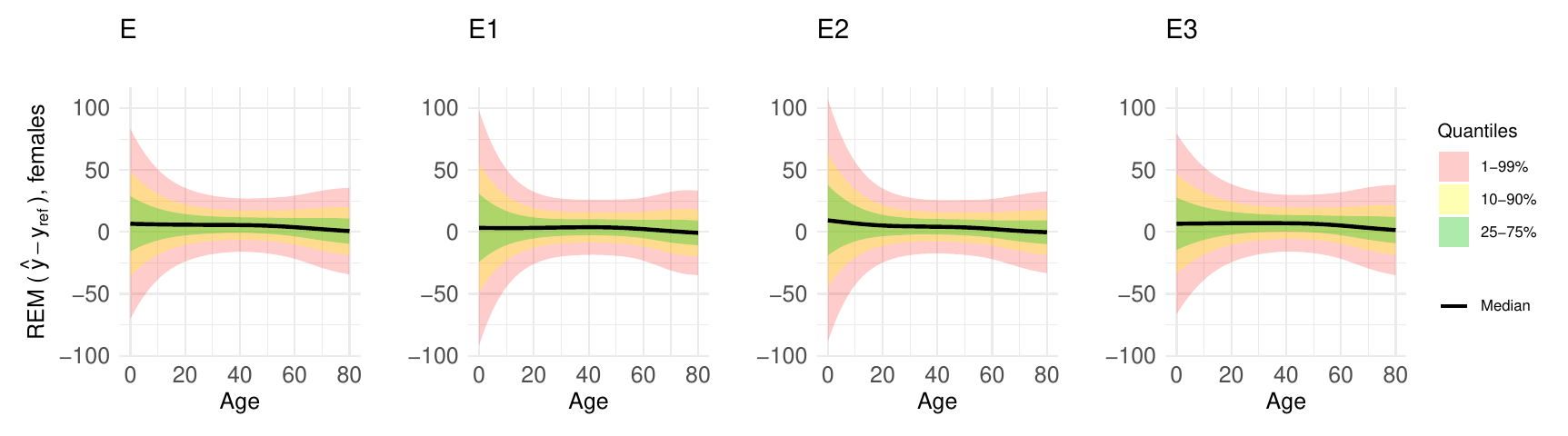}
    \caption{Age-conditioned expected distributions of stage-durations quantiles for females (under an optimistic scenario of AHI = PLMI = 0) across four sleep scoring models, i.e., SOMNUS, SOMNUS$_\text{U‑Sleep}$, SOMNUS$_\text{DeepResNet}$, and SOMNUS$_\text{SleepTransformer}$.}
    \label{supp_fig:markers_stages_female}
\end{figure}
\begin{figure}[htbp]
    \renewcommand{\figurename}{Supplementary Figure}
    \centering
    \includegraphics[width=\textwidth]{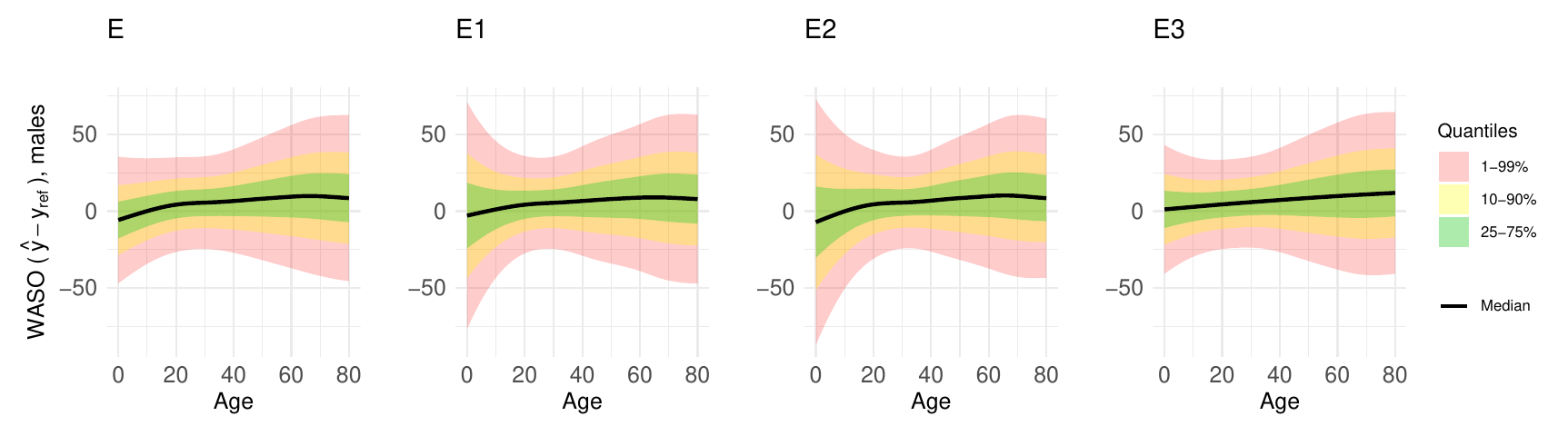}
    \includegraphics[width=\textwidth]{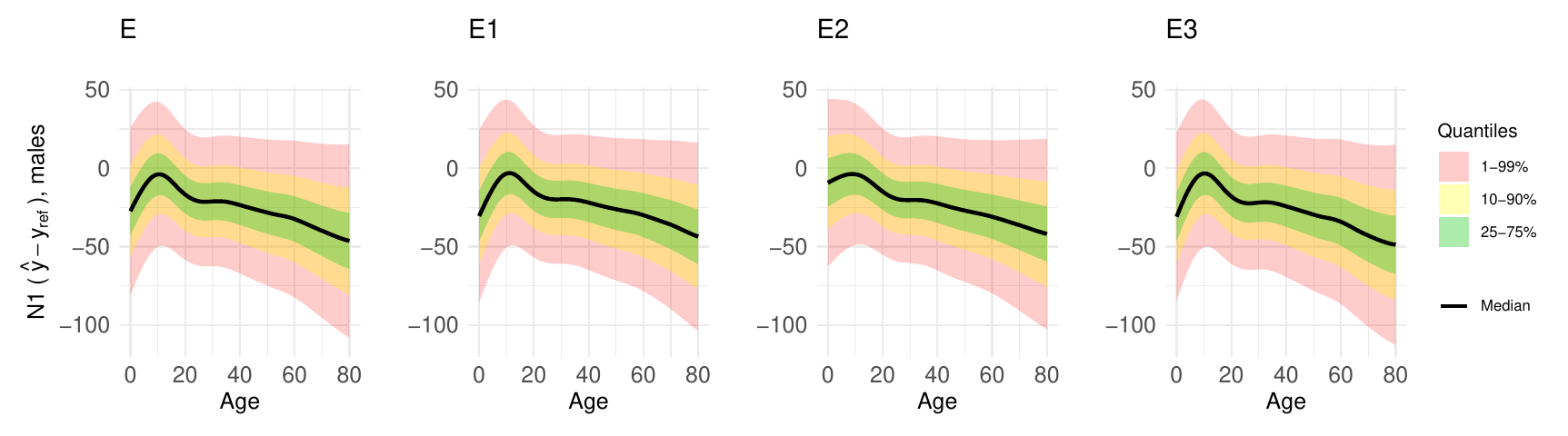}
    \includegraphics[width=\textwidth]{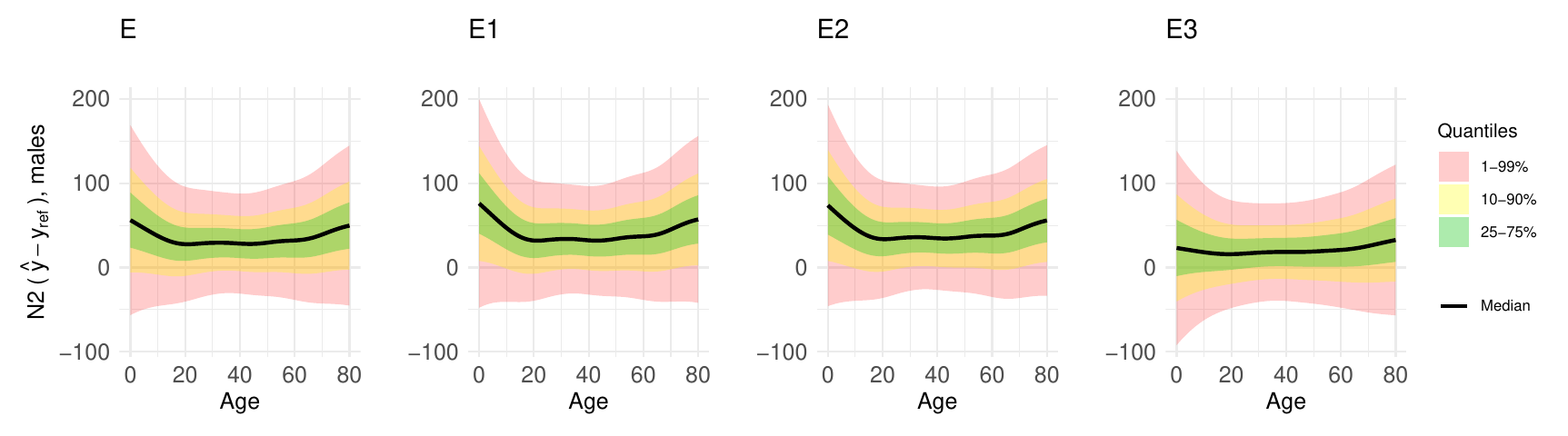}
    \includegraphics[width=\textwidth]{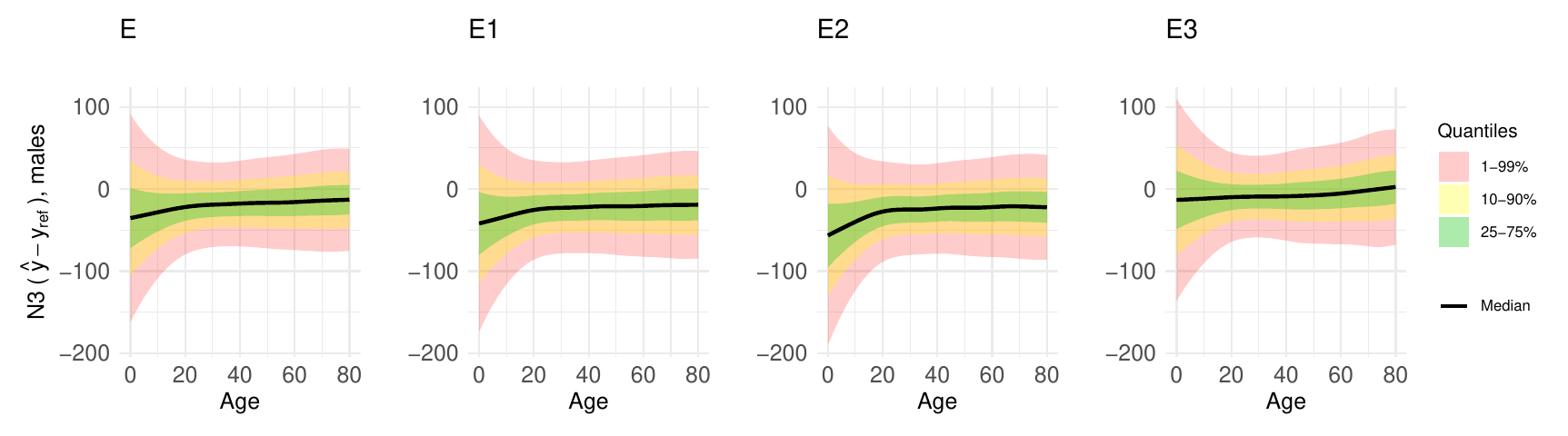}
    \includegraphics[width=\textwidth]{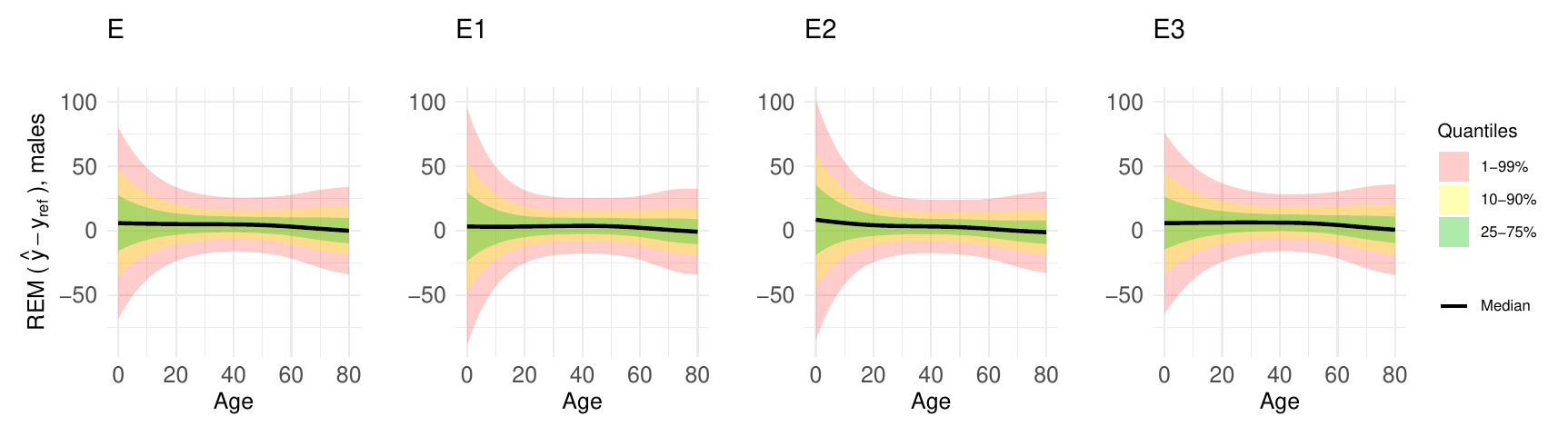}
    \caption{Age-conditioned expected distributions of stage-durations quantiles for males (under an optimistic scenario of AHI = PLMI = 0) across four sleep scoring models, i.e., SOMNUS, SOMNUS$_\text{U‑Sleep}$, SOMNUS$_\text{DeepResNet}$, and SOMNUS$_\text{SleepTransformer}$.}
    \label{supp_fig:markers_stages_male}
\end{figure}

\begin{figure}[htbp]
    \renewcommand{\figurename}{Supplementary Figure}
    \centering
    \includegraphics[width=\textwidth]{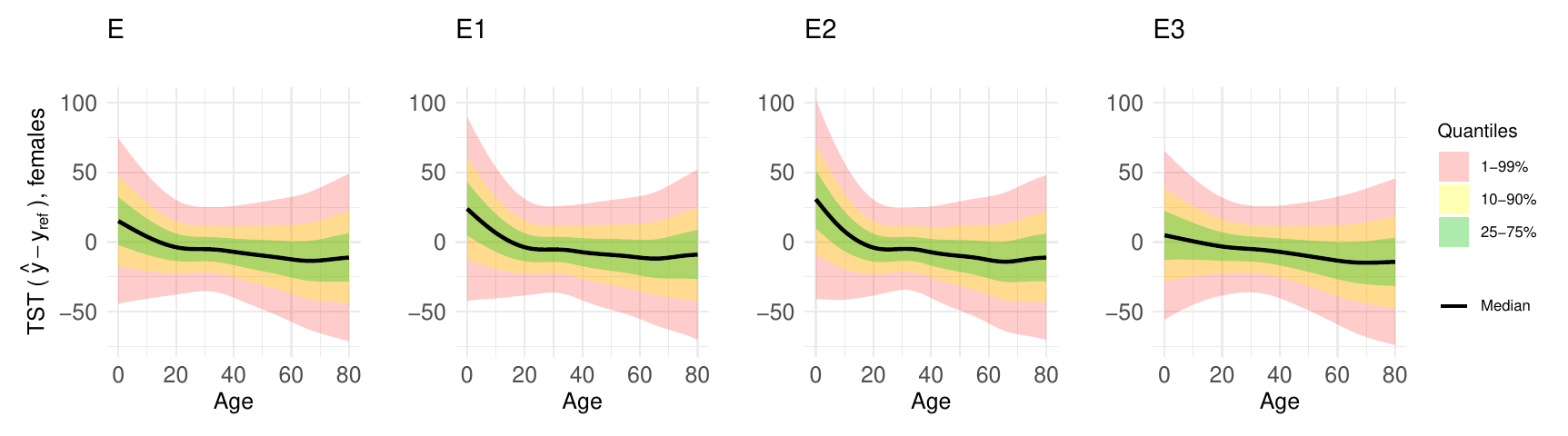}
    \includegraphics[width=\textwidth]{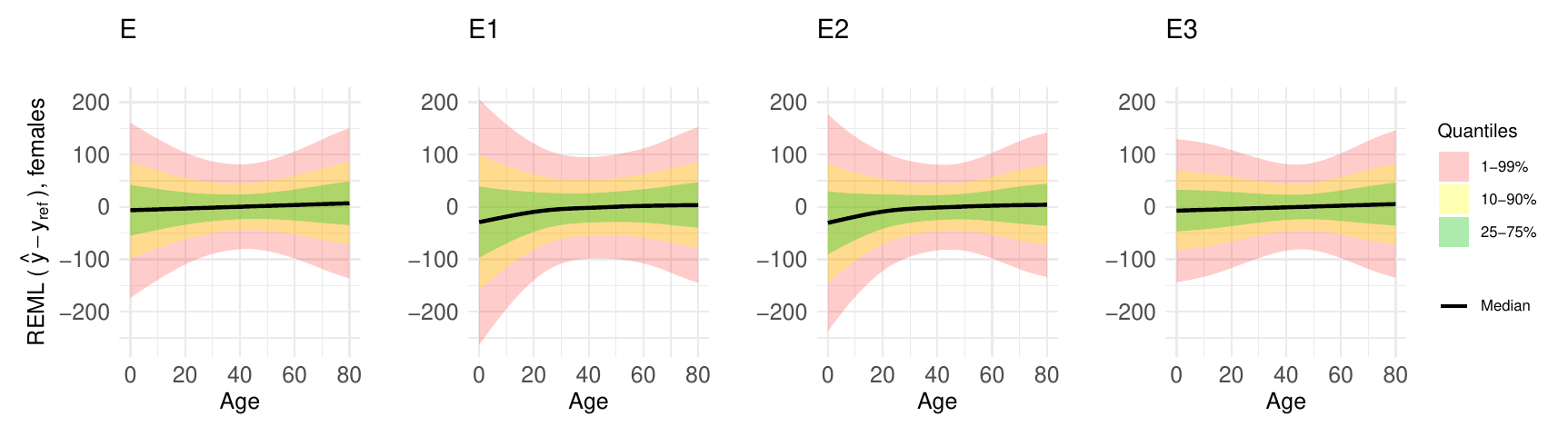}
    \includegraphics[width=\textwidth]{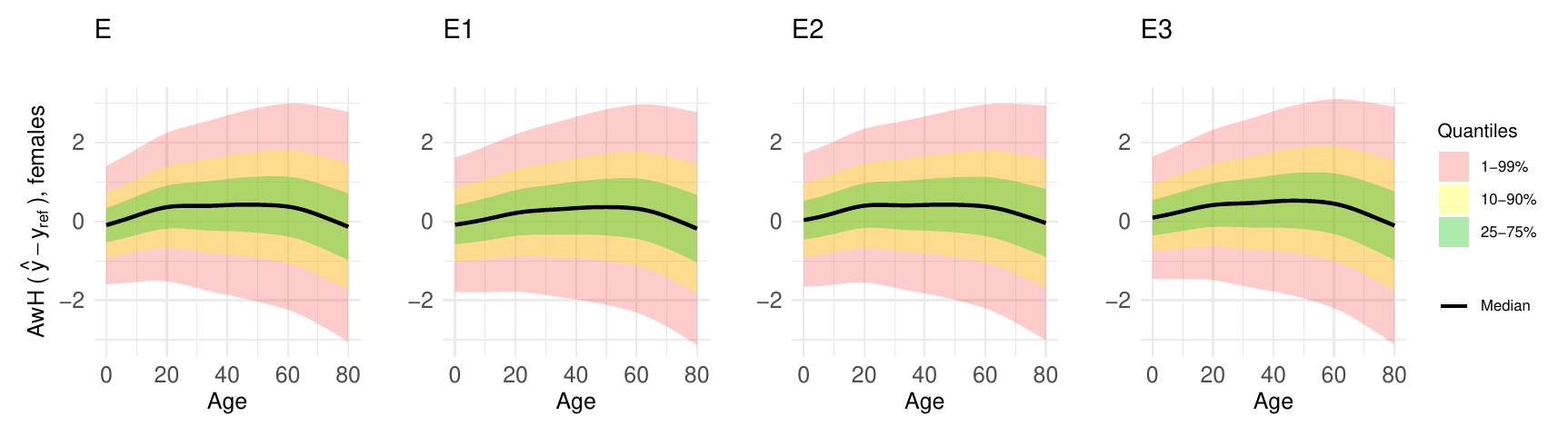}
    \includegraphics[width=\textwidth]{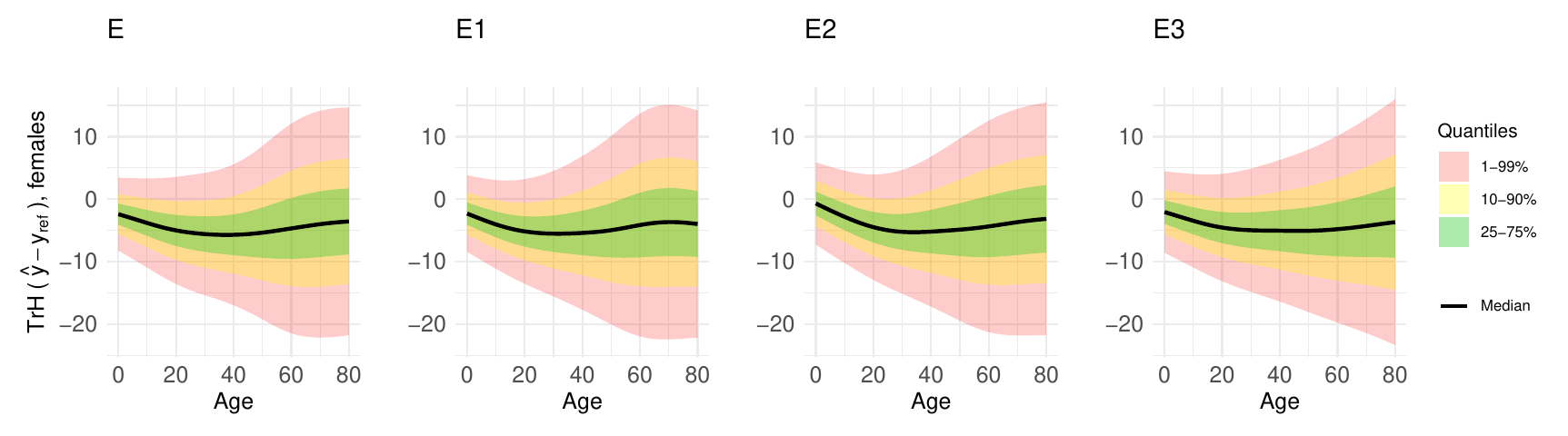}
    \caption{Age-conditioned expected distributions of Total Sleep Time (TST), REM latency (REML), Awakenings per Hour (AwH), and Tranistions per Hour (TrH) quantiles for females (under an optimistic scenario of AHI = PLMI = 0) across four sleep scoring models, i.e., SOMNUS, SOMNUS$_\text{U‑Sleep}$, SOMNUS$_\text{DeepResNet}$, and SOMNUS$_\text{SleepTransformer}$.}
    \label{supp_fig:bias_others_female}
\end{figure}

\begin{figure}[htbp]
    \renewcommand{\figurename}{Supplementary Figure}
    \centering
    \includegraphics[width=\textwidth]{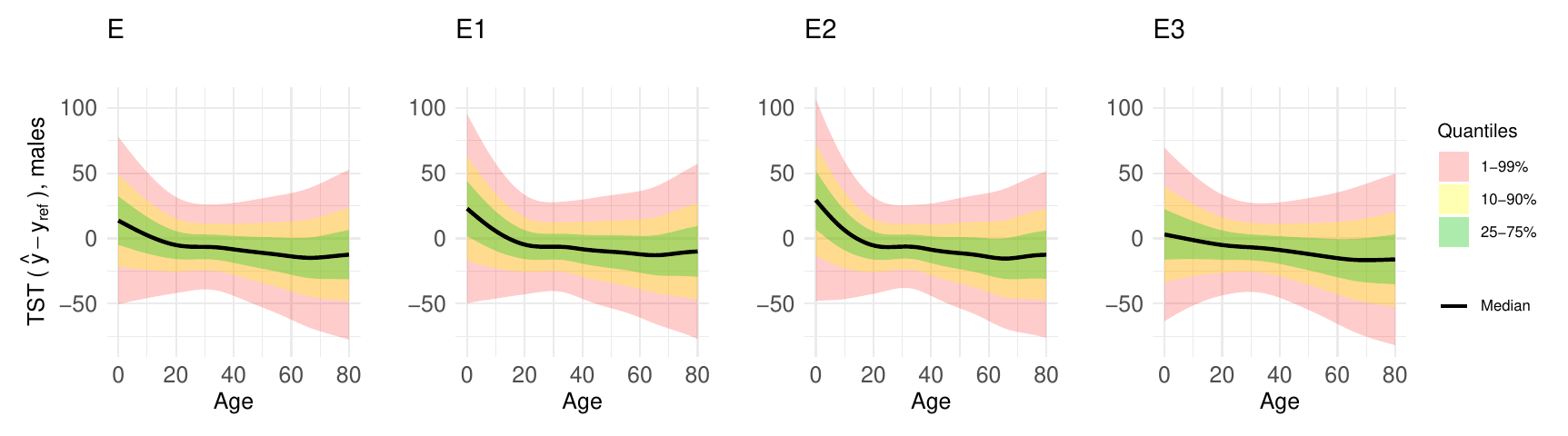}
    \includegraphics[width=\textwidth]{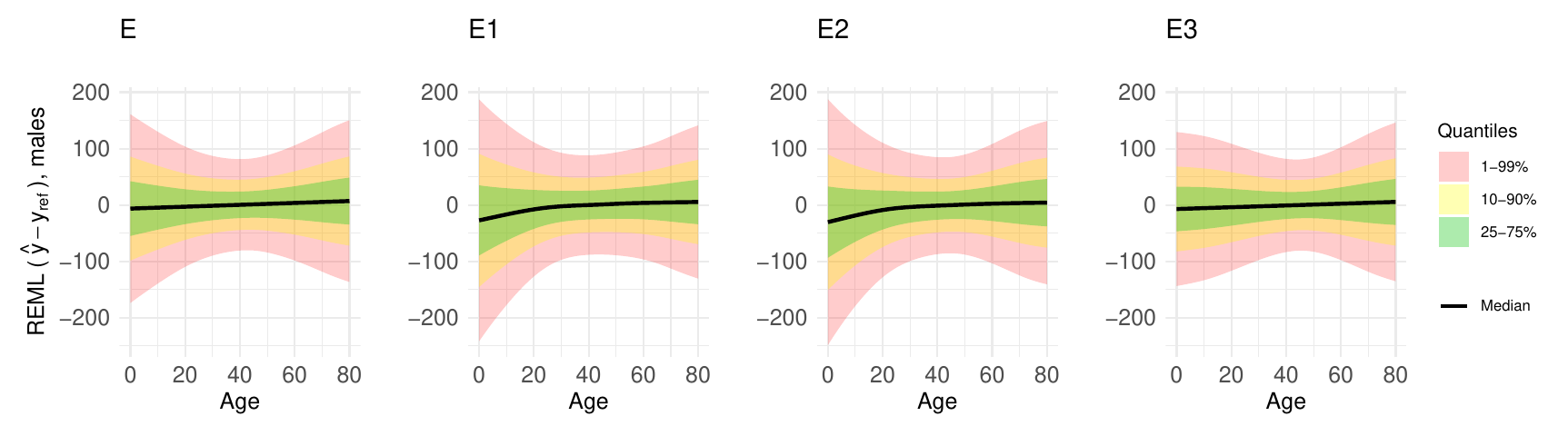}
    \includegraphics[width=\textwidth]{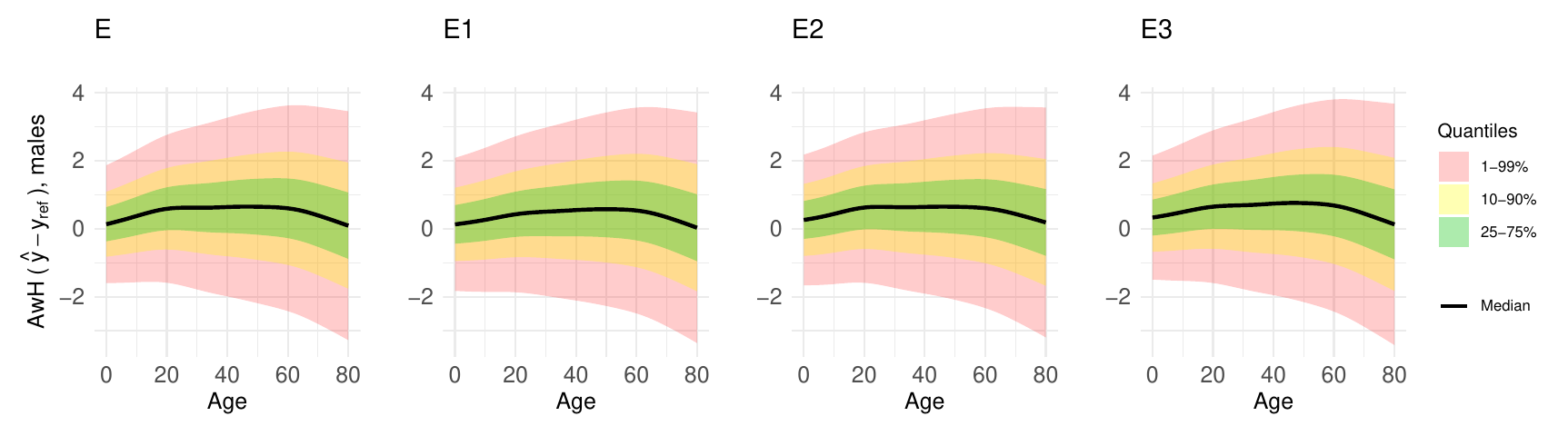}
    \includegraphics[width=\textwidth]{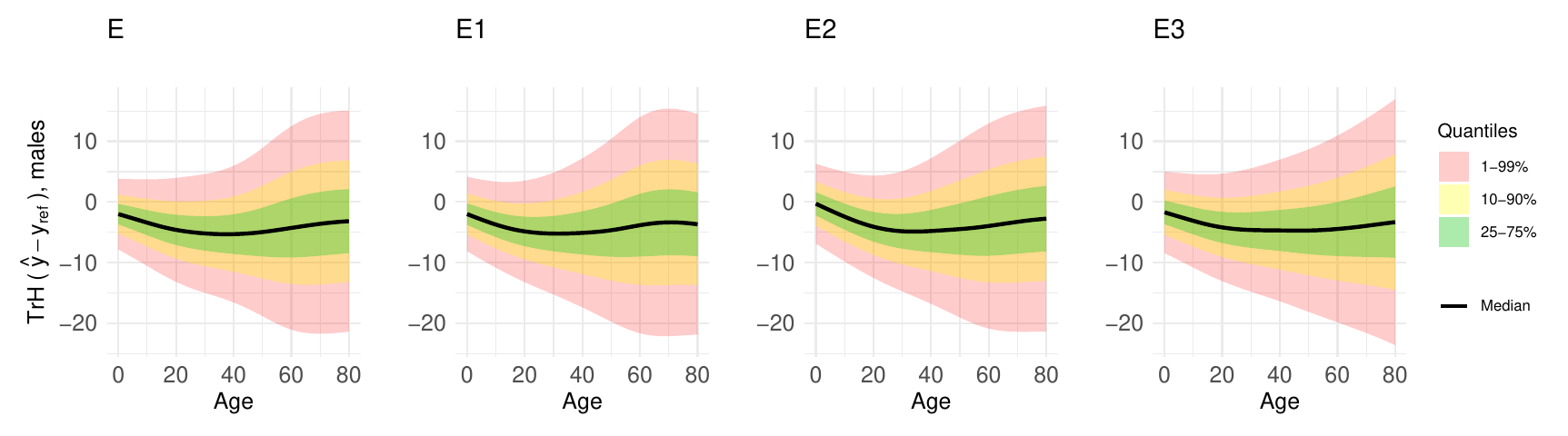}
    \caption{Age-conditioned expected distributions of Total Sleep Time (TST), REM latency (REML), Awakenings per Hour (AwH), and Tranistions per Hour (TrH) quantiles for males (under an optimistic scenario of AHI = PLMI = 0) across four sleep scoring models, i.e., SOMNUS, SOMNUS$_\text{U‑Sleep}$, SOMNUS$_\text{DeepResNet}$, and SOMNUS$_\text{SleepTransformer}$.}
    \label{supp_fig:bias_others_male}
\end{figure}

\begin{figure}[h]
    \renewcommand{\figurename}{Supplementary Figure}
    \centering
    \includegraphics[width=1\linewidth]{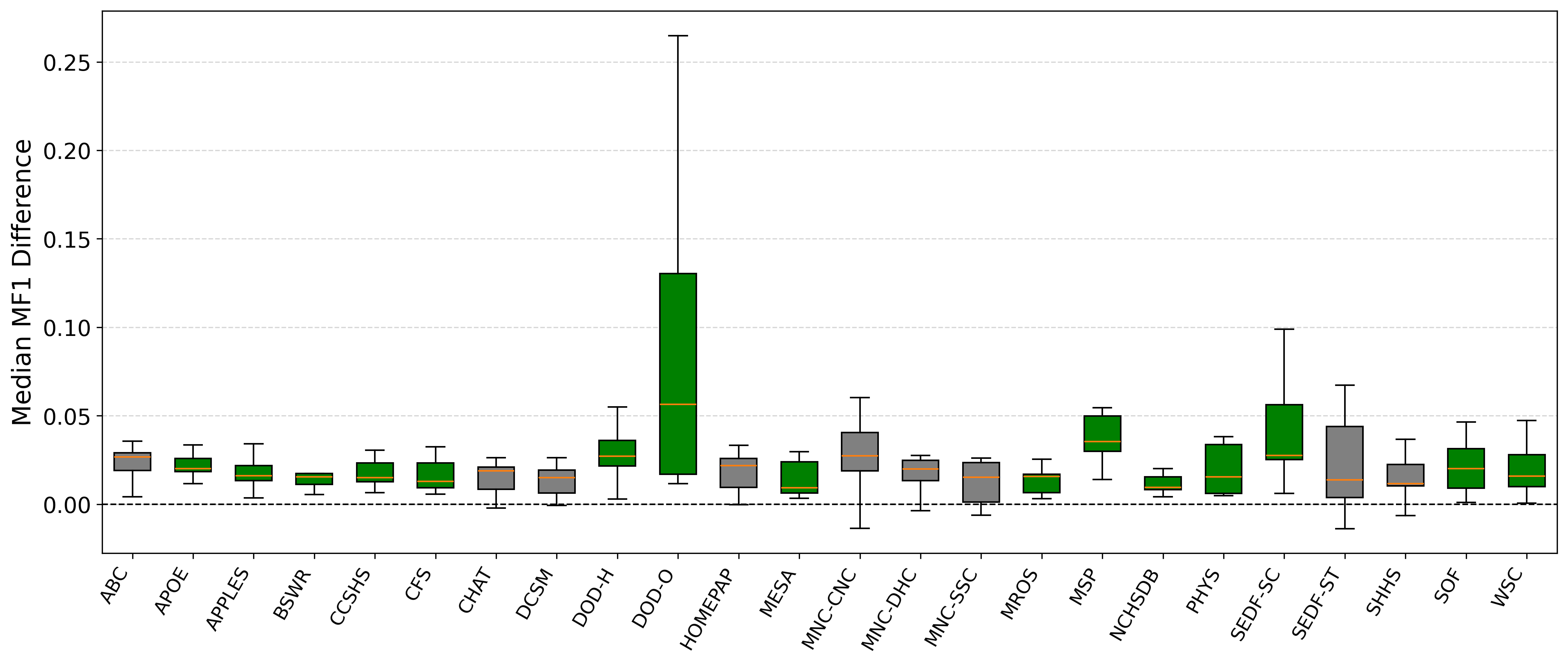}
    \caption{Distribution of the differences in median macro-F1 performance between SOMNUS and individual models (specific architecture and channel configuration) for each dataset. Each boxplot represents the variability in median performance differences across all model comparisons for a given dataset. Green boxes indicate datasets where SOMNUS consistently outperformed all individual models, while gray boxes indicate datasets with mixed outcomes. The dashed horizontal line at zero serves as a reference for equal performance. }
    \label{supp_fig:diff_SOMNUS_individual_models}
\end{figure}

\begin{figure}[h]
    \renewcommand{\figurename}{Supplementary Figure}
    \centering
    \includegraphics[width=1\linewidth]{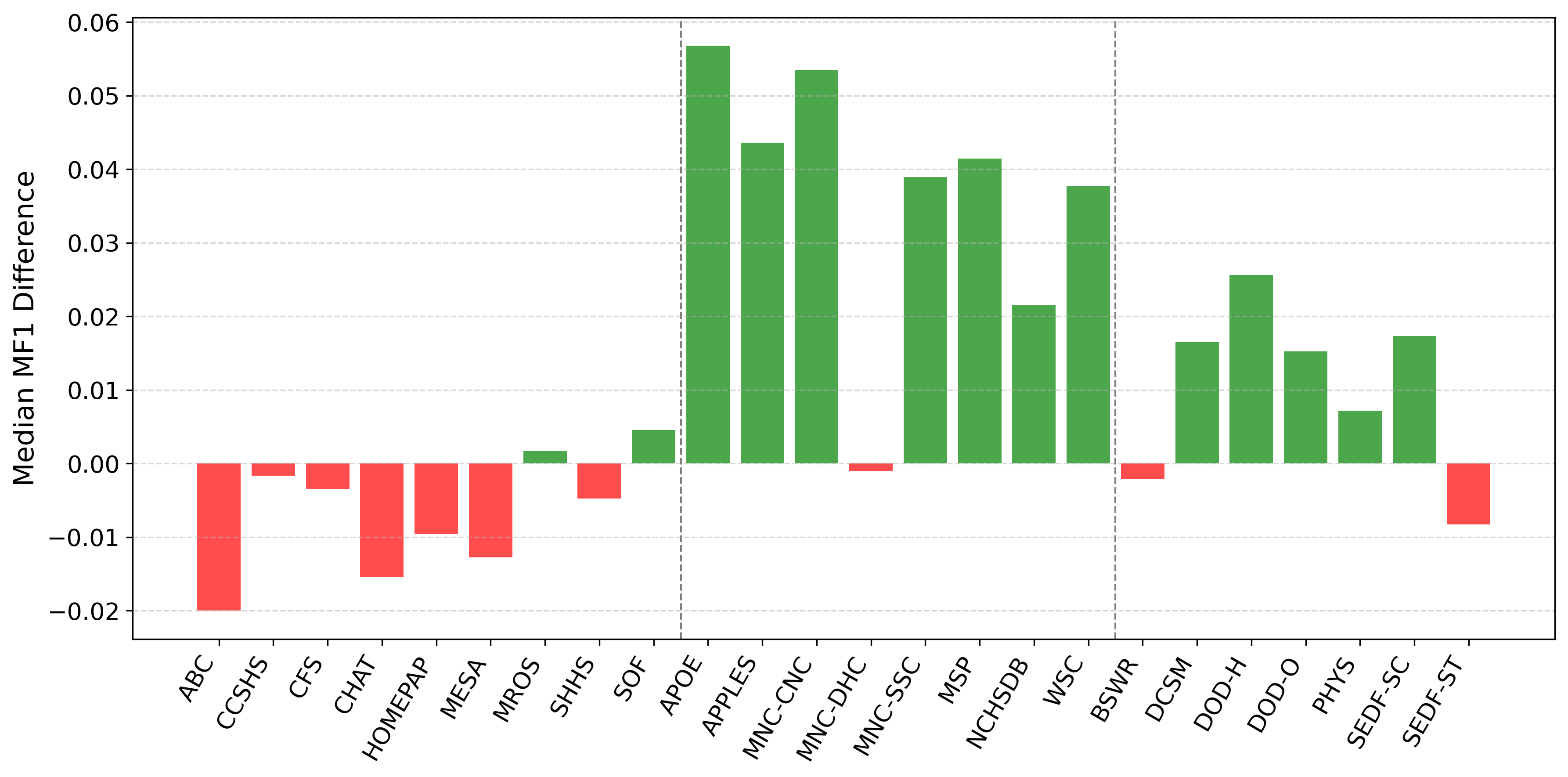}
    \caption{Difference in median macro‑F1 performance between two SOMNUS configurations: one trained on all 17 NSRR datasets, and another trained on a reduced subset excluding APOE, APPLES, MNC, MSP, NCHSDB, and WSC. Green bars indicate higher performance for the more comprehensively trained model, while red bars indicate decreased performance. Vertical dashed lines delineate: (from left to right) datasets that are in-domain for both models, datasets in-domain only for the version trained on the full dataset collection, and datasets that are out-of-domain for both models.}
    \label{supp_fig:diff_SOMNUS_22_24}
\end{figure}


\title[SLEEPYLAND]{SLEEPYLAND: trust begins with fair evaluation of automatic sleep staging models}


\author[1,2]{\fnm{Alvise} \sur{Dei Rossi}}\email{alvise.dei.rossi@usi.ch}

\author[2]{\fnm{Matteo} \sur{Metaldi}}\email{matteo.metaldi@supsi.ch}

\author[2,3]{\fnm{Michal} \sur{Bechny}}\email{michal.bechny@supsi.ch}

\author[4]{\fnm{Irina} \sur{Filchenko}}\email{irina.filchenko@insel.ch}

\author[4]{\fnm{Julia} \sur{van der Meer}}\email{julia.vandermeer@insel.ch}

\author[4]{\fnm{Markus H.} \sur{Schmidt}}\email{markus.schmidt@insel.ch}

\author[4]{\fnm{Claudio L.A.} \sur{Bassetti}}\email{claudio.bassetti@insel.ch}

\author[3,4]{\fnm{Athina} \sur{Tzovara}}\email{athina.tzovara@inf.unibe.ch}

\author[2]{\fnm{Francesca D.} \sur{Faraci}}\email{francesca.faraci@supsi.ch}

\author*[2,5]{\fnm{Luigi} \sur{Fiorillo}}\email{luigi.fiorillo@supsi.ch}

\affil[1]{\orgdiv{Faculty of informatics}, \orgname{Università della Svizzera Italiana}, \orgaddress{\street{Via Giuseppe Buffi 13}, \city{Lugano}, \postcode{6900}, \country{Switzerland}}}

\affil*[2]{\orgdiv{Institute of Digital Technologies for Personalized Healthcare $\vert$ MeDiTech}, Department of Innovative Technologies, \orgname{University of Applied Sciences and Arts of Southern Switzerland}, \orgaddress{\street{Via la Santa 1}, \city{Lugano}, \postcode{6962}, \country{Switzerland}}}

\affil[3]{\orgdiv{Institute of Informatics}, \orgname{University of Bern}, \orgaddress{\street{Neubr{\"u}ckstrasse 10}, \city{Bern}, \postcode{3012}, \country{Switzerland}}}

\affil[4]{\orgdiv{Sleep Wake Epilepsy Center $\vert$ NeuroTec},  Department of Neurology, \orgname{Inselspital, Bern University Hospital, University of Bern}, \orgaddress{\street{Freiburgstrasse}, \city{Bern}, \postcode{3010}, \country{Switzerland}}}

\affil*[5]{\orgdiv{Neurocenter of Southern Switzerland}, \orgname{Ente Ospedaliero Cantonale}, \orgaddress{\street{Via Tesserete 46}, \city{Lugano}, \postcode{6900}, \country{Switzerland}}}

\maketitle

\newpage

\section*{List of Supplementary Tables}

\begin{table}[h]
\renewcommand{\tablename}{Supplementary Table}
\caption{\textbf{EEG and EOG derivations available within each in-domain (ID) datasets.}}
\label{supp_tab:eeg_eog_channels}
\begin{center}
\begin{tabular}{lll}
Datasets & Channel type & Channel derivations \\
\hline
\noalign{\vskip 1mm} 
ABC & EEG & F3-M2, F4-M1, C3-M2, \\
& & C4-M1, O1-M2, O2-M1\\
& EOG & E1-M2, E2-M1\\ \hline
\noalign{\vskip 1mm}
APOE & EEG & C3-A1, C3-A2, C3-O1, C4-A1, C4-A2, F1-A2,\\
& & F2-C4, F2-T4, FP1-C3, FP1-C3, FP2-C4, FP2-C4,\\
& & Fz-A1, Fz-A2, O1-A2, O1-A2, O2-A1, T3-O1, T4-O2\\
& EOG & LOC-A2, ROC-A1, LOC, ROC\\ \hline
APPLES & EEG & C3-M2, C4-M1, O1-M2, O2-M1 \\
& EOG & LOC, ROC\\ \hline
CCSHS & EEG & C3-A2, C4-A1 \\
& EOG & LOC-A2, ROC-A1\\ \hline
CFS & EEG & C3-A2, C4-A1 \\
& EOG & LOC-A2, ROC-A1\\ \hline
CHAT & EEG & F3-M2, F4-M1, C3-M2, C4-M1, \\
& & T3-M2, T4-M1, O1-M2, O2-M1\\
& EOG & E1-M2, E2-M1, E1, E2\\ \hline
HOMEPAP & EEG & F4-M1, C4-M1, O2-M1, C3-M2, F3-M2, O1-M2,\\
& & F4, C4, O2, C3, F3, O1\\
& EOG & E1-M2, E2-M1, E1, E2, LOC, ROC, E1-E2\\ \hline
MESA & EEG & Fpz-Cz, Cz-Oz, C4-M1 \\
& EOG & E1-Fpz, E2-Fpz\\ \hline
\noalign{\vskip 1mm}
MNC & EEG & C3-M2, C3, C4-M1, C4, Cz, F3-M2, F3,\\
& & F4-M1, F4, O1-M2, O1, O2-M1, O2\\
& EOG & E1-M2 E1 E2-M1 E2 \\ \hline
MROS & EEG & C3-M2, C4-M1 \\
& EOG & E1-M2, E2-M1, E1, E2\\ \hline
\noalign{\vskip 1mm}
MSP & EEG & C3-M2, C4-M1, F3-M2, F4-M1, O1-M2, O2-M1\\
& EOG & LOC, ROC\\ \hline
NCHSDB & EEG & FP1, FP2, FZ, CZ, PZ, OZ, FPZ, P3-M2, P4-M1,\\ 
& & F3-M2, F4-M1, F4-M2, C3-M2, C4-M1, C4-M2,\\
& & T3-M2, T4-M1, O1-M2, O2-M1, F4, O1, O2\\
& EOG & E1-M2, E2-M1, E1-M1, E2-M2, E1, E2\\ \hline
SHHS & EEG & C4-A1, C3-A2 \\
& EOG & EOGL-PG1, EOGR-PG1\\ \hline
SOF & EEG & C3-A2, C4-A1 \\
& EOG & LOC-A2, ROC-A1\\ \hline
WSC & EEG & F3-M1, F3-M2, F3-AVG, F4-M1, F4-M2, F4-AVG, \\
& & Fz-M1, Fz-M2, Fz-AVG, Cz-M1, Cz-M2, Cz-AVG, \\
& & C3-M1, C3-M2, C3-AVG, C4-M1, C4-M2, C4-AVG, \\
& & Pz-M1, Pz-M2, Pz-AVG, Pz-Cz, O1-M1, O1-M2, \\
& & O1-AVG, O2-M1, O2-M2, O2-AVG\\
& EOG & E1, E2\\ \hline
\end{tabular}
\end{center}
\end{table}

\begin{table}[h]
\renewcommand{\tablename}{Supplementary Table}
\caption{\textbf{Data split on the in-domain (ID) datasets.} We report the total number of recordings, and the number of recordings used to train, validate and test all the models in SLEEPYLAND.\\}
\label{supp_tab:data_split_ID}
\begin{center}
\begin{tabular}{lllll}
Datasets   & Recordings & Train & Valid & Test \\ \hline \noalign{\vskip 1mm}
ABC        &   132  &  97  & 13  & 22  \\ \hline \noalign{\vskip 1mm}
APOE       &   712  &  562 & 50  & 100 \\ \hline \noalign{\vskip 1mm}
APPLES     &  1094  &  944 & 50  & 100 \\ \hline \noalign{\vskip 1mm}
CCSHS      &   515  & 387  & 50  & 78  \\ \hline \noalign{\vskip 1mm}
CFS        &   730  & 545  & 98  & 87  \\ \hline \noalign{\vskip 1mm}
CHAT       &  1638  & 1439 & 60  & 139 \\ \hline \noalign{\vskip 1mm}
HOMEPAP       &   246  & 184  & 25  & 37  \\ \hline \noalign{\vskip 1mm}
MESA       &  2056  & 1906 & 50  & 100 \\ \hline \noalign{\vskip 1mm}
MNC\text{-}CNC   &    78  &  58  & 8   & 12  \\ \hline \noalign{\vskip 1mm}
MNC\text{-}DHC   &    83  &  61  & 9   & 13  \\ \hline \noalign{\vskip 1mm}
MNC\text{-}SSC   &   767  & 617  & 50  & 100 \\ \hline \noalign{\vskip 1mm}
MROS       &  3930  & 3735 & 67  & 128 \\ \hline \noalign{\vskip 1mm}
MSP        &   105  & 78   & 11  & 16  \\ \hline \noalign{\vskip 1mm}
NCHSDB     &  3950  & 3789 & 55  & 106    \\ \hline \noalign{\vskip 1mm}
SHHS       &  8444  & 8223 & 75  & 146 \\ \hline \noalign{\vskip 1mm}
SOF        &   453  & 339  & 46  & 68  \\ \hline \noalign{\vskip 1mm}
WSC        &  2569  & 2222 & 119 & 228 \\ \hline
\end{tabular}
\end{center}
\end{table}

\begin{table}[h]
\renewcommand{\tablename}{Supplementary Table}
\caption{\textbf{SOMNUS dataset‐wise performance overview.} MF1, Accuracy and Cohen’s $\kappa$, plus Class‐Wise F1 score computed across all test recordings in each dataset (majority voting across channel derivations).}
\label{supp_tab:epoch_wise_best}
\begin{center}
\scriptsize
\begin{tabular}{lllllllll}
\hline
\noalign{\vskip 1mm}
\textbf{ID Dataset} & \textbf{MF1} & \textbf{Acc} & \textbf{$\kappa$} & \textbf{W} & \textbf{N1} & \textbf{N2} & \textbf{N3} & \textbf{REM} \\
\hline
\noalign{\vskip 1mm}
ABC      & $0.825$ & $0.854$ & $0.796$ & $0.907$ & $0.629$ & $0.878$ & $0.775$ & $0.934$ \\
\hline
\noalign{\vskip 1mm}
APOE     & $0.752$ & $0.822$ & $0.739$ & $0.892$ & $0.448$ & $0.862$ & $0.671$ & $0.888$ \\
\hline
\noalign{\vskip 1mm}
APPLES   & $0.731$ & $0.835$ & $0.761$ & $0.923$ & $0.516$ & $0.868$ & $0.443$ & $0.906$ \\
\hline
\noalign{\vskip 1mm}
CCSHS    & $0.875$ & $0.927$ & $0.899$ & $0.974$ & $0.655$ & $0.921$ & $0.887$ & $0.939$ \\
\hline
\noalign{\vskip 1mm}
CFS      & $0.827$ & $0.906$ & $0.865$ & $0.964$ & $0.522$ & $0.895$ & $0.836$ & $0.916$ \\
\hline
\noalign{\vskip 1mm}
CHAT     & $0.848$ & $0.890$ & $0.855$ & $0.969$ & $0.600$ & $0.866$ & $0.900$ & $0.905$ \\
\hline
\noalign{\vskip 1mm}
HOMEPAP  & $0.775$ & $0.830$ & $0.766$ & $0.910$ & $0.414$ & $0.841$ & $0.798$ & $0.911$ \\
\hline
\end{tabular}
\end{center}
\end{table}

\newpage

\begin{landscape}
\begin{table}[h]
\renewcommand{\tablename}{Supplementary Table}
\caption{\textbf{SOMNUS$_{\text{U-Sleep}}$ benchmarking.} MF1, Accuracy and Cohen’s $\kappa$, plus Class‐Wise F1 score on recording‐level for U‐Sleep considering both single and multi‐channel configurations.}
\label{supp_tab:results_usleep_all}
\begin{center}
\begin{tabular}{lllllllll}
\hline
\noalign{\vskip 1mm}
\textbf{ID Dataset} & \textbf{MF1} & \textbf{Acc} & \textbf{$\kappa$} & \textbf{W} & \textbf{N1} & \textbf{N2} & \textbf{N3} & \textbf{REM} \\
\hline
\noalign{\vskip 1mm}
ABC      & $0.776\pm0.094$ & $0.847\pm0.051$ & $0.773\pm0.063$ & $0.888\pm0.047$ & $0.594\pm0.084$ & $0.862\pm0.072$ & $0.621\pm0.306$ & $0.895\pm0.201$ \\
CCSHS    & $0.865\pm0.052$ & $0.923\pm0.038$ & $0.892\pm0.053$ & $0.969\pm0.022$ & $0.624\pm0.151$ & $0.916\pm0.050$ & $0.878\pm0.099$ & $0.936\pm0.035$ \\
CFS      & $0.813\pm0.081$ & $0.904\pm0.060$ & $0.854\pm0.098$ & $0.963\pm0.037$ & $0.536\pm0.153$ & $0.885\pm0.099$ & $0.777\pm0.222$ & $0.905\pm0.106$ \\
CHAT     & $0.830\pm0.044$ & $0.883\pm0.047$ & $0.842\pm0.063$ & $0.955\pm0.033$ & $0.551\pm0.114$ & $0.853\pm0.080$ & $0.895\pm0.072$ & $0.897\pm0.060$ \\
HOMEPAP  & $0.750\pm0.084$ & $0.829\pm0.084$ & $0.747\pm0.133$ & $0.901\pm0.082$ & $0.406\pm0.151$ & $0.824\pm0.095$ & $0.750\pm0.228$ & $0.894\pm0.125$ \\
MESA     & $0.743\pm0.088$ & $0.866\pm0.074$ & $0.798\pm0.115$ & $0.945\pm0.075$ & $0.492\pm0.137$ & $0.845\pm0.109$ & $0.528\pm0.293$ & $0.901\pm0.077$ \\
MROS     & $0.749\pm0.085$ & $0.894\pm0.041$ & $0.831\pm0.067$ & $0.957\pm0.037$ & $0.450\pm0.159$ & $0.875\pm0.063$ & $0.575\pm0.281$ & $0.880\pm0.142$ \\
\hline
\end{tabular}
\end{center}
\end{table}
\end{landscape}

\begin{landscape}
\begin{table}[h]
\renewcommand{\tablename}{Supplementary Table}
\caption{\textbf{SOMNUS$_{\text{DeepResNet}}$ benchmarking.} MF1, Accuracy and Cohen’s $\kappa$, plus Class‐Wise F1 score on recording‐level for DeepResNet considering both single and multi‐channel configurations.}
\label{supp_tab:results_deepresnet_all}
\begin{center}
\begin{tabular}{lllllllll}
\hline
\noalign{\vskip 1mm}
\textbf{ID Dataset} & \textbf{MF1} & \textbf{Acc} & \textbf{$\kappa$} & \textbf{W} & \textbf{N1} & \textbf{N2} & \textbf{N3} & \textbf{REM} \\
\hline
\noalign{\vskip 1mm}
ABC      & $0.776\pm0.094$ & $0.844\pm0.051$ & $0.769\pm0.062$ & $0.887\pm0.046$ & $0.596\pm0.088$ & $0.857\pm0.077$ & $0.633\pm0.298$ & $0.886\pm0.200$ \\
CCSHS    & $0.861\pm0.049$ & $0.921\pm0.037$ & $0.889\pm0.051$ & $0.970\pm0.018$ & $0.613\pm0.141$ & $0.914\pm0.049$ & $0.878\pm0.091$ & $0.930\pm0.039$ \\
CFS      & $0.807\pm0.085$ & $0.903\pm0.062$ & $0.852\pm0.101$ & $0.962\pm0.040$ & $0.509\pm0.180$ & $0.885\pm0.099$ & $0.770\pm0.236$ & $0.912\pm0.082$ \\
CHAT     & $0.827\pm0.048$ & $0.880\pm0.048$ & $0.838\pm0.065$ & $0.953\pm0.039$ & $0.546\pm0.125$ & $0.852\pm0.080$ & $0.897\pm0.072$ & $0.887\pm0.063$ \\
HOMEPAP  & $0.734\pm0.135$ & $0.809\pm0.127$ & $0.722\pm0.177$ & $0.888\pm0.117$ & $0.402\pm0.155$ & $0.794\pm0.158$ & $0.720\pm0.277$ & $0.884\pm0.170$ \\
MESA     & $0.743\pm0.090$ & $0.869\pm0.069$ & $0.801\pm0.110$ & $0.950\pm0.061$ & $0.504\pm0.141$ & $0.848\pm0.108$ & $0.512\pm0.305$ & $0.898\pm0.072$ \\
MROS     & $0.740\pm0.085$ & $0.893\pm0.041$ & $0.829\pm0.066$ & $0.955\pm0.036$ & $0.414\pm0.157$ & $0.875\pm0.058$ & $0.576\pm0.290$ & $0.874\pm0.144$ \\
\hline
\end{tabular}
\end{center}
\end{table}
\end{landscape}

\begin{landscape}
\begin{table}[h]
\renewcommand{\tablename}{Supplementary Table}
\caption{\textbf{SOMNUS$_{\text{SleepTransformer}}$ benchmarking.} MF1, Accuracy and Cohen’s $\kappa$, plus Class‐Wise F1 score on recording‐level for SleepTransformer considering both single and multi‐channel configurations.}
\label{supp_tab:results_sleeptransformer_all}
\begin{center}
\begin{tabular}{lllllllll}
\hline
\noalign{\vskip 1mm}
\textbf{ID Dataset} & \textbf{MF1} & \textbf{Acc} & \textbf{$\kappa$} & \textbf{W} & \textbf{N1} & \textbf{N2} & \textbf{N3} & \textbf{REM} \\
\hline
\noalign{\vskip 1mm}
ABC      & $0.790\pm0.084$ & $0.851\pm0.042$ & $0.780\pm0.048$ & $0.892\pm0.047$ & $0.606\pm0.090$ & $0.862\pm0.066$ & $0.683\pm0.272$ & $0.891\pm0.200$ \\
CCSHS    & $0.866\pm0.049$ & $0.923\pm0.033$ & $0.893\pm0.047$ & $0.973\pm0.016$ & $0.629\pm0.148$ & $0.915\pm0.047$ & $0.876\pm0.097$ & $0.935\pm0.038$ \\
CFS      & $0.811\pm0.081$ & $0.901\pm0.062$ & $0.850\pm0.101$ & $0.962\pm0.039$ & $0.542\pm0.154$ & $0.880\pm0.101$ & $0.764\pm0.228$ & $0.910\pm0.085$ \\
CHAT     & $0.848\pm0.039$ & $0.889\pm0.042$ & $0.850\pm0.057$ & $0.963\pm0.030$ & $0.622\pm0.097$ & $0.855\pm0.076$ & $0.897\pm0.071$ & $0.903\pm0.054$ \\
HOMEPAP  & $0.760\pm0.082$ & $0.833\pm0.079$ & $0.750\pm0.128$ & $0.910\pm0.073$ & $0.452\pm0.143$ & $0.824\pm0.097$ & $0.734\pm0.234$ & $0.900\pm0.119$ \\
MESA     & $0.751\pm0.091$ & $0.873\pm0.069$ & $0.807\pm0.110$ & $0.953\pm0.061$ & $0.509\pm0.142$ & $0.850\pm0.105$ & $0.540\pm0.297$ & $0.901\pm0.079$ \\
MROS     & $0.751\pm0.086$ & $0.896\pm0.037$ & $0.835\pm0.060$ & $0.957\pm0.036$ & $0.464\pm0.162$ & $0.877\pm0.054$ & $0.570\pm0.289$ & $0.875\pm0.150$ \\
\hline
\end{tabular}
\end{center}
\end{table}
\end{landscape}

\begin{landscape}
\begin{table}[h]
\renewcommand{\tablename}{Supplementary Table}
\caption{\textbf{SOMNUS benchmarking.} MF1, Accuracy and Cohen’s $\kappa$, plus Class‐Wise F1 score on recording‐level for soft unweighted voting ensemble considering all deep learning scorers in SLEEPYLAND.}
\label{supp_tab:results_somnus}
\begin{center}
\begin{tabular}{lllllllll}
\hline
\noalign{\vskip 1mm}
\textbf{ID Dataset} & \textbf{MF1} & \textbf{Acc} & \textbf{$\kappa$} & \textbf{W} & \textbf{N1} & \textbf{N2} & \textbf{N3} & \textbf{REM} \\
\hline
\noalign{\vskip 1mm}
ABC      & $0.789\pm0.087$ & $0.853\pm0.046$ & $0.783\pm0.054$ & $0.895\pm0.045$ & $0.605\pm0.087$ & $0.866\pm0.068$ & $0.666\pm0.288$ & $0.895\pm0.201$ \\
CCSHS    & $0.872\pm0.049$ & $0.926\pm0.036$ & $0.896\pm0.050$ & $0.973\pm0.017$ & $0.649\pm0.141$ & $0.919\pm0.048$ & $0.881\pm0.094$ & $0.938\pm0.035$ \\
CFS      & $0.817\pm0.080$ & $0.906\pm0.061$ & $0.857\pm0.099$ & $0.963\pm0.039$ & $0.539\pm0.162$ & $0.888\pm0.099$ & $0.781\pm0.223$ & $0.916\pm0.082$ \\
CHAT     & $0.842\pm0.043$ & $0.889\pm0.045$ & $0.849\pm0.061$ & $0.960\pm0.031$ & $0.589\pm0.114$ & $0.858\pm0.078$ & $0.899\pm0.072$ & $0.901\pm0.057$ \\
HOMEPAP  & $0.755\pm0.083$ & $0.833\pm0.082$ & $0.751\pm0.131$ & $0.908\pm0.078$ & $0.416\pm0.155$ & $0.827\pm0.091$ & $0.744\pm0.253$ & $0.904\pm0.111$ \\
MESA     & $0.752\pm0.092$ & $0.875\pm0.069$ & $0.809\pm0.112$ & $0.953\pm0.061$ & $0.510\pm0.146$ & $0.852\pm0.111$ & $0.536\pm0.297$ & $0.906\pm0.071$ \\
MROS     & $0.752\pm0.084$ & $0.898\pm0.038$ & $0.838\pm0.061$ & $0.958\pm0.035$ & $0.459\pm0.161$ & $0.880\pm0.055$ & $0.575\pm0.291$ & $0.880\pm0.143$ \\
\hline
\end{tabular}
\end{center}
\end{table}
\end{landscape}

\begin{landscape}
\begin{table}[h]
\renewcommand{\tablename}{Supplementary Table}
\caption{\textbf{U-Sleep$_{\text{EEG+EOG}}$ benchmarking.} MF1, Accuracy and Cohen’s $\kappa$, plus Class‐Wise F1 score on recording‐level for U‐Sleep using both EEG and EOG channel derivations.}
\label{supp_tab:results_usleep_eegeog}
\begin{center}
\begin{tabular}{lllllllll}
\hline
\noalign{\vskip 1mm}
\textbf{ID Dataset} & \textbf{MF1} & \textbf{Acc} & \textbf{$\kappa$} & \textbf{W} & \textbf{N1} & \textbf{N2} & \textbf{N3} & \textbf{REM} \\
\hline
\noalign{\vskip 1mm}
ABC      & $0.785\pm0.092$ & $0.848\pm0.050$ & $0.776\pm0.061$ & $0.887\pm0.046$ & $0.612\pm0.079$ & $0.863\pm0.069$ & $0.648\pm0.285$ & $0.895\pm0.201$ \\
CCSHS    & $0.859\pm0.050$ & $0.919\pm0.038$ & $0.886\pm0.053$ & $0.965\pm0.033$ & $0.611\pm0.141$ & $0.912\pm0.048$ & $0.879\pm0.093$ & $0.930\pm0.038$ \\
CFS      & $0.809\pm0.079$ & $0.901\pm0.060$ & $0.851\pm0.097$ & $0.962\pm0.037$ & $0.520\pm0.156$ & $0.883\pm0.097$ & $0.780\pm0.215$ & $0.902\pm0.110$ \\
CHAT     & $0.822\pm0.047$ & $0.881\pm0.047$ & $0.838\pm0.063$ & $0.952\pm0.036$ & $0.520\pm0.121$ & $0.851\pm0.078$ & $0.896\pm0.071$ & $0.892\pm0.063$ \\
HOMEPAP  & $0.739\pm0.100$ & $0.821\pm0.092$ & $0.737\pm0.143$ & $0.890\pm0.095$ & $0.417\pm0.152$ & $0.811\pm0.117$ & $0.726\pm0.260$ & $0.878\pm0.167$ \\
MESA     & $0.742\pm0.089$ & $0.859\pm0.079$ & $0.788\pm0.117$ & $0.937\pm0.095$ & $0.511\pm0.128$ & $0.838\pm0.103$ & $0.531\pm0.299$ & $0.889\pm0.095$ \\
MROS     & $0.752\pm0.080$ & $0.891\pm0.040$ & $0.826\pm0.067$ & $0.955\pm0.038$ & $0.445\pm0.150$ & $0.870\pm0.067$ & $0.605\pm0.250$ & $0.879\pm0.140$ \\
\hline
\end{tabular}
\end{center}
\end{table}
\end{landscape}

\begin{landscape}
\begin{table}[h]
\renewcommand{\tablename}{Supplementary Table}
\caption{\textbf{DeepResNet$_{\text{EEG+EOG}}$ benchmarking.} MF1, Accuracy and Cohen’s $\kappa$, plus Class‐Wise F1 score on recording‐level for DeepResNet using both EEG and EOG channel derivations.}
\label{supp_tab:results_deepresnet_eegeog}
\begin{center}
\begin{tabular}{lllllllll}
\hline
\noalign{\vskip 1mm}
\textbf{ID Dataset} & \textbf{MF1} & \textbf{Acc} & \textbf{$\kappa$} & \textbf{W} & \textbf{N1} & \textbf{N2} & \textbf{N3} & \textbf{REM} \\
\hline
\noalign{\vskip 1mm}
ABC      & $0.764\pm0.096$ & $0.834\pm0.053$ & $0.756\pm0.065$ & $0.879\pm0.050$ & $0.569\pm0.097$ & $0.849\pm0.082$ & $0.616\pm0.292$ & $0.884\pm0.199$ \\
CCSHS    & $0.856\pm0.050$ & $0.918\pm0.039$ & $0.884\pm0.054$ & $0.969\pm0.019$ & $0.600\pm0.142$ & $0.910\pm0.050$ & $0.875\pm0.089$ & $0.925\pm0.045$ \\
CFS      & $0.804\pm0.085$ & $0.901\pm0.061$ & $0.849\pm0.098$ & $0.960\pm0.039$ & $0.501\pm0.179$ & $0.883\pm0.096$ & $0.775\pm0.222$ & $0.901\pm0.115$ \\
CHAT     & $0.820\pm0.048$ & $0.877\pm0.047$ & $0.833\pm0.063$ & $0.950\pm0.040$ & $0.528\pm0.129$ & $0.847\pm0.080$ & $0.896\pm0.071$ & $0.880\pm0.065$ \\
HOMEPAP  & $0.726\pm0.138$ & $0.803\pm0.130$ & $0.715\pm0.182$ & $0.885\pm0.119$ & $0.382\pm0.157$ & $0.785\pm0.165$ & $0.731\pm0.275$ & $0.875\pm0.191$ \\
MESA     & $0.733\pm0.087$ & $0.861\pm0.077$ & $0.790\pm0.113$ & $0.942\pm0.086$ & $0.487\pm0.135$ & $0.841\pm0.103$ & $0.500\pm0.309$ & $0.889\pm0.078$ \\
MROS     & $0.736\pm0.084$ & $0.889\pm0.042$ & $0.822\pm0.068$ & $0.951\pm0.036$ & $0.403\pm0.148$ & $0.870\pm0.060$ & $0.579\pm0.278$ & $0.870\pm0.144$ \\
\hline
\end{tabular}
\end{center}
\end{table}
\end{landscape}

\begin{landscape}
\begin{table}[h]
\renewcommand{\tablename}{Supplementary Table}
\caption{\textbf{SleepTransformer$_{\text{EEG+EOG}}$ benchmarking.} MF1, Accuracy and Cohen’s $\kappa$, plus Class‐Wise F1 score on recording‐level for SleepTransformer using both EEG and EOG channel derivations.}
\label{supp_tab:results_sleeptransformer_eegeog}
\begin{center}
\begin{tabular}{lllllllll}
\hline
\noalign{\vskip 1mm}
\textbf{ID Dataset} & \textbf{MF1} & \textbf{Acc} & \textbf{$\kappa$} & \textbf{W} & \textbf{N1} & \textbf{N2} & \textbf{N3} & \textbf{REM} \\
\hline
\noalign{\vskip 1mm}
ABC      & $0.790\pm0.084$ & $0.851\pm0.042$ & $0.780\pm0.048$ & $0.892\pm0.047$ & $0.606\pm0.090$ & $0.862\pm0.066$ & $0.683\pm0.272$ & $0.891\pm0.200$ \\
CCSHS    & $0.866\pm0.049$ & $0.923\pm0.033$ & $0.893\pm0.047$ & $0.973\pm0.016$ & $0.629\pm0.148$ & $0.915\pm0.047$ & $0.876\pm0.097$ & $0.935\pm0.038$ \\
CFS      & $0.811\pm0.081$ & $0.901\pm0.062$ & $0.850\pm0.101$ & $0.962\pm0.039$ & $0.542\pm0.154$ & $0.880\pm0.101$ & $0.764\pm0.228$ & $0.910\pm0.085$ \\
CHAT     & $0.848\pm0.039$ & $0.889\pm0.042$ & $0.850\pm0.057$ & $0.963\pm0.030$ & $0.622\pm0.097$ & $0.855\pm0.076$ & $0.897\pm0.071$ & $0.903\pm0.054$ \\
HOMEPAP  & $0.760\pm0.082$ & $0.833\pm0.079$ & $0.750\pm0.128$ & $0.910\pm0.073$ & $0.452\pm0.143$ & $0.824\pm0.097$ & $0.734\pm0.234$ & $0.900\pm0.119$ \\
MESA     & $0.751\pm0.091$ & $0.873\pm0.069$ & $0.807\pm0.110$ & $0.953\pm0.061$ & $0.509\pm0.142$ & $0.850\pm0.105$ & $0.540\pm0.297$ & $0.901\pm0.079$ \\
MROS     & $0.751\pm0.086$ & $0.896\pm0.037$ & $0.835\pm0.060$ & $0.957\pm0.036$ & $0.464\pm0.162$ & $0.877\pm0.054$ & $0.570\pm0.289$ & $0.875\pm0.150$ \\
\hline
\end{tabular}
\end{center}
\end{table}
\end{landscape}

\begin{landscape}
\begin{table}[h]
\renewcommand{\tablename}{Supplementary Table}
\caption{\textbf{SOMNUS$_{\text{EEG+EOG}}$ benchmarking.} MF1, Accuracy and Cohen’s $\kappa$, plus Class‐Wise F1 score on recording‐level for soft unweighted voting ensemble considering all models using both EEG and EOG channel derivations.}
\label{supp_tab:results_suv_eegeog}
\begin{center}
\begin{tabular}{lllllllll}
\hline
\noalign{\vskip 1mm}
\textbf{ID Dataset} & \textbf{MF1} & \textbf{Acc} & \textbf{$\kappa$} & \textbf{W} & \textbf{N1} & \textbf{N2} & \textbf{N3} & \textbf{REM} \\
\hline
\noalign{\vskip 1mm}
ABC      & $0.785\pm0.088$ & $0.850\pm0.046$ & $0.778\pm0.055$ & $0.890\pm0.048$ & $0.596\pm0.085$ & $0.864\pm0.067$ & $0.667\pm0.277$ & $0.893\pm0.200$ \\
CCSHS    & $0.871\pm0.047$ & $0.925\pm0.035$ & $0.895\pm0.049$ & $0.972\pm0.017$ & $0.650\pm0.139$ & $0.918\pm0.047$ & $0.880\pm0.091$ & $0.936\pm0.037$ \\
CFS      & $0.818\pm0.079$ & $0.907\pm0.059$ & $0.858\pm0.097$ & $0.964\pm0.038$ & $0.539\pm0.157$ & $0.889\pm0.096$ & $0.787\pm0.207$ & $0.913\pm0.089$ \\
CHAT     & $0.838\pm0.045$ & $0.887\pm0.045$ & $0.847\pm0.061$ & $0.960\pm0.031$ & $0.572\pm0.122$ & $0.857\pm0.076$ & $0.899\pm0.071$ & $0.901\pm0.057$ \\
HOMEPAP  & $0.745\pm0.102$ & $0.826\pm0.089$ & $0.742\pm0.141$ & $0.899\pm0.087$ & $0.413\pm0.156$ & $0.822\pm0.095$ & $0.730\pm0.271$ & $0.883\pm0.167$ \\
MESA     & $0.749\pm0.088$ & $0.871\pm0.072$ & $0.805\pm0.113$ & $0.949\pm0.073$ & $0.503\pm0.138$ & $0.850\pm0.105$ & $0.539\pm0.300$ & $0.903\pm0.072$ \\
MROS     & $0.755\pm0.083$ & $0.898\pm0.039$ & $0.836\pm0.063$ & $0.957\pm0.035$ & $0.452\pm0.156$ & $0.879\pm0.059$ & $0.597\pm0.275$ & $0.881\pm0.143$ \\
\hline
\end{tabular}
\end{center}
\end{table}
\end{landscape}

\begin{landscape}
\begin{table}[h]
\renewcommand{\tablename}{Supplementary Table}
\caption{\textbf{U-Sleep$_{\text{EEG}}$ benchmarking.} MF1, Accuracy and Cohen’s $\kappa$, plus Class‐Wise F1 score on recording‐level for U‐Sleep using EEG channel derivations.}
\label{supp_tab:results_usleep_eeg}
\begin{center}
\begin{tabular}{lllllllll}
\hline
\noalign{\vskip 1mm}
\textbf{ID Dataset} & \textbf{MF1} & \textbf{Acc} & \textbf{$\kappa$} & \textbf{W} & \textbf{N1} & \textbf{N2} & \textbf{N3} & \textbf{REM} \\
\hline
\noalign{\vskip 1mm}
ABC      & $0.759\pm0.093$ & $0.838\pm0.051$ & $0.758\pm0.060$ & $0.882\pm0.048$ & $0.566\pm0.095$ & $0.853\pm0.076$ & $0.597\pm0.323$ & $0.878\pm0.201$ \\
CCSHS    & $0.855\pm0.060$ & $0.916\pm0.048$ & $0.882\pm0.066$ & $0.964\pm0.039$ & $0.607\pm0.154$ & $0.910\pm0.057$ & $0.869\pm0.108$ & $0.927\pm0.049$ \\
CFS      & $0.795\pm0.091$ & $0.893\pm0.063$ & $0.838\pm0.103$ & $0.959\pm0.040$ & $0.527\pm0.161$ & $0.875\pm0.101$ & $0.729\pm0.229$ & $0.883\pm0.137$ \\
CHAT     & $0.820\pm0.047$ & $0.873\pm0.050$ & $0.828\pm0.067$ & $0.946\pm0.040$ & $0.545\pm0.120$ & $0.845\pm0.087$ & $0.883\pm0.076$ & $0.883\pm0.067$ \\
HOMEPAP  & $0.742\pm0.070$ & $0.826\pm0.082$ & $0.742\pm0.130$ & $0.897\pm0.084$ & $0.375\pm0.132$ & $0.820\pm0.095$ & $0.757\pm0.227$ & $0.891\pm0.120$ \\
MESA     & $0.718\pm0.098$ & $0.856\pm0.078$ & $0.780\pm0.122$ & $0.936\pm0.092$ & $0.453\pm0.153$ & $0.841\pm0.101$ & $0.483\pm0.275$ & $0.868\pm0.123$ \\
MROS     & $0.738\pm0.086$ & $0.887\pm0.047$ & $0.821\pm0.073$ & $0.954\pm0.041$ & $0.448\pm0.161$ & $0.869\pm0.062$ & $0.558\pm0.277$ & $0.856\pm0.165$ \\
\hline
\end{tabular}
\end{center}
\end{table}
\end{landscape}

\begin{landscape}
\begin{table}[h]
\renewcommand{\tablename}{Supplementary Table}
\caption{\textbf{DeepResNet$_{\text{EEG}}$ benchmarking.} MF1, Accuracy and Cohen’s $\kappa$, plus Class‐Wise F1 score on recording‐level for DeepResNet using EEG channel derivations.}
\label{supp_tab:results_deepresnet_eeg}
\begin{center}
\begin{tabular}{lllllllll}
\hline
\noalign{\vskip 1mm}
\textbf{ID Dataset} & \textbf{MF1} & \textbf{Acc} & \textbf{$\kappa$} & \textbf{W} & \textbf{N1} & \textbf{N2} & \textbf{N3} & \textbf{REM} \\
\hline
\noalign{\vskip 1mm}
ABC      & $0.785\pm0.095$ & $0.846\pm0.049$ & $0.773\pm0.061$ & $0.890\pm0.047$ & $0.615\pm0.103$ & $0.858\pm0.071$ & $0.667\pm0.261$ & $0.882\pm0.200$ \\
CCSHS    & $0.857\pm0.051$ & $0.920\pm0.038$ & $0.887\pm0.052$ & $0.969\pm0.023$ & $0.599\pm0.148$ & $0.914\pm0.049$ & $0.877\pm0.090$ & $0.926\pm0.044$ \\
CFS      & $0.802\pm0.088$ & $0.898\pm0.063$ & $0.845\pm0.101$ & $0.959\pm0.041$ & $0.516\pm0.159$ & $0.881\pm0.098$ & $0.759\pm0.239$ & $0.893\pm0.120$ \\
CHAT     & $0.827\pm0.044$ & $0.878\pm0.047$ & $0.834\pm0.063$ & $0.952\pm0.036$ & $0.556\pm0.116$ & $0.849\pm0.078$ & $0.892\pm0.073$ & $0.884\pm0.064$ \\
HOMEPAP  & $0.754\pm0.085$ & $0.829\pm0.082$ & $0.748\pm0.133$ & $0.904\pm0.079$ & $0.421\pm0.144$ & $0.817\pm0.109$ & $0.756\pm0.240$ & $0.902\pm0.078$ \\
MESA     & $0.731\pm0.091$ & $0.860\pm0.074$ & $0.788\pm0.112$ & $0.939\pm0.087$ & $0.483\pm0.148$ & $0.844\pm0.099$ & $0.500\pm0.298$ & $0.880\pm0.090$ \\
MROS     & $0.738\pm0.085$ & $0.888\pm0.046$ & $0.821\pm0.074$ & $0.953\pm0.039$ & $0.422\pm0.153$ & $0.870\pm0.063$ & $0.581\pm0.275$ & $0.855\pm0.167$ \\
\hline
\end{tabular}
\end{center}
\end{table}
\end{landscape}

\begin{landscape}
\begin{table}[h]
\renewcommand{\tablename}{Supplementary Table}
\caption{\textbf{SleepTransformer$_{\text{EEG}}$ benchmarking.} MF1, Accuracy and Cohen’s $\kappa$, plus Class‐Wise F1 score on recording‐level for SleepTransformer using EEG channel derivations.}
\label{supp_tab:results_sleeptransformer_eeg}
\begin{center}
\begin{tabular}{lllllllll}
\hline
\noalign{\vskip 1mm}
\textbf{ID Dataset} & \textbf{MF1} & \textbf{Acc} & \textbf{$\kappa$} & \textbf{W} & \textbf{N1} & \textbf{N2} & \textbf{N3} & \textbf{REM} \\
\hline
\noalign{\vskip 1mm}
ABC      & $0.789\pm0.065$ & $0.846\pm0.046$ & $0.772\pm0.054$ & $0.893\pm0.049$ & $0.594\pm0.097$ & $0.858\pm0.067$ & $0.682\pm0.246$ & $0.904\pm0.119$ \\
CCSHS    & $0.861\pm0.049$ & $0.919\pm0.036$ & $0.886\pm0.050$ & $0.971\pm0.019$ & $0.623\pm0.143$ & $0.911\pm0.049$ & $0.875\pm0.089$ & $0.925\pm0.045$ \\
CFS      & $0.803\pm0.081$ & $0.897\pm0.061$ & $0.843\pm0.100$ & $0.958\pm0.041$ & $0.532\pm0.153$ & $0.878\pm0.094$ & $0.749\pm0.227$ & $0.902\pm0.091$ \\
CHAT     & $0.838\pm0.042$ & $0.883\pm0.044$ & $0.841\pm0.059$ & $0.961\pm0.029$ & $0.594\pm0.108$ & $0.850\pm0.074$ & $0.892\pm0.071$ & $0.895\pm0.059$ \\
HOMEPAP  & $0.744\pm0.085$ & $0.820\pm0.084$ & $0.734\pm0.136$ & $0.905\pm0.081$ & $0.412\pm0.136$ & $0.806\pm0.117$ & $0.757\pm0.241$ & $0.869\pm0.130$ \\
MESA     & $0.744\pm0.076$ & $0.872\pm0.055$ & $0.806\pm0.084$ & $0.953\pm0.052$ & $0.488\pm0.139$ & $0.857\pm0.076$ & $0.532\pm0.290$ & $0.885\pm0.088$ \\
MROS     & $0.743\pm0.083$ & $0.890\pm0.042$ & $0.825\pm0.069$ & $0.954\pm0.039$ & $0.448\pm0.160$ & $0.870\pm0.064$ & $0.570\pm0.278$ & $0.864\pm0.150$ \\
\hline
\end{tabular}
\end{center}
\end{table}
\end{landscape}

\begin{landscape}
\begin{table}[h]
\renewcommand{\tablename}{Supplementary Table}
\caption{\textbf{SOMNUS$_{\text{EEG}}$ benchmarking.} MF1, Accuracy and Cohen’s $\kappa$, plus Class‐Wise F1 score on recording-level for soft unweighted voting ensembling considering all models using EEG channel derivations.\\}
\label{supp_tab:results_suv_eeg}
\begin{center}
\begin{tabular}{lllllllll}
\hline
\noalign{\vskip 1mm}
\textbf{ID Dataset} & \textbf{MF1} & \textbf{Acc} & \textbf{$\kappa$} & \textbf{W} & \textbf{N1} & \textbf{N2} & \textbf{N3} & \textbf{REM} \\
\hline
\noalign{\vskip 1mm}
ABC & $0.785\pm0.092$ & $0.852\pm0.047$ & $0.780\pm0.056$ & $0.895\pm0.048$ & $0.602\pm0.094$ & $0.866\pm0.066$ & $0.651\pm0.300$ & $0.892\pm0.201$ \\
CCSHS & $0.870\pm0.050$ & $0.925\pm0.038$ & $0.895\pm0.053$ & $0.972\pm0.019$ & $0.645\pm0.145$ & $0.918\pm0.050$ & $0.881\pm0.092$ & $0.934\pm0.041$ \\
CFS & $0.812\pm0.082$ & $0.904\pm0.060$ & $0.853\pm0.098$ & $0.963\pm0.039$ & $0.545\pm0.164$ & $0.886\pm0.096$ & $0.760\pm0.222$ & $0.906\pm0.094$ \\
CHAT & $0.838\pm0.042$ & $0.885\pm0.045$ & $0.844\pm0.061$ & $0.958\pm0.032$ & $0.584\pm0.114$ & $0.856\pm0.078$ & $0.893\pm0.072$ & $0.898\pm0.059$ \\
HOMEPAP & $0.758\pm0.079$ & $0.833\pm0.083$ & $0.752\pm0.133$ & $0.907\pm0.082$ & $0.410\pm0.141$ & $0.821\pm0.104$ & $0.784\pm0.217$ & $0.900\pm0.116$ \\
MESA & $0.741\pm0.084$ & $0.871\pm0.066$ & $0.803\pm0.098$ & $0.947\pm0.074$ & $0.486\pm0.152$ & $0.857\pm0.087$ & $0.517\pm0.291$ & $0.894\pm0.080$ \\
MROS & $0.749\pm0.084$ & $0.896\pm0.041$ & $0.834\pm0.065$ & $0.957\pm0.037$ & $0.456\pm0.167$ & $0.878\pm0.057$ & $0.575\pm0.285$ & $0.868\pm0.155$ \\
\hline
\end{tabular}
\end{center}
\end{table}
\end{landscape}

\begin{landscape}
\begin{table}[h]
\renewcommand{\tablename}{Supplementary Table}
\caption{\textbf{U-Sleep$_{\text{EOG}}$ benchmarking.} MF1, Accuracy and Cohen’s $\kappa$, plus Class‐Wise F1 score on recording‐level for U‐Sleep using EOG channel derivations.}
\label{supp_tab:results_usleep_eog}
\begin{center}
\begin{tabular}{lllllllll}
\hline
\noalign{\vskip 1mm}
\textbf{ID Dataset} & \textbf{MF1} & \textbf{Acc} & \textbf{$\kappa$} & \textbf{W} & \textbf{N1} & \textbf{N2} & \textbf{N3} & \textbf{REM} \\
\hline
\noalign{\vskip 1mm}
ABC      & $0.766\pm0.097$ & $0.838\pm0.058$ & $0.758\pm0.075$ & $0.884\pm0.046$ & $0.581\pm0.079$ & $0.846\pm0.094$ & $0.612\pm0.300$ & $0.888\pm0.200$ \\
CCSHS    & $0.844\pm0.050$ & $0.910\pm0.038$ & $0.873\pm0.052$ & $0.965\pm0.028$ & $0.571\pm0.143$ & $0.901\pm0.051$ & $0.864\pm0.103$ & $0.919\pm0.045$ \\
CFS      & $0.791\pm0.082$ & $0.891\pm0.058$ & $0.835\pm0.094$ & $0.959\pm0.035$ & $0.488\pm0.158$ & $0.868\pm0.098$ & $0.753\pm0.236$ & $0.890\pm0.112$ \\
CHAT     & $0.824\pm0.046$ & $0.877\pm0.048$ & $0.832\pm0.065$ & $0.951\pm0.040$ & $0.544\pm0.115$ & $0.843\pm0.081$ & $0.891\pm0.074$ & $0.891\pm0.067$ \\
HOMEPAP  & $0.726\pm0.122$ & $0.809\pm0.100$ & $0.711\pm0.177$ & $0.873\pm0.149$ & $0.390\pm0.160$ & $0.812\pm0.090$ & $0.681\pm0.263$ & $0.884\pm0.171$ \\
MESA     & $0.728\pm0.094$ & $0.855\pm0.078$ & $0.781\pm0.123$ & $0.941\pm0.070$ & $0.466\pm0.147$ & $0.833\pm0.116$ & $0.504\pm0.299$ & $0.894\pm0.080$ \\
MROS     & $0.730\pm0.090$ & $0.885\pm0.043$ & $0.817\pm0.069$ & $0.953\pm0.036$ & $0.411\pm0.157$ & $0.864\pm0.063$ & $0.540\pm0.293$ & $0.870\pm0.158$ \\
\hline
\end{tabular}
\end{center}
\end{table}
\end{landscape}

\begin{landscape}
\begin{table}[h]
\renewcommand{\tablename}{Supplementary Table}
\caption{\textbf{DeepResNet$_{\text{EOG}}$ benchmarking.} MF1, Accuracy and Cohen’s $\kappa$, plus Class‐Wise F1 score on recording‐level for DeepResNet using EOG channel derivations.}
\label{supp_tab:results_deepresnet_eog}
\begin{center}
\begin{tabular}{lllllllll}
\hline
\noalign{\vskip 1mm}
\textbf{ID Dataset} & \textbf{MF1} & \textbf{Acc} & \textbf{$\kappa$} & \textbf{W} & \textbf{N1} & \textbf{N2} & \textbf{N3} & \textbf{REM} \\
\hline
\noalign{\vskip 1mm}
ABC      & $0.762\pm0.095$ & $0.834\pm0.059$ & $0.755\pm0.073$ & $0.879\pm0.044$ & $0.587\pm0.081$ & $0.846\pm0.092$ & $0.595\pm0.316$ & $0.880\pm0.199$ \\
CCSHS    & $0.841\pm0.048$ & $0.909\pm0.037$ & $0.872\pm0.051$ & $0.967\pm0.020$ & $0.562\pm0.133$ & $0.899\pm0.051$ & $0.865\pm0.093$ & $0.914\pm0.052$ \\
CFS      & $0.786\pm0.085$ & $0.889\pm0.064$ & $0.833\pm0.102$ & $0.956\pm0.044$ & $0.474\pm0.172$ & $0.869\pm0.100$ & $0.732\pm0.254$ & $0.902\pm0.089$ \\
CHAT     & $0.813\pm0.052$ & $0.870\pm0.050$ & $0.824\pm0.068$ & $0.943\pm0.048$ & $0.517\pm0.125$ & $0.841\pm0.082$ & $0.893\pm0.073$ & $0.871\pm0.078$ \\
HOMEPAP  & $0.720\pm0.135$ & $0.797\pm0.129$ & $0.704\pm0.177$ & $0.881\pm0.115$ & $0.405\pm0.155$ & $0.788\pm0.159$ & $0.656\pm0.290$ & $0.877\pm0.178$ \\
MESA     & $0.731\pm0.092$ & $0.858\pm0.072$ & $0.785\pm0.116$ & $0.943\pm0.064$ & $0.485\pm0.145$ & $0.837\pm0.112$ & $0.499\pm0.300$ & $0.888\pm0.076$ \\
MROS     & $0.724\pm0.090$ & $0.882\pm0.044$ & $0.812\pm0.069$ & $0.950\pm0.036$ & $0.396\pm0.149$ & $0.861\pm0.060$ & $0.532\pm0.302$ & $0.869\pm0.145$ \\
\hline
\end{tabular}
\end{center}
\end{table}
\end{landscape}

\begin{landscape}
\begin{table}[h]
\renewcommand{\tablename}{Supplementary Table}
\caption{\textbf{SleepTransformer$_{\text{EOG}}$ benchmarking.} MF1, Accuracy and Cohen’s $\kappa$, plus Class‐Wise F1 score on recording‐level for SleepTransformer using EOG channel derivations.}
\label{supp_tab:results_sleeptransformer_eog}
\begin{center}
\begin{tabular}{lllllllll}
\hline
\noalign{\vskip 1mm}
\textbf{Dataset} & \textbf{MF1} & \textbf{Acc} & \textbf{$\kappa$} & \textbf{W} & \textbf{N1} & \textbf{N2} & \textbf{N3} & \textbf{REM} \\
\hline
\noalign{\vskip 1mm}
ABC      & $0.772\pm0.090$ & $0.831\pm0.053$ & $0.754\pm0.066$ & $0.882\pm0.043$ & $0.609\pm0.080$ & $0.827\pm0.118$ & $0.646\pm0.277$ & $0.881\pm0.199$ \\
CCSHS    & $0.840\pm0.052$ & $0.905\pm0.040$ & $0.867\pm0.055$ & $0.967\pm0.022$ & $0.572\pm0.142$ & $0.892\pm0.055$ & $0.853\pm0.115$ & $0.913\pm0.058$ \\
CFS      & $0.787\pm0.085$ & $0.883\pm0.066$ & $0.825\pm0.106$ & $0.956\pm0.043$ & $0.483\pm0.166$ & $0.856\pm0.106$ & $0.747\pm0.239$ & $0.897\pm0.093$ \\
CHAT     & $0.840\pm0.041$ & $0.880\pm0.043$ & $0.838\pm0.058$ & $0.957\pm0.035$ & $0.614\pm0.096$ & $0.842\pm0.079$ & $0.891\pm0.073$ & $0.895\pm0.054$ \\
HOMEPAP  & $0.758\pm0.086$ & $0.828\pm0.072$ & $0.741\pm0.122$ & $0.904\pm0.068$ & $0.482\pm0.156$ & $0.819\pm0.100$ & $0.695\pm0.258$ & $0.908\pm0.076$ \\
MESA     & $0.737\pm0.095$ & $0.855\pm0.082$ & $0.784\pm0.119$ & $0.945\pm0.074$ & $0.504\pm0.139$ & $0.829\pm0.110$ & $0.512\pm0.296$ & $0.891\pm0.089$ \\
MROS     & $0.727\pm0.089$ & $0.880\pm0.044$ & $0.810\pm0.070$ & $0.951\pm0.039$ & $0.426\pm0.157$ & $0.856\pm0.062$ & $0.514\pm0.298$ & $0.874\pm0.144$ \\
\hline
\end{tabular}
\end{center}
\end{table}
\end{landscape}

\begin{landscape}
\begin{table}[h]
\renewcommand{\tablename}{Supplementary Table}
\caption{\textbf{SOMNUS$_{\text{EOG}}$ benchmarking.} MF1, Accuracy and Cohen’s $\kappa$, plus Class‐Wise F1 score on recording‐level for soft unweighted voting ensembling using EOG channel derivations.}
\label{supp_tab:results_suv_eog}
\begin{center}
\begin{tabular}{lllllllll}
\hline
\noalign{\vskip 1mm}
\textbf{Dataset} & \textbf{MF1} & \textbf{Acc} & \textbf{$\kappa$} & \textbf{W} & \textbf{N1} & \textbf{N2} & \textbf{N3} & \textbf{REM} \\
\hline
\noalign{\vskip 1mm}
ABC      & $0.783\pm0.091$ & $0.847\pm0.053$ & $0.775\pm0.066$ & $0.892\pm0.042$ & $0.611\pm0.082$ & $0.854\pm0.096$ & $0.650\pm0.298$ & $0.890\pm0.200$ \\
CCSHS    & $0.854\pm0.050$ & $0.915\pm0.036$ & $0.881\pm0.050$ & $0.970\pm0.019$ & $0.602\pm0.141$ & $0.906\pm0.051$ & $0.869\pm0.100$ & $0.924\pm0.048$ \\
CFS      & $0.801\pm0.081$ & $0.896\pm0.062$ & $0.842\pm0.099$ & $0.960\pm0.039$ & $0.506\pm0.162$ & $0.874\pm0.099$ & $0.760\pm0.239$ & $0.908\pm0.087$ \\
CHAT     & $0.839\pm0.044$ & $0.885\pm0.046$ & $0.844\pm0.062$ & $0.957\pm0.036$ & $0.591\pm0.108$ & $0.852\pm0.081$ & $0.897\pm0.073$ & $0.898\pm0.060$ \\
HOMEPAP  & $0.751\pm0.095$ & $0.826\pm0.083$ & $0.740\pm0.134$ & $0.903\pm0.075$ & $0.433\pm0.161$ & $0.825\pm0.087$ & $0.708\pm0.257$ & $0.899\pm0.122$ \\
MESA     & $0.741\pm0.094$ & $0.864\pm0.075$ & $0.794\pm0.119$ & $0.947\pm0.065$ & $0.498\pm0.149$ & $0.841\pm0.113$ & $0.515\pm0.305$ & $0.900\pm0.077$ \\
MROS     & $0.735\pm0.088$ & $0.889\pm0.042$ & $0.823\pm0.066$ & $0.955\pm0.036$ & $0.433\pm0.158$ & $0.868\pm0.059$ & $0.534\pm0.302$ & $0.876\pm0.149$ \\
\hline
\end{tabular}
\end{center}
\end{table}
\end{landscape}

\begin{landscape}
\begin{table}[h]
\renewcommand{\tablename}{Supplementary Table}
\caption{\textbf{Performance SleepTransformer SHHS.} MF1, Accuracy and Cohen’s $\kappa$, plus Class‐Wise F1 score on recording-level for SleepTransformer making use of EEG channel derivations (majority voting) and trained only on the SHHS dataset, as opposed to the entire NSRR repository.\\}
\label{supp_tab:results_sleeptransformer_shhs}
\begin{center}
\begin{tabular}{lllllllll}
\hline
\noalign{\vskip 1mm}
\textbf{Dataset} & \textbf{MF1} & \textbf{Acc} & \textbf{$\kappa$} & \textbf{W} & \textbf{N1} & \textbf{N2} & \textbf{N3} & \textbf{REM} \\
\hline
ABC & $0.674\pm0.083$ & $0.756\pm0.065$ & $0.643\pm0.089$ & $0.831\pm0.077$ & $0.307\pm0.109$ & $0.765\pm0.107$ & $0.616\pm0.236$ & $0.850\pm0.194$ \\
CCSHS & $0.716\pm0.086$ & $0.810\pm0.085$ & $0.738\pm0.113$ & $0.902\pm0.098$ & $0.344\pm0.151$ & $0.774\pm0.120$ & $0.746\pm0.156$ & $0.815\pm0.098$ \\
CFS & $0.693\pm0.116$ & $0.822\pm0.099$ & $0.737\pm0.140$ & $0.910\pm0.097$ & $0.320\pm0.179$ & $0.775\pm0.145$ & $0.694\pm0.225$ & $0.762\pm0.222$ \\
CHAT & $0.418\pm0.120$ & $0.522\pm0.127$ & $0.352\pm0.167$ & $0.673\pm0.186$ & $0.021\pm0.035$ & $0.315\pm0.198$ & $0.631\pm0.127$ & $0.450\pm0.272$ \\
HOMEPAP & $0.615\pm0.088$ & $0.722\pm0.109$ & $0.599\pm0.139$ & $0.851\pm0.097$ & $0.149\pm0.119$ & $0.676\pm0.131$ & $0.630\pm0.233$ & $0.789\pm0.235$ \\
MESA & $0.573\pm0.117$ & $0.739\pm0.096$ & $0.611\pm0.150$ & $0.860\pm0.112$ & $0.145\pm0.111$ & $0.704\pm0.157$ & $0.463\pm0.310$ & $0.694\pm0.238$ \\
MROS & $0.637\pm0.093$ & $0.811\pm0.069$ & $0.706\pm0.102$ & $0.931\pm0.048$ & $0.228\pm0.137$ & $0.744\pm0.107$ & $0.509\pm0.246$ & $0.761\pm0.199$ \\
\hline
\end{tabular}
\end{center}
\end{table}
\end{landscape}

\newpage

\begin{landscape}
\begin{table}
\renewcommand{\tablename}{Supplementary Table}
\caption{Agreement, expressed by Cohen's $\kappa$ (mean $\pm$ std), of models with human scorers and consensus on DOD-O and DOD-H datasets. The models are ordered by their agreement with respect to the consensus hypnogram. In the last columns averaged cosine similarity (ACS) quantifies the similarity between the hypnodensities implied by the soft-consensus and the predictions of the models.}
\label{supp_tab:consensus_aggrement_suppl}
\begin{tabular}{l|l|llllll|l|}
\toprule
 &  & Expert 1 & Expert 2 & Expert 3 & Expert 4 & Expert 5 & Consensus & ACS \\
 & Model &  &  &  &  &  &  &  \\
\midrule
\multirow{10}{*}{\rotatebox{90}{DOD-H}} & SOMNUS & 78.8 ± 17.0 & 81.7 ± 6.1 & 81.0 ± 10.2 & 74.7 ± 12.1 & 82.1 ± 8.7 & $\mathbf{88.9 \pm 4.7}$ & $\mathbf{0.947 \pm 0.021}$ \\
 & SOMNUS$_\text{SleepTransformer}$ & 78.1 ± 17.1 & 81.2 ± 6.3 & 81.4 ± 9.7 & 74.3 ± 11.6 & 81.9 ± 8.5 & 88.4 ± 4.7 & 0.945 ± 0.021 \\
 & SOMNUS$_\text{EEG/EOG}$ & 77.9 ± 16.7 & 81.0 ± 6.4 & 80.0 ± 10.6 & 74.2 ± 12.6 & 81.5 ± 9.2 & 87.9 ± 5.6 & 0.942 ± 0.023 \\
 & SOMNUS$_\text{DeepResNet}$ & 78.0 ± 16.8 & 80.6 ± 6.8 & 80.1 ± 10.8 & 73.8 ± 12.6 & 81.3 ± 9.7 & 87.7 ± 5.9 & 0.943 ± 0.023 \\
 & SOMNUS$_\text{EEG}$ & 78.3 ± 17.0 & 80.7 ± 6.4 & 80.6 ± 10.7 & 72.9 ± 11.8 & 80.8 ± 9.6 & 87.4 ± 5.8 & 0.938 ± 0.023 \\
 & Single Model (Best) & 76.3 ± 16.5 & 79.6 ± 6.2 & 80.6 ± 9.0 & 72.4 ± 11.2 & 80.2 ± 7.6 & 86.9 ± 4.7 & 0.939 ± 0.026 \\
 & SOMNUS$_\text{U-Sleep}$ & 77.1 ± 16.5 & 80.2 ± 7.1 & 78.5 ± 11.3 & 73.1 ± 13.2 & 80.0 ± 10.1 & 86.0 ± 6.7 & 0.936 ± 0.028 \\
 & SOMNUS$_\text{EOG}$ & 76.0 ± 17.3 & 78.8 ± 8.7 & 78.2 ± 11.4 & 73.0 ± 13.4 & 79.3 ± 10.7 & 85.5 ± 8.5 & 0.931 ± 0.047 \\
 & Single Model (Avg) & 74.5 ± 17.9 & 77.5 ± 9.7 & 76.8 ± 12.8 & 70.7 ± 14.3 & 77.6 ± 11.8 & 83.4 ± 9.4 & 0.924 ± 0.036 \\
 & Single Model (Worst) & 69.1 ± 21.9 & 72.5 ± 17.3 & 69.7 ± 18.9 & 66.7 ± 21.6 & 71.9 ± 18.7 & 77.1 ± 18.3 & 0.899 ± 0.059 \\
\cline{1-9}
\multirow{10}{*}{\rotatebox{90}{DOD-O}} & SOMNUS & 74.9 ± 13.6 & 77.5 ± 10.3 & 73.0 ± 12.7 & 78.9 ± 9.7 & 79.6 ± 9.6 & $\mathbf{85.4 \pm 8.0}$ & 0.939 ± 0.024 \\
 & SOMNUS$_\text{U-Sleep}$ & 73.7 ± 14.1 & 76.1 ± 10.4 & 73.4 ± 12.8 & 78.9 ± 9.8 & 79.6 ± 9.3 & 84.7 ± 7.7 & $\mathbf{0.941 \pm 0.026}$ \\
 & SOMNUS$_\text{EEG/EOG}$ & 73.7 ± 14.1 & 77.4 ± 11.0 & 72.4 ± 12.7 & 79.2 ± 9.0 & 79.3 ± 9.6 & 84.6 ± 8.2 & 0.935 ± 0.026 \\
 & SOMNUS$_\text{DeepResNet}$ & 74.1 ± 13.9 & 76.1 ± 10.6 & 72.9 ± 12.8 & 78.5 ± 9.4 & 79.0 ± 9.5 & 84.6 ± 7.9 & 0.939 ± 0.024 \\
 & Single Model (Best) & 73.4 ± 14.1 & 75.9 ± 10.8 & 72.7 ± 12.6 & 78.6 ± 8.9 & 78.7 ± 9.3 & 84.2 ± 7.6 & 0.939 ± 0.023 \\
 & SOMNUS$_\text{EOG}$ & 71.6 ± 14.0 & 72.6 ± 11.7 & 72.1 ± 13.3 & 77.5 ± 9.0 & 77.4 ± 9.0 & 81.7 ± 8.2 & 0.922 ± 0.028 \\
 & SOMNUS$_\text{EEG}$ & 73.3 ± 13.3 & 77.3 ± 9.9 & 69.4 ± 14.0 & 73.5 ± 13.4 & 75.2 ± 12.3 & 81.1 ± 12.0 & 0.915 ± 0.044 \\
 & Single Model (Avg) & 66.7 ± 14.2 & 69.2 ± 12.2 & 65.1 ± 13.8 & 70.2 ± 11.4 & 70.8 ± 11.7 & 74.9 ± 10.9 & 0.892 ± 0.048 \\
 & SOMNUS$_\text{SleepTransformer}$ & 66.7 ± 13.3 & 70.3 ± 11.9 & 63.4 ± 13.0 & 69.5 ± 11.1 & 70.0 ± 12.1 & 74.7 ± 11.8 & 0.881 ± 0.052 \\
 & Single Model (Worst) & 47.9 ± 14.6 & 52.8 ± 15.1 & 41.3 ± 17.2 & 44.7 ± 17.3 & 45.6 ± 18.2 & 49.3 ± 17.5 & 0.768 ± 0.091 \\
\cline{1-9}
\bottomrule
\end{tabular}
\end{table}
\end{landscape}


\title[SLEEPYLAND]{SLEEPYLAND: trust begins with fair evaluation of automatic sleep staging models}


\author[1,2]{\fnm{Alvise} \sur{Dei Rossi}}\email{alvise.dei.rossi@usi.ch}

\author[2]{\fnm{Matteo} \sur{Metaldi}}\email{matteo.metaldi@supsi.ch}

\author[2,3]{\fnm{Michal} \sur{Bechny}}\email{michal.bechny@supsi.ch}

\author[4]{\fnm{Irina} \sur{Filchenko}}\email{irina.filchenko@insel.ch}

\author[4]{\fnm{Julia} \sur{van der Meer}}\email{julia.vandermeer@insel.ch}

\author[4]{\fnm{Markus H.} \sur{Schmidt}}\email{markus.schmidt@insel.ch}

\author[4]{\fnm{Claudio L.A.} \sur{Bassetti}}\email{claudio.bassetti@insel.ch}

\author[3,4]{\fnm{Athina} \sur{Tzovara}}\email{athina.tzovara@inf.unibe.ch}

\author[2]{\fnm{Francesca D.} \sur{Faraci}}\email{francesca.faraci@supsi.ch}

\author*[2,5]{\fnm{Luigi} \sur{Fiorillo}}\email{luigi.fiorillo@supsi.ch}

\affil[1]{\orgdiv{Faculty of informatics}, \orgname{Università della Svizzera Italiana}, \orgaddress{\street{Via Giuseppe Buffi 13}, \city{Lugano}, \postcode{6900}, \country{Switzerland}}}

\affil*[2]{\orgdiv{Institute of Digital Technologies for Personalized Healthcare $\vert$ MeDiTech}, Department of Innovative Technologies, \orgname{University of Applied Sciences and Arts of Southern Switzerland}, \orgaddress{\street{Via la Santa 1}, \city{Lugano}, \postcode{6962}, \country{Switzerland}}}

\affil[3]{\orgdiv{Institute of Informatics}, \orgname{University of Bern}, \orgaddress{\street{Neubr{\"u}ckstrasse 10}, \city{Bern}, \postcode{3012}, \country{Switzerland}}}

\affil[4]{\orgdiv{Sleep Wake Epilepsy Center $\vert$ NeuroTec},  Department of Neurology, \orgname{Inselspital, Bern University Hospital, University of Bern}, \orgaddress{\street{Freiburgstrasse}, \city{Bern}, \postcode{3010}, \country{Switzerland}}}

\affil*[5]{\orgdiv{Neurocenter of Southern Switzerland}, \orgname{Ente Ospedaliero Cantonale}, \orgaddress{\street{Via Tesserete 46}, \city{Lugano}, \postcode{6900}, \country{Switzerland}}}

\maketitle

\clearpage

\section*{Supplementary notes}

\subsection*{Dataset}
\label{supp_note:dataset}

We report a detailed description of all the datasets used in our experiments.

\subsubsection*{NSRR datasets}

\textbf{ABC}. The Apnea, Bariatric surgery, and CPAP database consists of 132 recordings from 49 patients with severe obstructive sleep apnea and morbid obesity (BMI from 35 to 45) \cite{zhang2018national, bakker2018gastric}. EEG signals (F4-M1, F3-M2, C4-M1, C3-M2, O2-M1, O1-M2) and EOG signals (E2-M1, E1-M2) are considered in our experiments. The signals are recorded at 256Hz, and are hardware low-pass filtered at 105Hz and high-pass filtered at 0.16Hz. The recordings are manually scored by sleep experts according to the AASM rules. For more information we refer to \url{https://doi.org/10.25822/nx52-bc11} and \url{https://clinicaltrials.gov/ct2/show/NCT01187771}.\\

\textbf{APOE}. The Sleep Disordered Breathing, apolipoprotein E (ApoE), and Lipid Metabolism dataset is an NIH-supported study investigating genetic associations in ApoE e4-positive and e4-negative individuals with varying degrees of sleep apnea \cite{zhang2018national, moore2014design}. We consider 712 recordings from suspected but untreated sleep-disordered breathing participants. EEG signals (C3-M2, C4-M1, O2-M1, O1-M2, C3-M1, C4-M2, O2-M2, O1-M1, F1-M2, F2-C4, F2-T4, FP1-C3, FP1-C3, FP2-C4, Fz-M1, Fz-M2, T3-O1 T4-O2) and EOG signals (ROC-M1, LOC-M2) are included. The signals are recorded at 256Hz using the Sandman Elite system. The recordings are manually scored by sleep experts following the AASM rules. For more information we refer to \url{https://doi.org/10.25822/6ssj-2157}.\\

\textbf{APPLES}. \textbf{APPLES}. The Apnea Positive Pressure Long-term Efficacy Study (APPLES) is a 6-month, multi-center, randomized, double-blind, sham-controlled trial across five U.S. sites \cite{zhang2018national, quan2011association}. In our experiments we consider 1094 recordings from 1098 OSA participants. EEG signals (C3-M2, C4-M1, O2-M1, O1-M2) and EOG signals (ROC-M1, LOC-M2) are included. The signals are recorded at 128 Hz. The recordings are manually scored by sleep experts according to the Rechtschaffen and Kales rules, and re-aligned to the AASM rules. For more information we refer to \url{https://doi.org/10.25822/63pr-a591} and \url{https://clinicaltrials.gov/ct2/show/results/NCT00051363}.\\

\textbf{CCSHS}. The Cleveland Childrend's Sleep and Health Study consists of children and adolescents recordings. In our experiments we consider 515 recordings from adolescents aged 16-19 years. A small percentage of the subjects suffers from sleep related movement disorders. The recordings are collected in three different hospitals around Cleveland, Ohio, US \cite{zhang2018national, rosen2003prevalence}. EEG signals (C4-A1, C3-A2) and EOG signals (ROC-A1, LOC-A2) are considered in our experiments. The signals are recorded at 128Hz, and hardware high-pass filtered at 0.15Hz. The recordings are manually scored by sleep experts according to the AASM rules.  For more information we refer to \url{https://doi.org/10.25822/cg2n-4y91}.\\

\textbf{CFS}. The Cleveland Family Study is a family-based study on sleep apnea disordered subjects. The database consists of 2284 subjects from 361 families \cite{zhang2018national, redline1995familial}. We consider recordings of 730 subjects from 144 families (whence full whole-night PSG were available). For this specific database, the data split (train/val/test set) is done by considering subjects and family belonging (\textit{i.e.}, all the family members appear in the same data split). {More than half of the subjects are affected by sleep apnea disorder.} EEG signals (C4-A1, C3-A2) and EOG signals (ROC-A1, LOC-A2) are considered in our experiments. The signals are recorded at 128Hz, and hardware low-pass filtered and high-pass filtered at 105Hz and 0.16 Hz respectively. The recordings are manually scored by sleep experts according to the AASM rules. For more information we refer to \url{https://doi.org/10.25822/jmyx-mz90}.\\

\textbf{CHAT}. The Childhood Adenotonsillectomy Trial database consists of 1638 recordings (452 baseline, 407 follow-up and 779 control) from 1232 children post-adenotonsillectomy-surgery aged 5-10 years { with mild to moderate obstructive sleep apnea}. The recordings are collected in six different sleep centers in Massachusetts, Missouri, New York, Ohio and Pennsylvania \cite{zhang2018national, marcus2013randomized, redline2011childhood}.  EEG signals  (F4-M1, F3-M2, C4-M1, C3-M2, O2-M1, O1-M2, T4-M1, T3-M2) and EOG signals (E2-M1, E1-M2) are considered in our experiments. The signals are recorded at 200Hz (or higher in other sleep centers), and different hardware filtering given the different acquisition systems. One recording is excluded - EOG missing. The recordings are manually scored by sleep experts according to the AASM rules. For more information we refer to \url{https://doi.org/10.25822/d68d-8g03} and \url{https://clinicaltrials.gov/ct2/show/NCT00560859}.\\

\textbf{HOMEPAP}. The Home Positive Airway Pressure database consists of 373 recordings (246 considered in our experiments) from obstructive sleep apnea patients aged over 18 years. The recordings are collected in seven different US sleep centers \cite{zhang2018national, rosen2012multisite}. EEG signals  (F4-M1, F3-M2, C4-M1, C3-M2, O2-M1, O1-M2, T4-M1, T3-M2) and EOG signals (E2-M1, E1-M2) are considered in our experiments. The signals are recorded at 200Hz, no filtering applied. Nine recordings are excluded - EOG and/or reference channels missing. The recordings are manually scored by sleep experts according to the AASM rules. For more information we refer to \url{https://doi.org/10.25822/xmwv-yz91} and \url{https://clinicaltrials.gov/ct2/show/NCT00642486}.\\

\textbf{MESA}. The Multi-Ethnic Study of Atherosclerosis consists of 2237 recordings (2056 considered in our experiments) from a cohort of black, white, Hispanic and Chinese-American subjects aged 45-84 years \cite{zhang2018national, chen2015racial}. {About 15.0\% of individuals have severe SDB, 30.9\% short sleep duration, 6.5\% poor sleep quality and 13.9\%  daytime sleepiness.} EEG signals (Fz-Cz, C4-M1, Cz-Oz) and EOG signals (E2-Fpz, E1-Fpz) are considered in our experiments. The signals are recorded at 256Hz, and hardware low-pass filtered at 100Hz. The recordings are manually scored by sleep experts according to the AASM rules. For more information we refer to \url{https://doi.org/10.25822/n7hq-c406}.\\

\textbf{MNC}. The Mignot Nature Communications dataset contains raw polysomnography data first exploited in a neural-network–based automated sleep staging project \cite{stephansen2018neural}. We consider recordings from approximately 1000 normal and abnormal subjects. EEG signals (C3-M2, C3, C4-M1, C4, Cz, F3-M2, F3, F4-M1, F4, O1-M2, O1, O2-M1, O2) and EOG signals (E1-M2 E1 E2-M1 E2) are recorded at 128Hz. The recordings are manually scored by sleep experts according to the AASM rules. For more information we refer to \url{https://doi.org/10.25822/00tc-zz78}.\\

\textbf{MROS}. The database is a subset of the larger study Osteoporotic Fractures in Men, involving 5994 community-dwelling men aged over 65 years \cite{zhang2018national, blackwell2011associations, osteoporotic2015relationships}. In our experiments we consider 3930 recordings from subjects which underwent in-home overnight PSG. Most of the subjects are sleep breathing disorders patients. EEG signals (C4-A1, C3-A2) and EOG signals (ROC-A1, LOC-A2) are considered in our experiments. The signals are recorded at 256Hz, and hardware high-pass filtered at 0.15Hz. Seven recordings are excluded - EOG channels and/or sleep stage annotation files missing. The recordings are manually scored by sleep experts according to the AASM rules. For more information we refer to \url{https://doi.org/10.25822/kc27-0425}.\\

\textbf{MSP}. The Maternal Sleep in Pregnancy and the Fetus (MSP) dataset includes 106 overnight laboratory-based PSG recordings from women in their 36th week of pregnancy \cite{zhang2018national, dipietro2021fetal, dipietro2023fetal}.  Eligibility was restricted to non-smoking women with pre-pregnancy obesity (BMI $>$ 30~kg/m$^2$), without previously identified sleep disorders or significant conditions compromising mother or fetus (hypertension and diabetes were not exclusion criteria). In our experiments we consider recordings from 105 subjects. EEG signals (C3-M2, C4-M1, F3-M2, F4-M1, O1-M2, O2-M1) and EOG signals (LOC, ROC) are included. Signals are recorded at 256Hz. Recordings are manually scored by sleep experts according to AASM criteria. For more information we refer to \url{https://sleepdata.org/datasets/msp}.\\

\textbf{NCHSDB}. The Nationwide Children’s Hospital Sleep DataBank consists of 3984 pediatric sleep studies from 3673 patients aged 0–18 years, collected between 2017–2019 at Nationwide Children’s Hospital, Columbus, Ohio \cite{zhang2018national, lee2022large}. In our experiments we consider recordings from 3950 subjects. EEG signals (FP1, FP2, FZ, CZ, PZ, OZ, FPZ, P3-M2, P4-M1, F3-M2, F4-M1, F4-M2, C3-M2, C4-M1, C4-M2, T3-M2, T4-M1, O1-M2, O2-M1, F4, O1, O2) and EOG signals (E1-M2, E2-M1, E1, E2) are included. Signals are recorded at 256Hz. Recordings are manually scored by sleep technicians following AASM criteria and include longitudinal clinical data. For more information we refer to \url{https://sleepdata.org/datasets/nchsdb}.\\

\textbf{SHHS}. The Sleep Heart Health Study consists of 8444 recordings (5793 from visit 1 and 2651 from visit 2) from 5797 subjects aged over 40 years \cite{zhang2018national, quan1997sleep}. {Most of the subjects suffer from OSA or other SDB.} EEG signals (C4-A1, C3-A2) and EOG signals (ROC-A1, LOC-A2) are considered in our experiments. The EEG and EOG signals are recorded at 125Hz and 50Hz respectively, and hardware high-pass filtered at 0.15Hz. The recordings are manually scored by sleep experts according to the Rechtschaffen and Kales scoring rules, and re-aligned to the AASM rules. For more information we refer to \url{https://clinicaltrials.gov/ct2/show/NCT00005275} and \url{https://doi.org/10.25822/ghy8-ks59}.\\

\textbf{SOF}. The database is a subset of the larger study Osteoporotic Fractures. In our experiments we consider 453 recordings (from visit 8), which underwent in-home overnight PSG \cite{zhang2018national, cummings1990appendicular, spira2008sleep}. EEG signals (C4-A1, C3-A2) and EOG signals (ROC-A1, LOC-A2) are considered in our experiments. The EEG and EOG signals are recorded at 128Hz, and hardware high-pass filtered at 0.15Hz. The recordings are manually scored by sleep experts according to the Rechtschaffen and Kales scoring rules, and re-aligned to the AASM rules. For more information we refer to \url{https://doi.org/10.25822/e1cf-rx65}.\\

\textbf{WSC}. The Wisconsin Sleep Cohort is a longitudinal study of 1500 Wisconsin state employees, assessed at four-year intervals, investigating sleep disorders, particularly obstructive sleep apnea \cite{zhang2018national, young2009burden}.  In our experiments we consider 2569 recordings (recordings from visit 1 to visit 4). EEG signals (F3-M1, F3-M2, F3-AVG, F4-M1, F4-M2, F4-AVG, Fz-M1, Fz-M2, Fz-AVG, Cz-M1, Cz-M2, Cz-AVG, C3-M1, C3-M2, C3-AVG, C4-M1, C4-M2, C4-AVG, Pz-M1, Pz-M2, Pz-AVG, Pz-Cz, O1-M1, O1-M2,
O1-AVG, O2-M1, O2-M2, O2-AVG) and EOG signals (E1, E2) are included. Signals are recorded at 100Hz-200Hz during in-laboratory PSG. Recordings are manually scored by sleep experts following AASM criteria. For more information we refer to \url{https://sleepdata.org/datasets/wsc}.

\subsubsection*{Out-of-domain datasets}

\textbf{BSWR}.
The Bern Sleep-Wake Registry consists of 8950 recordings from 7985 subjects. A small percentage of the subjects is healthy (below 1\%). The rest of the subjects are patients with a single or multiple sleep disorders or with an uncertain diagnosis. The most common class of sleep disorders is sleep related breathing disorders, followed by central disorders of hypersomnolence, insomnia, parasomnias and sleep related movement disorders. A smaller percentage of patients with circadian rhythm sleep-wake disorders and isolated symptoms and normal variants is also present. EEG signals (F4-M1, F3-M2, C4-M1, C3-M2, O2-M1, O1-M2) and EOG signals (E2-M1, E1-M2) are considered in our experiments. The signals are recorded at 200Hz. The recordings are manually scored by sleep experts according to the AASM rules. Given the different scoring rules for infants ($\leq 2$ months) \cite{berry2017aasm}, it is important to specify, in the context of the following age analysis, that in the BSWR dataset there were no babies younger than two months.\\

\textbf{DCSM}. The Danish Centre for Sleep Medicine database consists of 255 recordings from patients with potential and non-specific sleep related disorders \cite{perslev2021u}. No demographic information is available for the database. EEG signals  (F4-M1, F3-M2, C4-M1, C3-M2, O2-M1, O1-M2, T4-M1, T3-M2) and EOG signals (E2-M1, E1-M2) are considered in our experiments. The signals are recorded at 256Hz, and band-pass filtered between 0.3Hz and 70Hz. The recordings are manually scored by sleep experts according to the AASM rules. For more information we refer to \url{https://sid.erda.dk/wsgi-bin/ls.py?share_id=fUH3xbOXv8}.\\

\textbf{DOD-H \& DOD-O}. The DOD-H dataset contains 25 recordings (19 males and 6 females) from healthy adult volunteers aged from 18 to 65 years. The recordings were collected at the French Armed Forces Biomedical Research Institute’s (IRBA) Fatigue and Vigilance Unit (Bretigny-Sur-Orge, France). EEG signals (C3-M2, C4-M1, F3-F4, F3-M2, F3-O1, F4-O2, O1-M2, O2-M1) and left/right EOG signals are considered. The recordings are sampled at 512 Hz. The DOD-O dataset contains 55 recordings (35 males and 20 females) from patients suffering from obstructive sleep apnea aged from 39 to 62 years. The recordings were collected at the Stanford Sleep Medicine Center. EEG signals (C3-M2, C4-M1, F4-M1, F3-F4, F3-M2, F3-O1, F4-O2, FP1-F3, FP1-M2, FP1-O1, FP2-F4, FP2-M1, FP2-O2) and left/right EOG signlas are considered. The recordings are sampled at 250 Hz. As in \cite{guillot2020dreem}, a band-pass Butterworth IIR filter is applied between [0.4, 18] Hz to remove residual PSG noise, and the signals are resampled at 100 Hz. The signals are then clipped and divided by 500 to remove extreme values. The recordings are scored by five physicians from three different sleep centers according to the AASM rules. For more information we refer to \cite{guillot2020dreem}. \\

\textbf{PHYS}. The database from the 1028 PhysioNet/CinC Challenge consists of 1985 recordings (994 labelled considered in our experiments) from patients with potential sleep disorders \cite{goldberger2000physiobank, ghassemi2018you}. EEG signals (F4-M1, F3-M2, C4-M1, C3-M2, O2-M1, O1-M2) and one EOG signal (E1-M2) are considered in our experiments. The signals are recorded at 200Hz. The recordings are manually scored by sleep experts according to the AASM rules. For more information we refer to \url{https://physionet.org/content/challenge-2018/1.0.0/}.\\

\textbf{SEDF-SC \& SEDF-ST}. The Sleep-EDF Expanded database consists of 197 recordings from two subset studies. The Sleep-EDF Sleep Cassette consists of 153 recordings from 78 healthy subjects aged 25-101 years. The Sleep-EDF Sleep Telemetry consists of 44 recordings from 22 healthy subjects with mild difficulty falling asleep (two recordings collected for each subject, \textit{i.e.}, one after temazepam intake and one after placebo intake) \cite{goldberger2000physiobank, kemp2000analysis}. EEG signals (Fpz-Cz, Pz-Oz) and one EOG signal (ROC-LOC) are considered in our experiments. The signals are recorded at 100Hz. The recordings are manually scored by sleep experts according to the Rechtschaffen and Kales scoring rules, and re-aligned to the AASM rules. For more information we refer to \url{https://doi.org/10.13026/C2C30J}.

\subsection*{Calculation of expected values from GAMLSS}
\label{supp_note:GAMLSS_expected_value}

In GAMLSS, distributional parameters are modeled using additive predictors and link functions. That is, each parameter $\theta$ is expressed as $\eta_\theta = g_\theta(\theta)$, where $g_\theta(\cdot)$ is a link function specific to that parameter and distribution. To compute interpretable quantities such as expectations, we apply the inverse link function: $\theta = g_\theta^{-1}(\eta_\theta)$.

\subsubsection*{Expected value formulas}

For the zeros-and-ones-inflated Beta (BEINF) distribution, used for bounded performance metrics such as MF1, the expected value is:

\[
\mathbb{E}[Y] = \tau + (1 - \nu - \tau) \cdot \mu
\]

where:\\

\begin{itemize}
    \item $\mu$ is the mean of the continuous Beta part,
    \item $\nu$ is the probability mass at 0,
    \item $\tau$ is the probability mass at 1.\\
\end{itemize}

These parameters are modeled using the following link functions:

\[
\begin{aligned}
\eta_\mu &= \text{logit}(\mu) \quad &\Rightarrow\quad \mu &= \text{logit}^{-1}(\eta_\mu) = \frac{1}{1 + \exp(-\eta_\mu)} \\
\eta_\nu &= \log(\nu) \quad &\Rightarrow\quad \nu &= \exp(\eta_\nu) \\
\eta_\tau &= \log(\tau) \quad &\Rightarrow\quad \tau &= \exp(\eta_\tau)
\end{aligned}
\]

For Gaussian-distributed bias outcomes (e.g., TST = $\widehat{\text{TST}}_{\text{algorithm}} - \text{TST}_{\text{reference}}$), the expectation is straightforward:
\[
\mathbb{E}[Y] = \mu
\]

with \[
\eta_\mu = \mu \quad \text{(identity link)}, \quad \Rightarrow \mu = \eta_\mu
\]

\subsubsection*{Worked examples}

We provide two examples using the SOMNUS model: one for a baseline subject and one for a non-baseline case. Refer to Table~4 and Table~5 in the manuscript to retrieve the values. \\

\noindent\textbf{Example 1: Baseline subject (female, age 50, AHI = 0, PLMI = 0)}\\

\textbf{\textit{MF1 (performance metric):}}\\

The intercepts for MF1 are:

\[
\eta_\mu = 1.12, \quad \eta_\nu = -22.54, \quad \eta_\tau = -22.63
\]

Transforming back via the inverse link functions:

\[\mu = \text{logit}^{-1}(1.12) = \frac{1}{1 + \exp(-1.12)} \approx 0.754, \] \\
\[\nu = \exp(-22.54) \approx 1.62 \times 10^{-10},\] \\
\[\tau = \exp(-22.63) \approx 1.48 \times 10^{-10} \]

Then the expected value is:

\[
\mathbb{E}[Y] \approx 1.48 \times 10^{-10} + (1 - 1.62 \times 10^{-10} - 1.48 \times 10^{-10}) \cdot 0.754 \approx 0.754
\]

\textbf{\textit{TST bias (in minutes):}}\\

The intercept for TST (Normal distribution) is:

\[
\eta_\mu = -9.05 \quad \text{(identity link)} \quad \Rightarrow \quad \mathbb{E}[Y] = -9.05
\]\\

\noindent\textbf{Example 2: Male, age 50, AHI = 30, PLMI = 10}\\

\textbf{\textit{MF1 (performance metric):}}\\

The relevant coefficients are:

\[
\beta_\mu^{\text{gender}} = -0.06, \quad \beta_\mu^{\text{AHI}} = -0.05, \quad \beta_\mu^{\text{PLMI}} = -0.02
\]

The linear predictor becomes:

\[
\eta_\mu = 1.12 - 0.06 - 3 \cdot 0.05 - 1 \cdot 0.02 = 0.89
\]

\[
\mu = \text{logit}^{-1}(0.89) = \frac{1}{1 + \exp(-0.89)} \approx 0.709
\]

With $\eta_\nu = -22.54$, $\eta_\tau = -22.63$ as before:

\[
\mathbb{E}[Y] \approx \exp(-22.63) + (1 - \exp(-22.54) - \exp(-22.63)) \cdot 0.709 \approx 0.709
\]

\textbf{\textit{TST bias (in minutes):}}\\

Coefficients for the linear predictor:

\[
\beta_\mu^{\text{gender}} = -1.37, \quad \beta_\mu^{\text{AHI}} = -1.95, \quad \beta_\mu^{\text{PLMI}} = -0.47
\]

\[
\eta_\mu = -9.05 - 1.37 - 3 \cdot 1.95 - 1 \cdot 0.47 = -16.74 \quad \Rightarrow \quad \mathbb{E}[Y] = -16.74
\]

\bibliography{sn-bibliography}